\documentclass[twoside,11pt]{article}

\usepackage[T1]{fontenc}
\usepackage[utf8]{inputenc}
\usepackage{textcomp}
\usepackage{amsmath}
\usepackage{amsthm}
\usepackage{amssymb}
\usepackage[dvipsnames]{xcolor}
\usepackage{pifont}
\usepackage{marvosym}
\usepackage[colorinlistoftodos]{todonotes}

\usepackage[colorlinks=true,linkcolor=blue,urlcolor=blue]{hyperref}

\makeatletter
\usepackage{wrapfig}
\usepackage{microtype}
\usepackage{graphicx}
\usepackage{subfigure}
\usepackage{booktabs} % for professional tables
\usepackage{xcolor}
\usepackage{bbm}

\usepackage{mathtools}
\usepackage{verbatim}
\usepackage[ruled,vlined]{algorithm2e}
\usepackage{algorithmic}
\usepackage{enumitem}
\setlist[itemize]{topsep=0pt, leftmargin=3mm}

\newcommand{\tick}{\textcolor{ForestGreen}{\ding{51}}}
\newcommand{\ok}{\textcolor{Dandelion}{\ding{108}}}
\newcommand{\cross}{\textcolor{BrickRed}{\ding{55}}}

\usepackage{tikz}
\usetikzlibrary{intersections,shapes.arrows}

\usepackage{amsthm,amsmath,amssymb,amsfonts}

\DeclareMathOperator*{\argmax}{arg\,max}

\newcommand{\jcom}[1]{\textcolor{red}{JPH: #1}}
\newcommand{\vcom}[1]{\textcolor{blue}{Vu: #1}}
\newcommand{\rr}[1]{\textcolor{green}{RR: #1}}

\newcommand{\bcom}[1]{\textcolor{cyan}{BZ: #1}}
\newcommand{\xcom}[1]{\textcolor{brown}{XS: #1}}

\usepackage{comment}

\usepackage{jair, theapa, rawfonts}

\jairheading{74}{2022}{517-568}{01/2022}{06/2022}
\ShortHeadings{AutoRL: A Survey and Open Problems}{Parker-Holder, Rajan, Song, et al.}
\firstpageno{517}
\begin{document}

\title{Automated Reinforcement Learning (AutoRL):\\ A Survey and Open Problems}

% The \author macro works with any number of authors. There are two commands
% used to separate the names and addresses of multiple authors: \And and \AND.
%
% Using \And between authors leaves it to LaTeX to determine where to break the
% lines. Using \AND forces a line break at that point. So, if LaTeX puts 3 of 4
% authors names on the first line, and the last on the second line, try using
% \AND instead of \And before the third author name.

\begin{comment}
\author{%
\name Jack Parker-Holder$^{*1}$, \name Raghu Rajan$^{*2}$, \name Xingyou Song$^{*3}$,
\\
\name André Biedenkapp$^2$, \name Yingjie Miao$^3$, \name Theresa Eimer$^4$, \name Baohe Zhang$^2$, \name Vu Nguyen$^5$,
\\
\name Roberto Calandra$^6$, \name Aleksandra Faust$^{\dagger3}$, \name Frank Hutter$^{\dagger2,7}$ \& \name Marius Lindauer$^{\dagger4}$
\AND
$^*$ co$-$first authors (alphabetical order) \\
$^\dagger$ co$-$last authors (alphabetical order) \\
$^1$ University of Oxford\\$^2$ University of Freiburg\\ $^3$ Google Research, Brain Team \\ $^4$ Leibniz University Hannover\\  $^5$ Amazon Australia \\ $^6$ Meta AI\\ $^7$ Bosch Center for Artificial Intelligence \\
\texttt{jackph@robots.ox.ac.uk, rajanr@cs.uni-freiburg.de, xingyousong@google.com} \\
}
\end{comment}

\author{\name Jack Parker-Holder$^*$ \email jackph@robots.ox.ac.uk\\
  \addr University of Oxford \vspace{0.23cm}\\
  \name Raghu Rajan$^*$ \email rajanr@cs.uni-freiburg.de\\
  \addr University of Freiburg \vspace{0.23cm}\\
  \name Xingyou Song$^*$ \email xingyousong@google.com\\
  \addr Google Research, Brain Team \vspace{0.23cm}\\
  \name André Biedenkapp \email biedenka@cs.uni-freiburg.de\\
  \addr University of Freiburg \vspace{0.23cm}\\
  \name Yingjie Miao \email yingjiemiao@google.com\\
  \addr Google Research, Brain Team \vspace{0.23cm}\\
  \name Theresa Eimer \email eimer@tnt.uni-hannover.de\\
  \addr Leibniz University Hannover \vspace{0.23cm}\\
  \name Baohe Zhang \email zhangb@cs.uni-freiburg.de\\
  \addr University of Freiburg \vspace{0.23cm}\\
  \name Vu Nguyen \email vutngn@amazon.com\\
  \addr Amazon Australia \vspace{0.23cm}\\
  \name Roberto Calandra \email rcalandra@fb.com\\
  \addr Meta AI \vspace{0.23cm}\\
  \name Aleksandra Faust$^\dagger$ \email sandrafaust@google.com\\
  \addr Google Research, Brain Team \vspace{0.23cm}\\
  \name Frank Hutter$^\dagger$ \email fh@cs.uni-freiburg.de\\
  \addr University of Freiburg \& Bosch Center for Artificial Intelligence \vspace{0.23cm}\\
  \name Marius Lindauer$^\dagger$ \email lindauer@tnt.uni-hannover.de\\
  \addr Leibniz University Hannover \vspace{0.23cm}\\
  \\
  %$^*$ co$-$first authors (alphabetical order) \\
  %$^\dagger$ co$-$last authors (alphabetical order) \\

  % \And
  % Coauthor \\
  % Affiliation \\
  % Address \\
  % \texttt{email} \\
  % \AND
  % Coauthor \\
  % Affiliation \\
  % Address \\
  % \texttt{email} \\
  % \And
  % Coauthor \\
  % Affiliation \\
  % Address \\
  % \texttt{email} \\
  % \And
  % Coauthor \\
  % Affiliation \\
  % Address \\
  % \texttt{email} \\
}
% Uncomment next 3 lines for JMLR
% \ShortHeadings{AutoRL: A Survey and Open Problems}{Parker-Holder, Rajan and Song}
% \firstpageno{1}
% \editor{TODO}

% \ShortHeadings{AutoRL Survey}{1st authors}
% \firstpageno{1}

\maketitle
\vspace{-1cm}
\begin{abstract}%
The combination of Reinforcement Learning (RL) with deep learning has led to a series of impressive feats, with many believing (deep) RL provides a path towards generally capable agents. However, the success of RL agents is often highly sensitive to design choices in the training process, which may require tedious and error-prone manual tuning. This makes it challenging to use RL for new problems and also limits its full potential. In many other areas of machine learning, AutoML has shown that it is possible to automate such design choices, and AutoML has also yielded promising initial results when applied to RL.
However, \emph{Automated Reinforcement Learning} (AutoRL) involves not only standard applications of AutoML but also includes additional challenges unique to RL, that naturally produce a different set of methods.
As such, AutoRL has been emerging as an important area of research in RL, providing promise in a variety of applications from RNA design to playing games, such as Go. Given the diversity of methods and environments considered in RL, much of the research has been conducted in distinct subfields, ranging from meta-learning to evolution.
In this survey, we seek to unify the field of AutoRL, provide a common taxonomy, discuss each area in detail and pose open problems of interest to researchers going forward.

\end{abstract}

% Required for JMLR but not JAIR:
% \begin{keywords}
% Automated Reinforcement Learning, AutoRL, AutoML%, TODO
% \end{keywords}

%\tableofcontents

\section{Introduction}
%Reinforcement learning (RL, \shortciteA{Sutton92}) is a paradigm whereby autonomous agents learn from interacting in an environment to maximize some notion of reward.
In the past decade, we have seen a series of breakthroughs using Reinforcement Learning (RL, \shortcite{sutton_book_2018}) to train agents in a variety of domains, such as games \shortcite{dqn,dota,alphago,alphastar} and robotics \shortcite{dexterous_openai}, with successes emerging in real world applications \shortcite{loon,nguyen2021deep,rl_plasma}. As such, there has been a surge of interest in the research community.

However, while RL has achieved some impressive feats, many of the headline results rely on heavily tuned implementations, which fail to generalize beyond the intended domain. Indeed, RL algorithms have been shown to be incredibly sensitive to hyperparameters and architectures of deep neural networks \shortcite{deeprlmatters,andrychowicz2021what,Engstrom2020Implementation}, while there are a growing number of additional design choices, such as the agent's objective \shortcite{inductivebiases} and update rule \shortcite{learnedPG}. It is tedious, expensive and potentially even error-prone for humans to manually optimize so many design choices at once. There has been significant success in other areas of machine learning (ML) with \emph{Automated Machine Learning} (AutoML, \shortciteA{automl_book}). However, these methods have not yet had a significant impact in RL, in part due to RL applications being typically challenging, due to the diversity of environments and algorithms, and the nonstationarity of the RL problem.

The goal of this survey is to present the field of \emph{Automated Reinforcement Learning} (AutoRL), as a suite of methods automating a varying degree of the RL pipeline. AutoRL serves to tackle a variety of challenges: on the one hand, the fragility of RL algorithms hinders applications in novel fields, especially those where practitioners lack vast resources to search for optimal configurations. In many settings it may be prohibitively expensive to manually find even a moderately strong set of hyperparameters for a completely unseen problem. AutoRL has already been shown to aid in such a situation for important problems, such as designing RNA~\shortcite{runge2018learning}. On the other hand, for those with the benefit of more compute, it is clear that increasing the flexibility of the algorithm can boost performance \shortcite{frodo,stac,pbt}. This has already been exhibited with the famed AlphaGo agent, which was significantly improved through the use of Bayesian Optimization (BO) \shortcite{bo_alphago}.

%This survey seeks to draw connections between a wide variety of methods, developed in a diverse set of communities.
Methods that can be considered AutoRL algorithms were shown to be effective as early as the 1980s \shortcite{barto1981goal}. However, in recent times the prevalence of AutoML has led to the nascent application of more advanced techniques \shortcite{runge2018learning,autorl_learn_reward_nn_arch_chiang_ral19}. Meanwhile the evolutionary community has been evolving neural networks alongside their weights for decades \shortcite{neat}, with methods inspiring those proving effective for modern RL \shortcite{pbt}. In addition, the recent popularity of meta-learning has led to a series of works that seek to automate the RL process \shortcite{evolvedpg,metagradients,learned_objectives_meta_rl_kirsch_iclr20}.

In this paper, we seek to provide a taxonomy of these methods. In doing so, we hope to open up a swathe of future work through the cross-pollination of ideas, while also introducing RL researchers to a suite of techniques to improve the performance of their algorithms. We believe AutoRL has a significant part to play in aiding the potential impact of reinforcement learning, both in open-ended research and practical real-world applications, and this survey could form a starting point for those looking to harness its potential.

Furthermore, we hope to attract researchers interested in AutoML more broadly to the AutoRL community, since AutoRL poses unique challenges. In particular, RL suffers from non-stationarity, as the data the agent is training on is a function of the current policy. In addition, AutoRL also encompasses environment and algorithm design specific to RL problems. We believe these challenges will require significant future work and thus outline open problems throughout this paper.

We structure the paper as follows. In Section \ref{sec:preliminaries}, we describe the background and notation needed to formalize the AutoRL problem and then formalize the problem and discuss various methodologies to evaluate it. We then briefly summarize various types of RL algorithms followed by a description of non-stationarity unique to the AutoRL problem. In Section \ref{sec:what_needs_to_be_automated}, we discuss the various components of the AutoRL problem that need automating including the environments, algorithms, their hyperparameters and architectures. In Section \ref{sec:methods}, we provide a taxonomy and survey current AutoRL methods in subsections following this taxonomy. In Section \ref{sec:benchmarks}, we discuss various publicly available benchmarks and their application domains. Finally, in Section \ref{sec:future_directions}, we discuss future directions for AutoRL.

%, which we discuss in more detail in Section \ref{sec:nonstationarity}.

\begin{table}[h]
\centering
{
\renewcommand{\arraystretch}{1.0}
\label{table:notation}
\scalebox{0.9}{
\begin{tabular}{ll}
\\
Notation & Meaning \\
\hline
$\mathcal{S}$ & State space \\
$\mathcal{A}$ & Action space \\
$P$ & Transition function \\
$R$ & Reward function \\
$\mathcal{O}$ & Observation space \\
$\rho_o$ & Initial state distribution \\
$\mathcal{T}$ & Set of terminal states \\
$\gamma$ & Discount factor \\
$t$ & Inner loop environment timestep \\
$T$ & Maximum inner loop environment timestep \\
$n$ & Outer loop trial index \\
$N$ & Maximum outer loop trial index \\
$V$ & State-value function \\
$Q$ & Action-value function \\
$J$ & Expected total reward function \\
$L$ & Loss function \\
$f$ & Objective function \\
$\theta$ & Neural network parameters \\
$\phi$ & Secondary neural network parameters \\
$B$ & Batch size \\
$\lambda$ & Bootstrapping hyperparameter \\
$\zeta$ & Underspecified components such as hyperparameters or selected algorithms\\
$\eta$ & Subset of hyperparameters that are differentiable \\
\hline
\end{tabular}}
\caption{Table of notation used.}
}
\end{table}

\section{Preliminaries and Notation}\label{sec:preliminaries}

We begin by defining a Markov Decision Process (MDP) as a 7-tuple $(\mathcal{S}, \mathcal{A}, P, R, \rho_o, \mathcal{T}, \gamma)$, where $\mathcal{S}$ is the set of states, $\mathcal{A}$ is the set of actions, $P: \mathcal{S} \times \mathcal{A} \rightarrow \mathcal{S}$ describes the transition dynamics, $R: \mathcal{S} \times \mathcal{A} \times \mathcal{S} \rightarrow \mathbb{R}$ describes the reward dynamics, $\rho_o: \mathcal{S} \rightarrow \mathbb{R}^+$ is the initial state distribution, $\mathcal{T}$ is the set of terminal states and $\gamma \leq 1$ is the discount factor\footnote{$\gamma$ in many instances is treated as a hyperparameter, e.g., to make optimization tractable during training, while evaluation usually reports undiscounted rewards (i.e. $\gamma$ = 1).}.

We further define a POMDP with two additional components: $\mathcal{O}$ represents the set of observations and $\Omega: \mathcal{S} \times \mathcal{A} \times \mathcal{O} \rightarrow \mathbb{R}^+$ describes the probability density function of an observation given a state and action. This makes it an 9-tuple $(\mathcal{S}, \mathcal{A}, P, R, \mathcal{O}, \Omega, \rho_o, \mathcal{T}, \gamma)$.
%To clarify terminology, following \shortciteA{pomdp_info_state_definitions} we will use \textit{information state} to refer to the state representation used by the agent and \textit{belief state} to refer to the posterior belief of the unobserved state given the full observation history.

Note that in many cases this is not sufficient to fully define the task considered (for example, if the dynamics are sampled from a parameterized distribution, such as having a cartpole environment where the length of the pole is drawn from a distribution). Where necessary we will discuss this alongside the methods which require it. Many of the algorithms we discuss are designed to optimize agents to maximize reward in a MDP/POMDP with a fixed parameterization. However, we also discuss those which seek to learn or automate components of the environment itself.

To provide sufficient background information, we introduce the predominant classes of RL algorithms. Deep RL typically considers the problem of finding a \emph{policy} $\pi_{\theta}:\mathcal{S}\rightarrow\mathcal{A}$ parameterized by $\theta$ (i.e. the weights of a neural network) to maximize cumulative reward:
\begin{equation}
  \max_{\theta} J(\theta; \zeta) \> \text{ where } \> J(\theta; \zeta) = \mathbb{E}_{\tau \sim \pi_\theta} \left[\sum_{t \ge 0} \gamma^{t} r_{t}\right],
\end{equation}
where $\tau = (s_{0}, a_{0}, r_{0}, s_{1},...)$ is a trajectory (consisting of observed states, actions and rewards denoted by $s_t$, $a_t$ and $r_t$ at timestep $t$) generated via $\pi_{\theta}$'s interaction with the MDP, and $\sum_{t \ge 0} \gamma^{t} r_{t}$ is the total discounted reward, where the length of $\tau$ is usually either determined by a maximum timestep count $T$, or early termination when the last state is part of the terminal states $\mathcal{T}$. The main focus of this survey is optimizing $\zeta$, which here corresponds to the \emph{underspecified} components of the RL training pipeline. As we will discuss in Section \ref{sec:what_needs_to_be_automated}, $\zeta$ may refer to anything from a single hyperparameter to an entire algorithm, which may not be known in advance.

To evaluate the success of our RL training procedure, we typically have a separate \textit{evaluation objective} $f(\zeta, \theta)$ that refers to the performance of our agent $\pi_\theta$ in a given distribution of MDPs.\footnote{Note that the distribution of MDPs might be just a single MDP depending on what the practitioner's goals are.} Thus, the \textbf{AutoRL problem} may be framed in terms of a general bi-level optimization, i.e.
\begin{equation}
\max_{\zeta} f(\zeta, \theta^{*}) \> \> \> \text{s.t. } \> \theta^{*} \in \argmax_{\theta} J(\theta; \zeta),
\end{equation}\label{bi-level}%
%\todo[inline]{Marius: in the automl literature, $\lambda$ is more common than $\zeta$. I saw that you later used $\lambda$ for TD-$\lambda$, which makes totally sense. Nevertheless, I would like to avoid to introduce a new notation for AutoRL. Maybe we could use $\vec{\lambda}$ or $\boldsymbol{\lambda}$ for a hyperparameter vector?}
%\todo[inline]{Richard: $\zeta$ is used in DARTS notation - IMO $\vec{\lambda}$ would make it more confusing; I'd rather stick with $\zeta$ unless there's a huge issue}
%\todo[inline]{Marius: I see your point. Please also consider that $\zeta$ is used for the learning rate; so some readers might be confused and believe that we only optimize step size. So, I would still vote for $\boldsymbol{\lambda}$}
%\todo[inline]{Richard: ([EDITED]: Just realized doc has changed, there's actually no specific learning rate variable) Just wondering, what is the main purpose of using the \textit{vec} command? Is it only because $\lambda$ will be overloaded, or is it trying to convey that the hyperparameters are a vector? If the latter, would $\vec{\zeta}$ make more sense? The tradeoff there might be that $\nabla_{\vec{\zeta}}$ looks a little weird.}
\noindent where $\max_{\zeta} f(\zeta, \theta^{*})$ can be considered the \textit{outer loop}, while $\max_{\theta} J(\theta; \zeta)$ can be considered the \textit{inner loop}, which may be equivalent to minimizing some \textit{loss}, i.e. $\min_{\theta} \mathcal{L}(\theta; \zeta)$, for some methods requiring automatic differentiation. The maximization procedure usually is constrained to a \textit{budget}, with the most common being a maximum trial count $N$ of allowed evaluations $f(\zeta_{1}, \theta^{*}_{1}),..., f(\zeta_{N}, \theta^{*}_{N})$. The notation used in the paper is summarized in Table \ref{table:notation}. \textbf{Optimizing $\zeta$ will be the main purpose of this survey}, with common definitions of $\zeta$ outlined in Section \ref{sec:what_needs_to_be_automated}, while optimization methods will be outlined in Section \ref{sec:methods}.

\subsection{Evaluation Methods} \label{subsec:eval_methods}
The most common inner-loop evaluation objective $f$ considered is simply the original cumulative reward, i.e. $f(\zeta, \theta^{*}) = J(\theta^{*}; \zeta)$, measured after a fixed number of timesteps have been used from the training environments. However, this assumes that the evaluation MDP is specified at training time, which may not always be the case (e.g. in Sim2Real problems encountered in real world robotics), and sometimes it may be desirable, for example, to train on a different distribution of tasks to improve performance on the target task (see Section \ref{sec:envdesign}) or even a distribution of tasks. Performance under distributional shifts can also be explicitly described using metrics from the recently emerging field of RL \textit{generalization} \shortcite{overfitting_rl,rl_generalization_survey}, where $f(\zeta, \theta^{*})$ can be defined as the \textit{validation} reward, i.e. the reward in the outer loop, whereas $J(\theta^{*};\zeta)$ can be considered the \textit{training} reward, i.e. the reward in the inner loop. In other cases, rather than optimizing purely reward-based metrics, properties such as efficiency can be optimized, where $f(\zeta, \theta^{*})$ may be defined with respect to the policy's inference speed and number of floating point operations. Optimizing multiple objectives in such a manner comes under the scope of Multi-Objective Optimization (MOO) \shortcite{moml_book_jin_2006}.\footnote{Such
%,deep_modl_ruchte_icdm_2021
Multi-Objective Optimization should not be confused with Multi-Objective Reinforcement Learning (MORL) \shortcite{morl_yang_neurips_19,morl_moffaert_jmlr_15}. The former refers to multiple objectives in the outer loop while the latter refers to a vector reward signal, e.g. $\overrightarrow{\text{\bf{R}}}$, in the MDP under consideration.}

Furthermore, there are various ways to evaluate the performance of the outer loop as well, many of which are used in the AutoML literature. By far the most common evaluation metric reported is the best value found so far, i.e. $\max \{f(\zeta_{1}, \theta^{*}_{1}),..., f(\zeta_{N}, \theta^{*}_{N})\}$, when there is a notion of a maximum budget of trials $N$, i.e., the hyperparameters are set $N$ times. This budget can also be defined using multiple fidelities \shortcite{cutler2014multifidelity_rl,bayesopt_multifidelity,multifidelity_bo}, such as the sum of timesteps used over the inner loops or total wallclock time. Wallclock time is especially useful as a metric when the outer-loop optimization may not be based on a multi-trial method, but rather a differentiable optimization process, e.g. DARTS \shortcite{darts}. Furthermore, other metrics, such as cumulative regret \shortcite{gp_ucb}, defined as $\sum_{n=1}^{N} \left[ f(\zeta^{*}, \theta^{*}_{max}) - f(\zeta_{n}, \theta^{*}_{n}) \right]$ or trial count needed to reach a certain performance threshold for $f$ \shortcite{expected_time_for_performance}, may also be used for evaluating the outer loop.

A unique challenge in regards to evaluation in AutoRL arises due to natural instabilities in training, which strongly affects the optimation procedure over $\zeta$. In RL, the result of optimizing $J(\theta^{*}, \zeta)$ can vary widely per training run and as a result some of the metrics mentioned above, such as the cumulative reward, cannot be robustly measured with just a single training run. As such, it is quite common in RL to occasionally have even the perfect training setup completely fail. Normally in RL, there is a common standard of reporting aggregate results from multiple training runs or \textit{seeds} \shortcite{deeprlmatters,how_many_rl_seeds}, usually involving the mean or median of rewards, in order to resolve this issue. However, even such metrics can be very unreliable, especially for ranking different optimization methods \shortcite{rliable} where the distribution of outcomes can be incredibly wide. \shortciteA{rliable} propose to correct this by using metrics based on robust statistics, which include the interquartile mean and the optimality gap. Such issues and improvements in robust evaluation are very important in ensuring that the outer-loop optimization remains effective; we discuss this matter throughout Section \ref{sec:methods}.

\subsection{Inner Loop Optimization}
We now provide a brief background on popular methods of optimizing the inner loop, i.e., $\theta$. There are several classes of methods used to optimize $\theta$ \shortcite{sutton_book_2018}. At a high level, such methods may be classified into \textit{model-free} and \textit{model-based} methods depending on whether they use a model of the environment or not.

\paragraph{Model-free methods:} A common class of model-free methods is \textit{policy gradients} \shortcite{policy_gradients}, which seeks to optimize $J(\theta)$ via gradient descent using the fact that
\begin{equation}
\nabla_{\theta} J(\theta) = \mathbb{E}_{\tau \sim \pi_{\theta}} \left[\sum_{t \ge 0} r_{t} \nabla_{\theta} \log \pi_{\theta}(a_{t} | s_{t}) \right],
\end{equation}
assuming the notion of an action distribution from the policy: $a \sim \pi_{\theta}(\cdot | s)$. Policy gradient algorithms have been used extensively in robotics \shortcite{sac-v2,tan2018sim}, with several large scale applications \shortcite{dota}.

In contrast, \emph{value-based} methods seek to learn an action-value function predicting the expected discounted cumulative reward from a given state if an action is taken \shortcite{watkins1992q} and policy $\pi_{\theta}$ is followed thereafter:
\begin{equation}
    Q_{\theta}(s, a) = \mathbb{E}_{\tau \sim \pi_{\theta}} \left\lbrack \sum_{t \geq 0} r_t \middle| s_0 = s, a_0 = a \right\rbrack \,,
\end{equation}
for each initial state-action pair $(s, a) \in \mathcal{S} \times \mathcal{A}$.  The policy therefore seeks to take the action maximizing the so-called \emph{Q-values}, i.e. $\pi_{\theta}(s) \in \argmax_a Q_{\theta}(s, a)$. A prototypical method based on temporal difference learning \shortcite{sutton_book_2018} is Q-learning \shortcite{watkins1992q};
in its most basic form, it maintains an estimate \mbox{$Q_{\theta} : \mathcal{S} \times \mathcal{A} \rightarrow \mathbb{R}$} of the optimal value function, and given a sequence of transition tuples $(s_t, a_t, r_t, s_{t + 1})_{t \geq 0}$, updates $Q_{\theta}(s_t, a_t)$ towards the target $r_t + \gamma \max_{a \in \mathcal{A}} Q_{\theta}(s_{t + 1}, a)$, for each $t \geq 0$. The canonical value-based method in deep RL is DQN \shortcite{dqn}. Since then, there has been an explosion of interest in the field, and a swathe of improvements to this algorithm \shortcite{doubleQ,prioritized_exprep,dueling,distributional_og,fortunato2018noisy}. These innovations were combined in the Rainbow algorithm \shortcite{rainbow}.

\textit{Actor-critic} methods (\shortciteA{sutton_book_2018}) usually combine policy gradient and value-based methods. In this case, the actor represents the policy which is learned using policy gradients based on the value function learnt by the critic \cite<see e.g.,>{ddpg,a3c_mnih_icml_16}.

The model-free methods mentioned above can be practically organized into three components: a learner, a replay buffer, and potentially multiple actors. The \textit{learner} performs gradient descent optimization over the \textit{replay buffer}, which contains trajectory data collected by \textit{actors} interacting with the MDP.
%Many of the best known results in RL stem from these two classes of methods, but there have also been successes for other approaches. For example,

%However, a final class of algorithms avoids the notion of time-step specific data altogether, and instead simply apply blackbox optimization over the input space $\theta$ to optimize the total reward $\sum_{t \ge 0} \gamma^{t} r_{t}$. These are considered Evolutionary Strategies (ES) \textcolor{red}{...}

\paragraph{Model-based methods:} Alternatively, \emph{Model-Based} Reinforcement Learning (MBRL) methods make use of models of the environment, such as the transition function $P$ and the reward function $R$. This may be in the form of the \textit{true} model provided to the agent, famously used by the AlphaGo \shortcite{alphago} algorithm, or the model may also be \textit{learned} during training to predict the environment dynamics, either for training a policy \shortcite{dyna,dreamer,worldmodels,simple} or for planning \shortcite{planet,pets,tassa_ilqg_iros_12,schrittwieser2020mastering}. In addition to minimizing the losses involved for model-free methods above, this may entail further minimizing, e.g., a mean-squared error (MSE) loss or negative log likelihood (NLL) loss on learning $P$ and $R$.

Planning-based approaches led to one of the most prominent successes in RL, with the AlphaGo agent making use of Monte Carlo Tree Search (MCTS, \shortciteA{mcts_browne_ieee_12}) guided by neural network approximators for the policy and value functions. MCTS performs a heuristic search by randomly sampling rollouts for a given state. This requires the true model of the environment. MuZero \shortcite{schrittwieser2020mastering} takes these ideas one step further. It matches the performance of AlphaZero on Go, chess and shogi, while also performing well on Atari games with a learned model. An alternative planning approach is Model Predictive Control (MPC, \shortciteA{mpc_garcia_journals_automatica_89}), which solves an optimization problem online. This has been used in Deep RL, for example, in the PETS algorithm  \shortcite{pets} or PlaNet \shortcite{planet}.

In contrast, Dyna \shortcite{dyna} is a canonical MBRL algorithm that essentially treats a learned dynamics model as an environment for model-free RL. Dyna-style approaches are popular since they facilitate faster inference at deployment time (forward-pass of the policy rather than an optimization loop) and have been shown to be highly sample-efficient both from proprioceptive states \shortcite{mbpo} and pixels \shortcite{dreamer}.

% For a more detailed treatment of MBRL we would refer the reader to \shortciteA{mbrl_survey}.

\subsection{Non-Stationarity of the Optimization Problem}
\label{sec:nonstationarity}
In various domains of AI, such as evolutionary algorithms or deep learning, it is well known that some hyperparameters need to be adapted over time to achieve the best performance \cite<see e.g.,>{moulines-neurips11,doerr-toec20}.
A prime example for this are learning rates in deep learning.
Keeping static learning rates might result in very slow learning or even in diverging learning behavior.
As deep neural networks are the commonly used method of function approximation in the current landscape of RL, it is to be expected that similar dynamic treatments of hyperparameters are needed in RL.
In fact, existing optimization experiments have observed that dynamically adapting learning rates is also beneficial for model-free and model-based RL \shortcite{pbt,pbtbt}.

Additionally, the nature of RL further amplifies the potential non-stationarity of hyperparameters, even in stationary environments \shortcite{igl2021transient}.
With the trial-and-error approach to learning, an RL agent constantly (re-)generates its own training data, which is highly dependent on the current instantiation of the agent.
Thus, hyperparameter settings might require very different values over the whole training run.
Indeed, besides classical static optimization approaches, methods for AutoRL typically include approaches for dynamically selecting configurations. We will discuss these more in Section~\ref{sec:methods} when we introduce the current state of methods for AutoRL.

%\section{What Needs to be Optimized?}
%\xcom{Maybe this is also worth organizing? Raw Reward, Efficiency, Generalization, Sample Efficiency, etc.}
%\jcom{this should be included in the other section}

\section{What Needs to be Automated?}
\label{sec:what_needs_to_be_automated}
\begin{figure}[h]
    \center
    \includegraphics[keepaspectratio, width=0.8\textwidth]{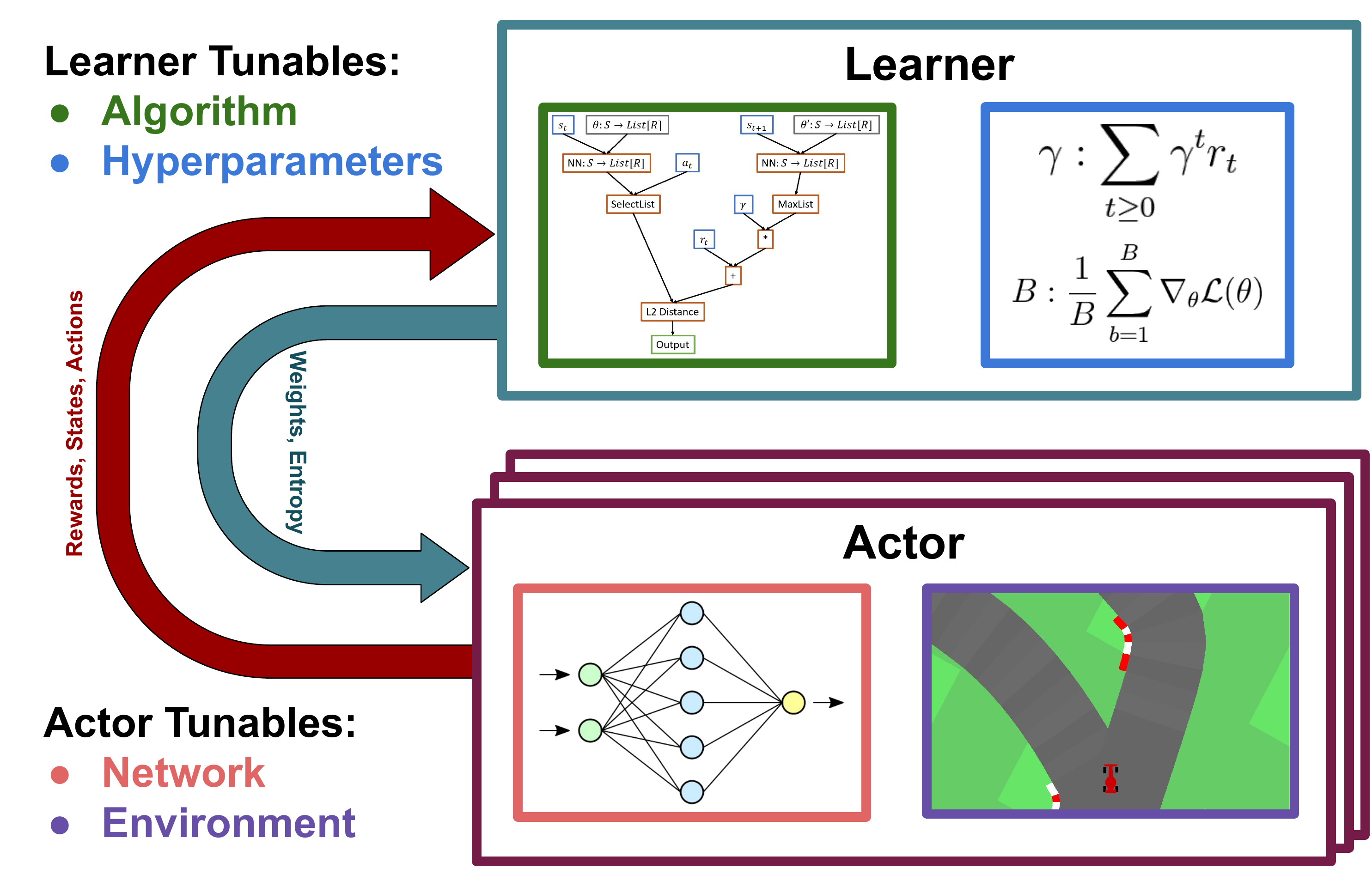}
\caption{Components, or definitions of $
\zeta$, that have been commonly tuned in AutoRL approaches. Most pipelines consist of a learner and actors. The learner possesses tunable algorithms (e.g. Q-learning computation graph by \shortciteA{evolving_algos}) and hyperparameters (e.g.: discount factor $\gamma$ and batch size $B$). The actors possess the network with weights, and the environment (e.g. Car-Racing by \shortciteA{gym}).}
\label{fig:tunable_components}
\end{figure}

In this section, we introduce the set of problems AutoRL seeks to solve. Figure \ref{fig:tunable_components} demonstrates some of the tunable components (i.e. possible definitions of $\zeta$) in a standard RL pipeline, which range from entire algorithms to a single hyperparameter. The following subsections begin at a high level, approximately equating to the sequence of decisions that need to be made when first considering using RL to solve a problem.

\subsection{How Do We Design Tasks?}

When using RL for a new problem setting, designing the learning environment is usually the first problem to be considered. First of all, the user must specify a reward function $R(s, a, s')$. This is typically assumed to be provided in many RL papers and common benchmarks \shortcite{ale,gym}, yet it is certainly non-trivial to be set for many real world problems. For example, for robotic manipulation, the reward could be provided once the task is completed, or given based on some intermediate goal, such as reaching the object of interest. These differences can often have a dramatic impact on the success of policy optimization \shortcite{gleave2021quantifying}.

Often, it is also necessary to determine what should be provided as an observation for the agent. For example, when training from pixels, it is common to deploy a series of preprocessing steps that may be customized per environment. Indeed, one of the very first successes of deep RL made use of greyscale to simplify the learning problem and to stack consecutive frames to form the observation \shortcite{dqn}. While such additional processing is not always needed, using raw frames directly can be demanding in terms of computation and memory requirements \shortcite{dqn}. In addition, a large action space makes it significantly harder for an agent to begin learning; so it is often required for researchers to pick the available actions for an agent, or consider learning them \shortcite{farquhar2020growing}.

Finally, the environments to train on can also be controlled by the designer. We discuss a wide variety of automated environment design methods in Section \ref{sec:envdesign}.

\subsection{Which Algorithm Do We Use?}

Once we are able to define an MDP (or POMDP) to solve, the next question is the choice of algorithm. In cases where RL has already demonstrated strong performance, the choice may seem rather simple. For example, when playing a new Atari game, a variant of DQN can be expected to perform well. For a continuous control task from MuJoCo \shortcite{todorov2012mujoco}, an actor critic algorithm such as SAC \shortcite{sac-v2} would likely be a strong baseline. However, even for these standard benchmarks, RL agents are known to be brittle \shortcite{deeprlmatters,andrychowicz2021what,Engstrom2020Implementation} and this choice is not as simple as it seems. For a completely new problem, the challenge is significantly greater, which prevents the uptake of RL for real world problems. Without expertise in the field or vast computational resources, many use cases may result in sub-optimal solutions as arbitrary algorithms are used based on success in somewhat unrelated problems.

To tackle the problem of which algorithm to use, there are multiple possible avenues. One could use a learning to learn approach \shortcite{l2lbGDbGD_andrychowicz_neurips_16,l2lwGDbGD_chen_icml_17} where a meta-learner learns an agent that can perform well on a set of related tasks. The agent would be trained over a distribution of tasks and learn to recognize the task during rollouts at test time. Such recognition capabilities can be achieved by a Recurrent Neural Network (RNN), for instance, that rolls out a policy depending on its state that would encode the task at hand.

One could also learn the objective function itself, where each objective function defines a new algorithm. In the most general form, RL's objective is to maximize expected cumulative reward $J(\theta; \zeta)$ in a given environment. In the context of deep RL, this objective is not differentiable with respect to model parameters; therefore proxy objectives are used in practice, which have a strong impact on the learning dynamics. Indeed, many advancements in deep RL result from improvements in the objective functions, such as double Q-learning \shortcite{doubleQ}, distributional value functions \shortcite{distributional_og} and yet more \shortcite{ddpg,schulman2017proximal,sac-v2}. These algorithms have fundamental differences in the objective functions, which are designed by human experts. Even smaller tweaks to known objective functions can make a big difference, e.g. Munchausen-DQN \shortcite{m-dqn} and CQL \shortcite{cql}, yet these tweaks require significant theoretical analysis and empirical experiments to validate their effectiveness.

One could also make a categorical choice between existing algorithms based on multiple evaluations. In AutoML, such a choice of algorithm can be automated through algorithm selection \shortcite{rice1976algorithm}. In algorithm selection, a meta-learning model is used to decide which algorithm to use for an environment at hand. The so-called selector is trained based on past observed performances and features of the environment. While this alleviates the need of expert knowledge, it requires potentially many resources to gather enough performance data to learn a well-performing selector.

We discuss works related to learning algorithms in RL in Section~\ref{sec:discovering_algs}.

\subsection{What about Architectures?}
Many of the large breakthroughs in machine learning have come through the introduction of new architectures \shortcite{alexnet,resnet,vaswanitransformers}. To automate this discovery, the field of \emph{Neural Architecture Search} (NAS) \shortcite{ElskenMH19}
%,nas_best_practices_jmlr_19}
has become an active area of research over the past few years. In contrast to supervised learning, very little attention has been paid to the design of neural architectures in RL. For tasks from proprioceptive states, it is common to use two or three hidden layer feedforward MLPs, while many works learning from pixels still make use of the convolutional neural network (CNN) architecture used in the original DQN paper, known as ``Nature CNN''. More recently, it has become commonplace to make use of the network proposed in the IMPALA paper \shortcite{impala}, which is now referred to as ``IMPALA-CNN''. While the IMPALA-CNN has been shown to be a stronger architecture for vision and generalization \shortcite{procgen,coinrun}, there has been little research into alternatives, although some have focused on using attention modules for policies \shortcite{transformersforrl,attentionagent2020,relational_rl}.

Along with IMPALA-CNN, there is evidence that deeper and denser networks, use of different nonlinearities, as well as normalizers, such as Batch Normalization, can improve current methods across a suite of manipulation and locomotion tasks \shortcite{impact_nn_arch_rl_sinha_neurips20_deep_rl_workshop,observational_overfitting}, %\todo{Check how to have both arxiv and workshop info in the reference entry}
even for the MLP setting. \shortciteA{kumar2020implicit} further expanded upon the downside of underparameterization for value-based methods. In general, there remains little conceptual understanding (and uptake) on architectural design choices and their benefits in the RL domain. While \shortciteA{procgen,coinrun} have shown \textit{overparameterization} and batch normalization \shortcite{batch_norm} effects in RL generalization, it is unclear whether they can be explained by supervised learning theories, i.e. \textit{implicit regularization} \shortcite{implicit_regularization,practical_overparameterization_neyshabur15}, Neural Tangent Kernels \shortcite{ntk,cntk}, complexity measures \shortcite{width_complexity_measure_neyshabur19} and landscape smoothness \shortcite{batch_norm_smoothness}.

We discuss works seeking to address neural architectures in RL in a variety of sections, given that it can be addressed using many different approaches.

\subsection{Last but not Least: What about Hyperparameters?}
Having defined the task and selected (or learned) an algorithm and architecture, the last remaining challenge is selecting hyperparameters.\footnote{We note, however, that to search for neural architectures one should already have meaningful hyperparameters. This poses a chicken-and-egg problem, which can be side-stepped by jointly searching for neural architectures and hyperparameters as, for example, done by \shortciteA{runge2018learning}.} The most widely studied area of AutoRL is the sensitivity of RL algorithms to hyperparameters. Indeed, in one of the best-known studies,  \shortciteA{deeprlmatters} found that many of the most successful recent algorithms were brittle with respect to hyperparameters, implementation, or even seed, while \shortciteA{reproducibility_cont_control_islam_arxiv17} noted that it is challenging to compare benchmark algorithms from different papers given the impact of hyperparameters on performance.

One of the better understood hyperparameters is the discount factor $\gamma$, which determines the time-scale of the return. \shortciteA{adaptivecriticdesigns}, as well as \shortciteA{neurodynamic}, found that lower discount factors led to faster convergence, with the potential risk of leading to myopic policies. \shortciteA{singhdayan} explored the manner in which TD learning is sensitive to the choice of its step size and eligibility trace parameters.

%\todo{Marius: Inconsistent style of citing: (a) cite\{XYZ\} did ..., (b) XYZ et al. cite\{XYZ\} did ..., (c) approach cite\{XYZ\}, (d) the authors of cite\{XYZ\} .... Please make it consistent. I (and I think also Frank) would be in strong favor of variant (b).\\
%Frank: I agree. Also, while the citation style in 3.4 is now consistent, it is different than 3.1-3.3 (which never give author names) ... both are possible, I would just argue for consistency.}

Another important hyperparameter, which could be possibly considered as choosing the whole algorithm itself, is the \textit{exploration-exploitation trade-off} in RL. This trade-off is a significant determinant of performance and has been explored in several papers \shortcite{agent57,dabney_temporally_extended_e_greedy_iclr_21,temporl}. At a higher level, the outer loop algorithm must also consider a similar trade-off, which remains a key component of research in methods such such as Bayesian optimization and Population-Based Training %\shortcite{pbt}
and deserves special attention for an AutoRL pipeline.

Looking at specific algorithms, \shortciteA{andrychowicz2021what} conducted an extensive investigation into design choices for on-policy actor critic algorithms. They found significant differences in performance across loss functions, architectures and even initialization schemes, with significant dependence between these choices. \shortciteA{obando20revisiting} also explored design choices for off-policy RL, highlighting the differing performance for MSE vs. Huber loss functions, while also assessing the importance of n-step returns, which \shortciteA{Fedus2020RevisitingFO} and \shortciteA{rowland2020adaptive} also studied. \shortciteA{logistic-q} showed that the performance improved by ensuring the convexity in the Q-learning using logistic Bellmann error function. Furthermore, \shortciteA{liu2021regularization} showed that the choice of regularizer can also significantly impact performance.

Finally, beyond the well-flagged hyperparameters, there are even significant \emph{code level} implementation details. \shortciteA{deeprlmatters} identified this and showed that different codebases for the \emph{same algorithm} produced markedly different results. Furthermore, \shortciteA{Engstrom2020Implementation} investigated implementation details in popular policy gradient algorithms (PPO, \shortciteA{schulman2017proximal} and TRPO \shortciteA{schulman2015trust}), where details such as reward normalization were found to play a crucial role in RL performance.
%This section will be discussing results which show AutoRL is important.
With this area being the most studied component of AutoRL, we discuss various different solution approaches to the hyperparameter optimization problem in Sections \ref{sec:rs} to \ref{sec:blackbox_online_tuning}.
%\ref{sec:generic}, \ref{sec:neuroevolution}, \ref{sec:metagradients} and \ref{sec:blackbox_online_tuning}.

\section{Methods for Automating Reinforcement Learning}
\label{sec:methods}

\begin{figure}[h]
    \center
    \includegraphics[keepaspectratio, width=0.99\textwidth]{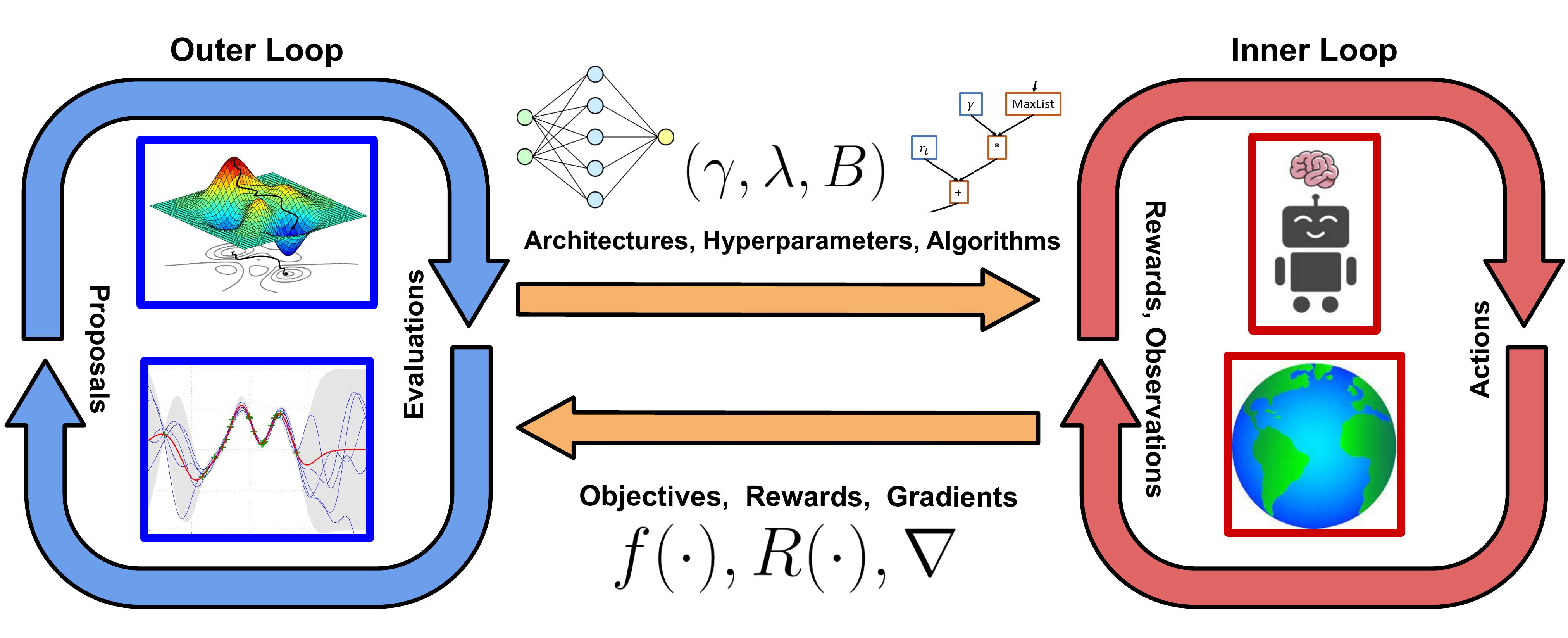}
\caption{Most approaches can be organized in terms of an outer loop and inner loop, where the outer loop sends specifications, such as architecture specification, hyperparameters, and algorithms, while the inner loop sends back objectives, rewards, and gradients. Images from \shortcite{gp_picture,evolving_algos,optimization_picture}.}
\label{fig:autorl_pipeline}
\end{figure}

In this section, we survey methods for AutoRL, spanning a wide range of communities and encompassing a broad array of techniques. In general, most methods can be conveniently organized in terms of a combination of an \textit{inner loop}, which consists of a standard RL pipeline, and an \textit{outer loop}, which optimizes the agent configuration. Each loop can be optimized via either blackbox optimization or gradient-based methods, although the combination of gradients for the outer loop and blackbox for the inner loop blackbox is not possible, as the inner loop blackbox setting would make gradients unavailable. This is shown in Figure \ref{fig:autorl_pipeline} and Table \ref{table:gradient_vs_blackbox}.

\begin{table}[h]
\begin{center}
\scalebox{0.94}{
\begin{tabular}{ ccc }
    & Benefits & Downsides \\
 \hline
 Blackbox & Flexible/No assumptions on pipeline & Can be sample-inefficient \\
 Gradient & Scalable and sample-efficient & Inflexible/Requires differentiable pipelines \\
 \hline
\end{tabular}
}
\caption{Comparison between gradient-based and blackbox approaches.}
\label{table:gradient_vs_blackbox}
\end{center}
\end{table}

We summarize the taxonomy of AutoRL methods along broad classes (which often overlap) in Table \ref{table:autorl_summary}. Each has its own pros and cons, as there are inevitably trade-offs, such as sample vs. compute efficiency. The scope for each method is also vastly different, from tuning a single hyperparameter, to learning a reward function, or an entire algorithm from scratch. In each subsection we first discuss related methods before proposing open problems.

\begin{table}[h]
{
\vspace{-5mm}
\renewcommand{\arraystretch}{1.16}
\caption{A high level summary of each class of AutoRL algorithms. In each case, the properties are a generalization, and specific algorithms in each class may vary.}
\label{table:autorl_summary}
\scalebox{0.88}{
\begin{tabular}{lllllll}
\\
Class & \multicolumn{5}{l}{Algorithm properties} & What is automated? \\
\hline
Random/Grid Search (\ref{sec:rs}) & \Ankh\Ankh \Ankh & $\blacksquare$ & $\rightrightarrows$ & \tick & $\doteq$ & hyperparameters, architecture, algorithm \\
Bayesian Optimization (\ref{sec:bo}) & \Ankh\Ankh \Ankh & $\blacksquare$ & $\rightrightarrows$ & \tick & $\doteq$ & hyperparameters, architecture, algorithm \\
Evolutionary Approaches (\ref{sec:neuroevolution}) & \Ankh\Ankh\Ankh & $\blacksquare$ & $\rightrightarrows$ & \tick & $\approx$ & hyperparameters, architecture, algorithm \\
% Population Based Training (\ref{sec:pbt}) &  \Ankh\Ankh \Ankh & $\blacksquare$ & $\rightrightarrows$ & \tick  & hyperparameters, architecture \\
Meta-Gradients (\ref{sec:metagradients}) & \Ankh & $\nabla$ & $\rightarrow$ & \ok & $\approx$ & hyperparameters \\
Blackbox Online Tuning (\ref{sec:blackbox_online_tuning}) &  \Ankh & $\blacksquare$ & $\rightarrow$ & \tick & $\approx$ & hyperparameters \\
Learning Algorithms (\ref{sec:discovering_algs}) &  \Ankh \Ankh \Ankh & $\blacksquare$ & $\rightrightarrows$ & \ok & $\doteq$ & algorithm \\
Environment Design (\ref{sec:envdesign}) &  \Ankh \Ankh \Ankh & $\blacksquare$ & $\rightrightarrows$ & \ok & $\approx$ & environment \\
\hline
\multicolumn{7}{l}{
\begin{small}
\Ankh \ only uses a single trial, \Ankh \Ankh \Ankh \ requires multiple trials
\end{small}} \\
\multicolumn{7}{l}{
\begin{small}
$\nabla$ requires differentiable variables, $\blacksquare$ works with non-differentiable hyperparameters
\end{small}} \\
\multicolumn{7}{l}{
\begin{small}
$\rightrightarrows$ parallelizable $\rightarrow$ not parallelizable
\end{small}} \\
\multicolumn{7}{l}{
\begin{small}
\tick \ works for any RL algorithm, \ok \ works for only some classes of RL algorithms %\cross \ requires a specific setup.
\end{small}} \\
\multicolumn{7}{l}{
\begin{small}
$\doteq$ static optimization, $\approx$ dynamic optimization
\end{small}}
\end{tabular}}
}
\end{table}

% \subsection{Generic Approaches}
% \label{sec:generic}
% We begin by discussing generic approaches to optimization that are commonly applied in different communities.
% We however split this section in sequential and parallel approaches to allow for a more fine-grained distinction of the methods presented here.

\subsection{Random/Grid Search Driven Approaches}
\label{sec:rs}
%\todo{Frank: Why do we not start with these methods? They are clearly the simplest methods. And as it stands, we already say in the preceding subsection on BO that BO is better than Hyperband without having introduced Hyperband. Introducting Hyperband would also be good before introducing BOHB in the BO section. Raghu: Moved this sub-section to before BO methods.}

%Random Search, Grid search, Hyperband et al.

 %However, they are both typically inferior to existing batch (or parallel) BO approaches \shortcite{CoCaBO,kandasamy2020tuning,turner2021bayesian}. This is because there is no intelligent way for Random Search and Grid Search to leverage historical observations to make informed decisions.

%Existing batch BO approaches \shortcite{gp_bucb,bayeslocalpen} are mostly greedy by sequentially visiting all the maxima of the acquisition function. After the first maximum is found, such methods adjust the acquisition function by suppressing the current maximum and then move on to find the next best maximum. This is repeated till a batch of maxima is collected. Alternative to such greedy techniques, we can collect a batch of points directly by (i) exploiting the property of Thompson sampling in drawing independent samples \shortcite{hernandez2017parallel,kandasamy2018parallelised} or (ii) approximating the acquisition function as a Gaussian mixture model and selecting the Gaussian means \shortcite{nguyen2016budgeted}.

We begin by discussing the most simple methods: Random Search \shortcite{randomsearch} and Grid Search \cite{automl_book}. As the names imply, Random Search randomly samples hyperparameter configurations from the search space and Grid Search divides the search space into fixed grid points which are evaluated. Due to their simplicity, Random Search and Grid Search can be used to select a list of hyperparameters\footnote{As mentioned in Table \ref{table:autorl_summary}, these approaches are applicable to architectures and algorithms as well, but they are usually only applied to hyperparameters. They are applied to architectures in the form of selecting, e.g., number of layers of the network or the width of a layer, but we include such basic architectural choices under hyperparameters.}, exhaustively evaluate all of them and pick the best configuration, making them very general approaches. Indeed, Grid Search is still the most commonly used method in RL, featuring in the vast majority of cases where hyperparameters are tuned, but should not be considered as the most efficient approach.

\begin{figure}[h]
    \center
    \subfigure[Grid Search at $n=0$, $n=N$ and averaged over $n$.]{
        \includegraphics[width=0.9\textwidth]{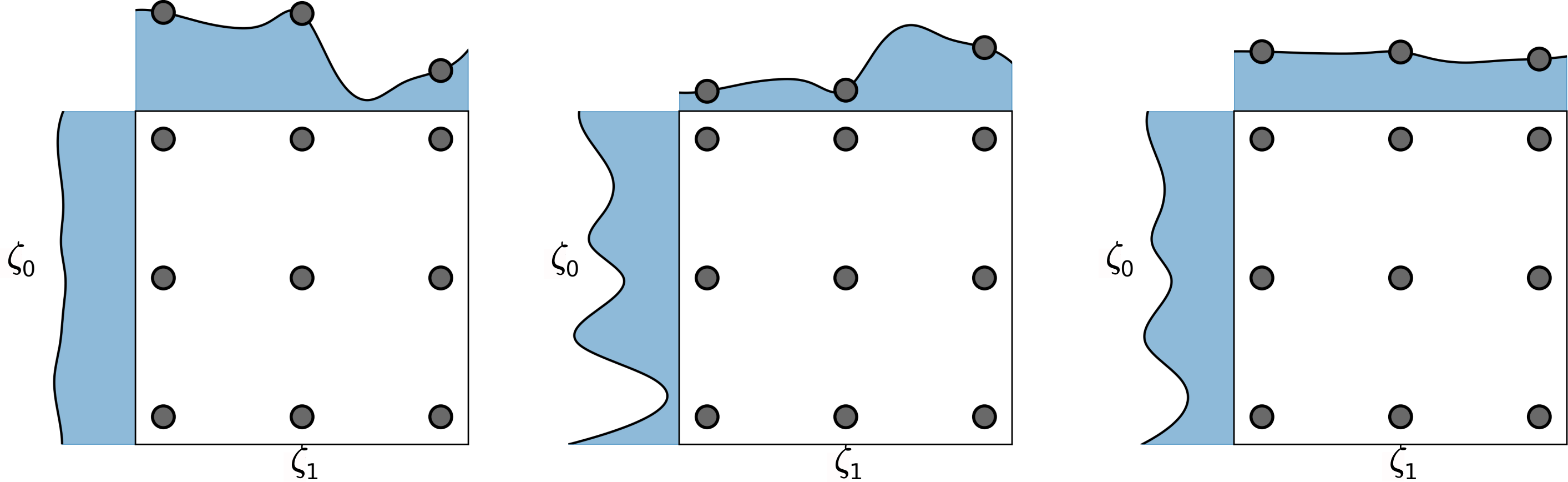}\label{fig:gs}
    }

    \subfigure[Random Search at $n=0$, $n=N$ and averaged over $n$.]{
        \includegraphics[width=0.9\textwidth]{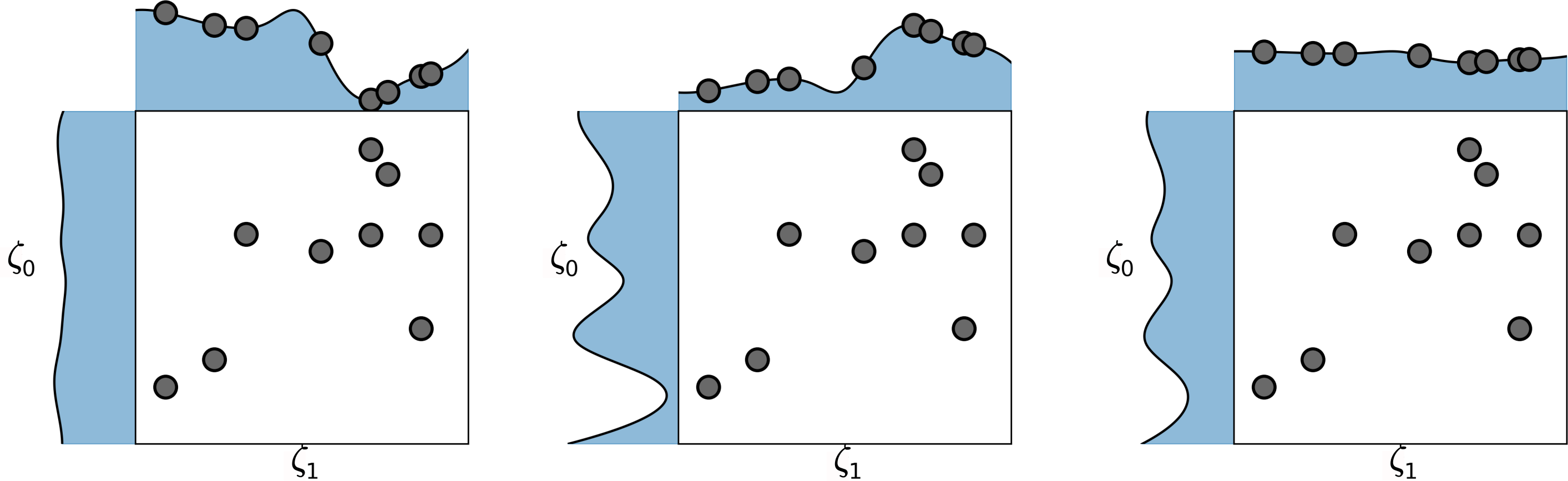}\label{fig:rs}
    }
\caption{The non-stationarity of the AutoRL problem poses new challenges for classical (hyperparameter) optimization methods. Generally, grid and random search keep the hyperparameters $\zeta_0$ and $\zeta_1$ fixed throughout a run. If the loss landscape changes during the run (see difference between leftmost and middle images) these methods optimize the average performance over time (right most images). Circles represent hyperparameters that are sampled at the beginning of the optimization procedure and the blue shaded area depicts the RL agent's loss. This figure is based on a figure for the stationary case by~\shortciteA{randomsearch}.}%~\shortciteA{automl_book}.}
\label{fig:gsVSrs}
\end{figure}

These classical approaches do not take the potential non-stationarity of the optimization problem (recall Section~\ref{sec:nonstationarity}) into account. We describe the issue below, using Figure~\ref{fig:gsVSrs}. Assume we use these procedures to \textit{minimize} the loss of an RL algorithm that has two continuous hyperparameters $\zeta_0$ and $\zeta_1$ (see Figure~\ref{fig:gsVSrs}).
We evaluate $9$ hyperparameter settings in parallel. At $n=0$ (First Column of Figure~\ref{fig:gsVSrs}), we observe the performance of all settings as indicated by the blue shaded areas.
It is clear that changes in $\zeta_0$ result in little difference in the agents performance, whereas changes in $\zeta_1$ result in large performance differences. While random search is likely to observe the performance of $9$ unique values per hyperparameter, grid search only observes $3$. At $n=N$ (Middle Column of Figure~\ref{fig:gsVSrs}), we observe that the loss landscape has shifted and that now both hyperparameters have an impact on the final performance. Furthermore, while in the early phase of the optimization procedure, large values of $\zeta_1$ were beneficial, now smaller values are preferable.
% With both static variants of random and grid search we keep the same set of hyperparameter values until the end of the optimization procedure, i.e. until the training budget for each hyperparameter setting has been reached.
% With the dynamic variant of random search however, we change the hyperparameter settings on the fly.
% This new batch recovers the momentary optima of both hyperparameters in the example, successfully adapting to the non-stationarity of the hyperparameters.
Looking at the loss averaged over the whole optimization procedure (Last Column of Figure~\ref{fig:gsVSrs}) we can see that this view, although often used with static configuration approaches, abstracts away a lot of information. For example, both approaches would view $\zeta_1$ as much less important than $\zeta_0$ even though it had an overall larger impact during the whole optimization procedure.
At $n=0$, low values for $\zeta_1$ resulted in bad performance whereas at $n=N$, large values resulted in bad performance. If we now average the performance over time, it appears to have little to no impact.
% Furthermore, by not adapting to the change in the optimization landscape, both static approaches result in a worse final loss than the dynamic random search.
Similarly, looking only at the final performance hides a lot of valuable information and can lead to sub-optimal hyperparameter configurations at different stages of the run.
Thus, these methods might be insufficient in cases where changes in hyperparameters are required during the run to achieve the best performance.

A common way to improve the performance of Random Search is with Hyperband \shortcite{hyperband}, a bandit-based configuration evaluation for hyperparameter optimization. It focuses on speeding up Random Search through adaptive resource allocation and early-stopping. The hyperparameter optimization task is considered as a pure-exploration non-stochastic infinite-armed bandit problem where a predefined resource like iterations, data samples, or features is allocated to randomly sampled configurations. In particular, Hyperband uses Successive Halving \shortcite{successive_halving_jamieson_aistats_16} which allocates a budget to a set of hyperparameter configurations. This is done uniformly, and after this budget is depleted, half of the configurations are rejected based on performance. The top $50\%$ are kept and trained further with twice the budget. This process is repeated until only one configuration remains. \shortciteA{pbtbt} used Random Search and Hyperband for tuning hyperparameters of their MBRL algorithm. Furthermore, \shortciteA{pbtbt} analyzed the correlation of configuration performance across the considered budgets, and based on low correlations, they moved from static to dynamic configuration methods.

\paragraph{Open Problems}
The key drawback of random/grid search approaches is being unable to leverage the information regarding promising regions of the hyperparameter search space for making informed decisions. During the optimization, this information becomes clearer as more and more hyperparameter configurations are tried out. However, as these approaches solely focus on exploration and do not exploit this information, they typically scale poorly as the search space increases in size.

%\vcom{we have not highlighted the use of Random Search/Grid Search/Hyperband in parallelly optimizing hyperparameters for AutoRL. This section is currently disconnected to other sections.}

\subsection{Bayesian Optimization}
\label{sec:bo}
\begin{figure}[h]
    \center
    \includegraphics[keepaspectratio, width=0.8\textwidth]{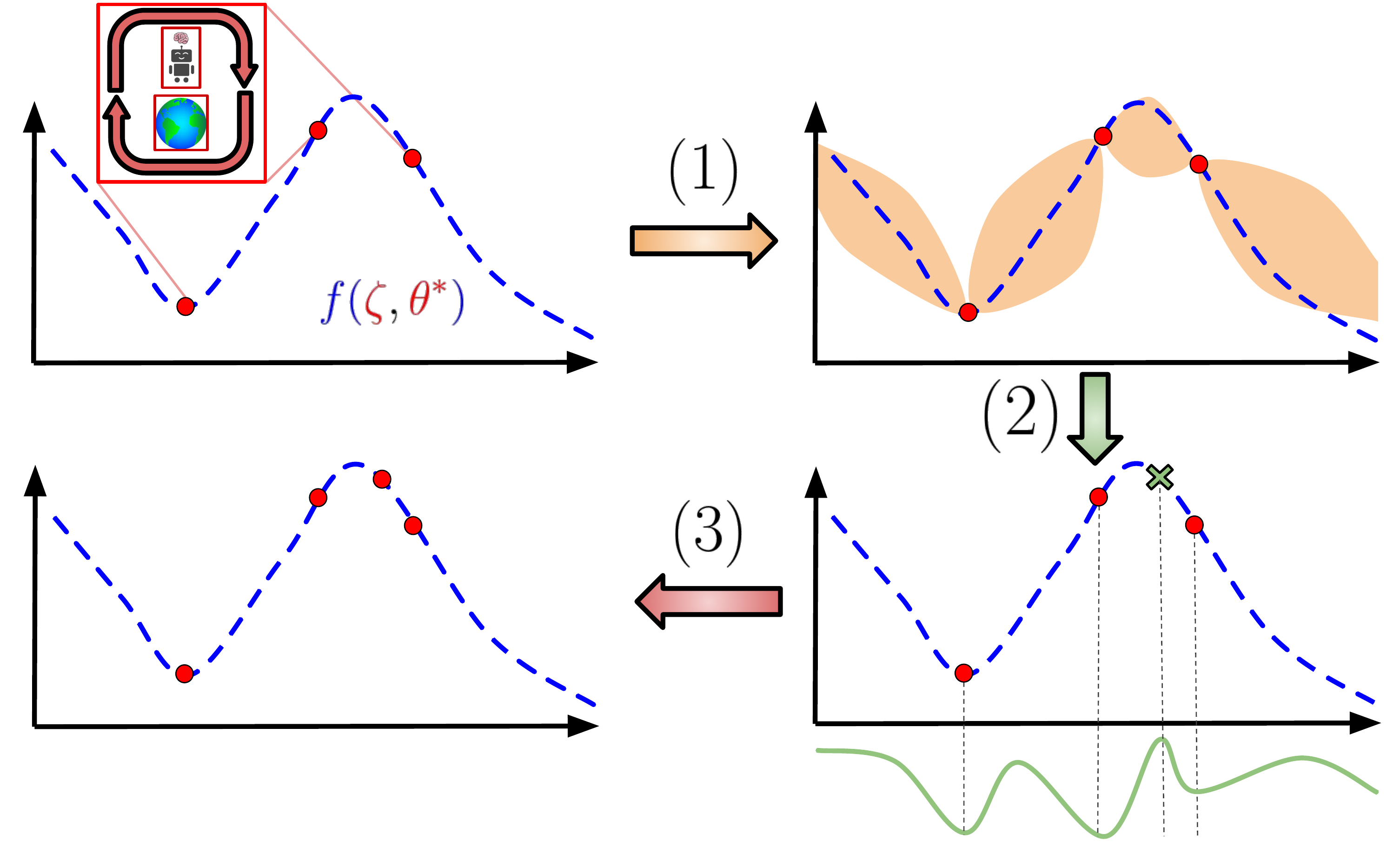}
\caption{Bayesian Optimization, which consists of 3 main steps. (1): A regressor (commonly a Gaussian Process) is used to construct uncertainty estimates on the blackbox function $f(\cdot)$ given previous evaluation data. (2): An acquisition function is constructed from the regressor, which represents the explore-exploit tradeoff on $f$. (3): The argmax of the acquisition function is used for the next trial.}
\label{fig:bayesopt}
\end{figure}

The next set of approaches involved are those which inherently have a notion of sequential decision making. Bayesian Optimization (BO,  \shortciteA{original_bayesopt,jones1998efficient,bayesopt_nando}) is one of the most popular approaches to date, used for industrial applications \shortcite{vizier,botorch,perrone2021amazon} and a variety of scientific experimentation \shortcite{materials_bayesopt_Frazier_2015,hernandez2017parallel,li2018accelerating,griffiths2020constrained,tran2021simulation,van2021personalized}. For RL applications, one of the most prominent uses of BO was for tuning AlphaGo's hyperparameters, which include its core Monte Carlo Tree Search (MCTS) \shortcite{mcts_browne_ieee_12} hyperparameters and time-control settings. This led to an improvement of AlphaGo's win-rate from $50\%$ to $66.5\%$ in self-play games \shortcite{bo_alphago}. In Figure \ref{fig:bayesopt}, we demonstrate the general concept of Bayesian Optimization in the RL case.

%\todo{Frank: I would first mention that standard BO requires expensive black-box evaluations, and then discuss speedup techniques, in particular BOHB and BOIL. The BOHB paper did have RL experiments, and we used BOHB for the RNA application. We could also discuss different fidelities one can use (at least number of seeds and number of epochs.)} \vu{done, thanks!}

\paragraph{Multi-Fidelity Algorithms:}
Because of the fairly large optimization overhead, standard BO methods require expensive black-box evaluations, such as training an ML algorithm to the end to observe the accuracy. This way of training is time-consuming and costly because each set up may take hours or days to run. An alternative choice to the standard BO for tuning RL hyperparameters is leveraging different fidelities \shortcite{cutler2014multifidelity_rl,bayesopt_multifidelity,falkner2018bohb,multifidelity_bo}, such as different number of seeds and epochs, or runtimes.
%Particularly, recent works have attempted to exploit internal information readily available from RL applications for improving the optimization, which we describe below. %We review below one particular methods BOHB \shortcite{falkner2018bohb} and BOIL \shortcite{boil}.
While one option is to use the model to choose both a configuration and a fidelity to evaluate it at~\shortcite{bayesopt_multifidelity,klein-aistats17,multifidelity_bo}, a simpler and more robust approach is to decide about the fidelities using Hyperband (HB) and only choose the configurations that Hyperband should consider at the beginning of each iteration with BO instead of random selection. This is the method used in the popular multi-fidelity BO method BOHB \shortcite{falkner2018bohb}, which trains BO's model for a given fidelity on the configurations that were evaluated at this fidelity so far and which can also efficiently and effectively take advantage of parallel resources. BOHB has been shown to be orders of magnitude faster than vanilla BO in some cases and to find dramatically better results of PPO on the cartpole swing-up task from Gym using seeds as a fidelity~\shortcite{falkner2018bohb}. It also powered one of the first full AutoRL applications: optimizing a PPO approach that was tasked to learn to design RNA~\shortcite{runge2018learning}. In that application, BOHB was used with a runtime budget as fidelity to jointly optimize the neural architecture and hyperparameters of PPO, along with the formulation of the problem as a Markov decision process (MDP), ultimately learning an RNA design method that was over 1000 times faster than the previous state of the art. It is noteworthy that AutoRL made this project possible to be carried out by two undergraduate students with no prior experience with RL.

\paragraph{Exploiting additional information:}
Recent works have attempted to exploit internal information readily available from RL applications for improving the optimization, which we describe now.
BOIL \shortcite{boil} enhances tuning performance by modelling the training curves,  providing a signal to guide search. It transforms the whole training curve into a numeric score to represent high vs low-performing curves. Then, BOIL introduces a data augmentation technique leveraging intermediate information from the training curve to inform the underlying GP model. The algorithm not only selects a hyperparameter setting to evaluate, but also how many epochs it should evaluate. Thus, BOIL makes it possible to run a larger number of cheap (but high-utility) experiments, when cost-ignorant algorithms would only be able to run a few expensive ones, resulting in greater sample-efficiency than traditional BO approaches when tuning deep RL agents.
%
% BOHB’s strong anytime performance stems from its use of Hyperband. Quickly evaluating lots of configurations on small budgets allows BOHB to quickly find some configurations that are promising. The strong final performance stems from BOHB’s BO part as the guided search in the end is able to refine the selected configurations.
%
It is also possible to exploit \textit{external} knowledge  of the maximum achievable return, i.e. knowing $\max_{\tau} \sum_{t \ge 0} \gamma^{t} r_{t}$ to improve hyperparameter tuning in RL. This optimum value is available in advance for some RL tasks, such as the maximum reward (when $\gamma=1$) being $200$ in CartPole. To utilize such knowledge, \shortciteA{nguyen2019knowing} proposed to (i) transform the surrogate model and (ii) select the next point using the optimum value. They showed that exploiting this external information can improve the optimization.

%\todo{ML: inconsistent use of tense. Sometimes simple past and sometimes simple present is used. Please make it consistent}
\paragraph{Empirical results:}
BO has been successfully used in various RL case studies. For example,
\shortciteA{hpo_deep_rl_hertel_arxiv_20} employed successive halving \shortcite{successive_halving_jamieson_aistats_16}, random search and BO and concluded that BO with a noise robust acquisition function is the best choice for hyperparameter optimization in RL tasks. Finally, \shortciteA{lu2021revisiting} showed it was possible to improve the performance of agents trained with offline RL when tuning hyperparameters using BO \shortcite{wan2021casmopolitan}.

\paragraph{Open Problems} BO-based approaches usually perform static tuning, which may not be the most effective for RL. While there are BO approaches \shortcite{pb2} that take the temporal nature of the optimization, especially relevant for RL, into account, these are scarce to the best of our knowledge.

\subsection{Evolutionary Approaches}
\label{sec:neuroevolution}

\begin{figure}[h]
    \center
    \includegraphics[keepaspectratio, width=0.8\textwidth]{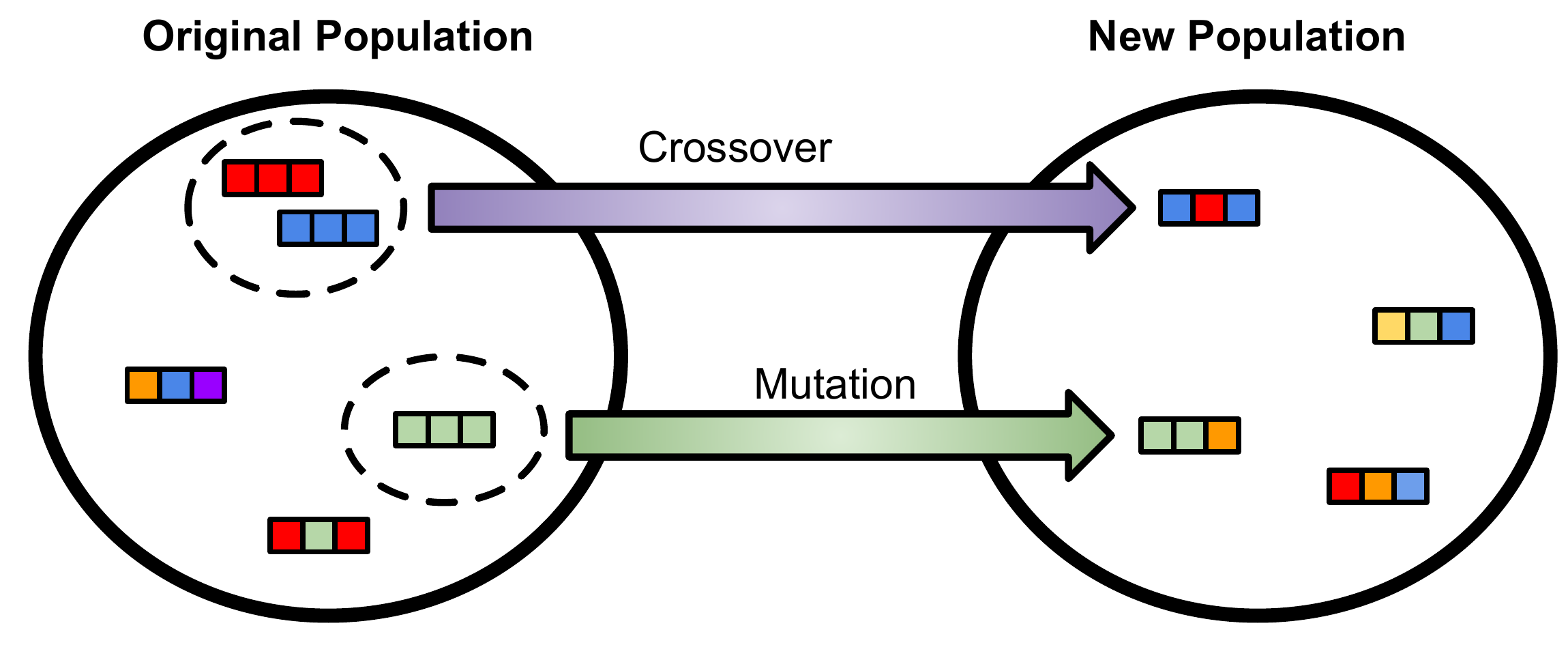}
\caption{Most evolutionary algorithms consist of updating entire populations, usually by mechanisms such as crossover (between multiple genomes) or mutation (on a single genome).}
\label{fig:evolutionary}
\end{figure}

Evolutionary approaches are widely used for various optimization tasks in different fields. These may have different ways of representing the tasks and mechanisms to mutate, recombine, evaluate and select parameters. Such mechanisms are represented in Figure \ref{fig:evolutionary}.

\paragraph{Neuroevolution:} While many in the deep RL community are familiar with Evolution strategies (ES, \shortciteA{es1973}), a specific kind of evolutionary approach that can be used in lieu of an RL algorithm to optimize policies \shortcite{such2018deep,lehman2018es,chrabaszcz-ijcai18a}, there remains little cross pollination with the broader evolutionary computation community. One particular subfield relevant for AutoRL is neuroevolution \shortcite{neuroevolution_nature}, which has been used to evolve both weights and architectures, inspired by biological brains. In particular, the NEAT algorithm was shown to effectively evolve architectures for Pole Balancing in the early 2000s \shortcite{neat}, predating the surge of interest in deep RL. NEAT was made more expressive with the extension to HyperNEAT \shortcite{hyperneat}, which was evaluated in a robotic control task in the original paper. Since then, NEAT and HyperNEAT have been evaluated in a series of control problems \shortcite{hyperneatgaits2013,neuroevolutioncontrol,hyperneat2009gaits,morphologycontrol2013}, even training agents to play Atari games prior to the seminal DQN paper \shortcite{hyperneat_atari}. More recently WANNs \shortcite{wanns_gaier_neurips19} showed it was possible to use NEAT to solve RL tasks by only evolving network topology, using a single randomly initialized weight parameter.

\paragraph{Hyperparameter Optimization:} Evolutionary methods have also been used to search for the hyperparameters of RL algorithms. \shortciteA{hpoGA} used a real-number Genetic Algorithm (GA, \shortciteA{michalewicz2013genetic}), which encodes the hyperparameters of the RL algorithm via genes in every individual of the population, to tune the hyperparameters of SARSA($\lambda$), applying the approach to control a mobile robot. \shortciteA{hpoeo} used GAs to tune the hyperparameters of RL algorithms in simple settings and achieved good performance by combining with automatic restarting strategies \shortcite{RestartGA} to escape from local minima. Similarly, \shortciteA{ddpg_ga} also showed that GAs can be applied to improve the performance of DDPG with Hindsight Experience Replay \shortcite{HER} by tuning hyperparameter configurations. \shortciteA{WhaleOptforRL} used a Whale Optimization Algorithm (WOA, \shortciteA{WhaleOptAlgo}), which is inspired by the hunting strategies of humpback whales, to optimize the hyperparameters of DDPG in various RL tasks in order to improve the performance.
\shortciteauthor{elfwing-arxiv17} proposed \emph{online meta-learning by parallel algorithm competition} (OMPAC; \shortcite{elfwing-arxiv17,elfwing-gecco18}) which similarly to PBT \shortcite{pbt} proposed to use a population of RL agents that is trained in parallel. OMPAC has shown to be successful for tuning hyperparameters of RL agents learning to play Tetris and Atari games.

\paragraph{Multi-Fidelity Algorithms:} Like with Bayesian optimization, evolutionary methods can be sped up by evaluating at different fidelities. DEHB \shortcite{awad2021dehb} combined Hyperband with Differential Evolution (DE), yielding an approach that was up to 1000x faster than random search for the hyperparameter optimization of neural networks and 5x faster to optimize seven hyperparameters of the PPO algorithm (see Figure 9 in the paper).

\paragraph{Population Based Training:}
\label{sec:pbt}

In the RL community \emph{Population Based Training} (PBT, \shortciteA{pbt}) refers to \emph{a specific class of methods} that has found widespread and successful use for hyperparameter optimization but also neuroevolution for RL.
PBT inherits ideas from many evolutionary approaches \shortcite{AdaptingCrossOverinEA,SurveyOfEA,Gloger2004SelfadaptiveEA,Clune2008EA}.
PBT seeks to replicate how a human would observe experiments; it trains multiple agents in parallel and periodically replaces weaker agents with copies of stronger ones. Taking inspiration from Lamarckian Evolutionary Algorithms \shortcite{lamarckian}, where parameters are inherited whilst hyperparameters are evolved, PBT seeks to ``exploit'' stronger agent weights, while ``exploring'' the hyperparameter space, typically through random perturbations. The benefit of this procedure is that it is possible to explore the hyperparameter space with the same wall clock time as a single training run, given access to parallel computational resources.

Another key benefit of PBT and evolutionary approaches is the ability to learn hyperparameter \emph{schedules}, which were shown in the original paper to be particularly effective in RL, likely due to the non-stationarity of the problem (see: Section \ref{sec:nonstationarity}). As such, PBT has had a prominent role in many high profile RL publications \shortcite{kickstarting,liu2018emergent,pbtfootball,impala,Jaderberg859,xland2021,pbtbt}.

\begin{figure}[t]
    \center
    \includegraphics[width=0.9\textwidth]{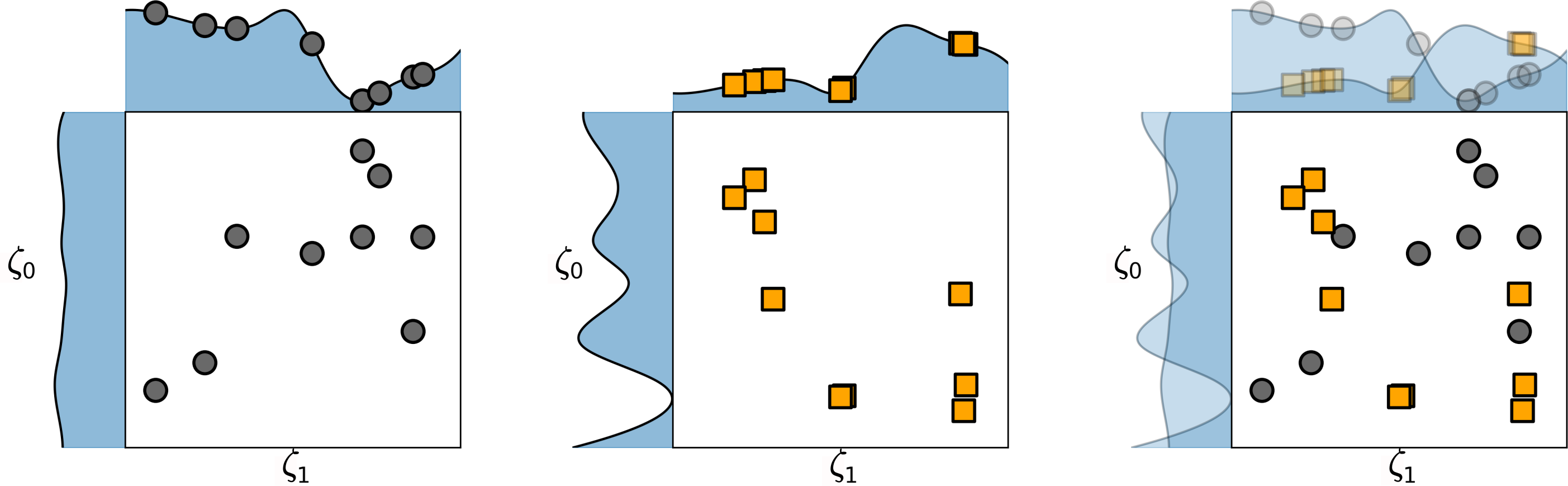}
\caption{Dynamic random search at $t=0$ (first column), $n=N$ (second column) and averaged over $t$. Dynamic random search resamples the hyperparameters $\zeta_0$ and $\zeta_1$ after a fixed time interval $N$.
Circles represent hyperparameters that are sampled at the beginning ($n=0$) of the optimization procedure, orange squares hyperparameters that were sampled at $n=N$ and the blue shaded area depicts the RL agent's loss. This figure is based on Figure 1.1 from~\shortciteA{automl_book}.}
\label{fig:drs}
\end{figure}

In the following we give a simplified view on how evolutionary (and PBT-like) approaches are able to produce hyperparameter schedules. For this we consider a dynamic version of random search: a version that resamples the hyperparmeters after a fixed time interval (see Figure~\ref{fig:drs}).
Assume we use these procedures to \textit{minimize} the loss of an RL algorithm that has two continuous hyperparameters $\zeta_0$ and $\zeta_1$, as in Figure~\ref{fig:gsVSrs}.
With the dynamic variant of random search however, we change the hyperparameter settings on the fly.
In this example, changing the hyperparameters on the fly allows to find better performing hyperparameter schedules than keeping the hyperparameters fixed throughout the run.
Although taking the non-stationarity of the optimization problem into account can result in better performing agents, it poses interesting new challenges for the field of AutoRL.
PBT-like and evolutionary approaches are a natural way of handling the non-stationarity, although they might require far too many resources in cases where hyperparameters need not be changed on the fly.

In recent times, there has been a series of additional improvements over PBT-style algorithms. \shortciteA{franke2021sampleefficient} proposed SEARL and showed it is possible to share experience amongst agents in an off-policy setting, leading to significant efficiency gains. \shortciteA{pbtbt} proposed PBT-BT which included a backtracking component to the exploit step of PBT, and evaluated the approach on a challenging MBRL setting. Finally, \shortciteA{dalibard2021faster} addressed the greediness of the exploitation phase of PBT, and showed that higher long term performance could be achieved by maintaining sub-populations, and using a metric based on expected improvement.

\paragraph{Open Problems} One of the key challenges of evolutionary approaches is their inefficiency, often requiring thousands of CPU cores to achieve strong results. While this may be possible for well-equipped industrial labs, it renders many methods impractical for small to medium sized groups and in many practical applications. Furthermore, there remain very few cases of large-scale evolutionary approaches being applied to both hyperparameters ($\zeta$) and neural network parameters ($\theta$), although recent efforts such as ES-ENAS \shortcite{es_enas_song21} have attempted to do so by combining different evolutionary algorithms.

A disadvantage of PBT-style methods is the relative data inefficiency. This will be particularly important when increasing the search space beyond a handful of hyperparameters as typically tuned with PBT. In addition, the population size is currently a meta-parameter which has a significant impact on the results; so far there has been very little research understanding how the efficacy of PBT works when this changes.

\subsection{Meta-Gradients for Online Tuning}
\label{sec:metagradients}

Meta-gradients provide an alternative approach to dealing with the non-stationarity of RL hyperparameters. The meta-gradient formulation is inspired by meta-learning methods, such as MAML \shortcite{maml_finn_icml17}, which optimize both an inner and outer loop with \emph{gradients}. In particular, meta-gradient methods specify a subset of their (differentiable) hyperparameters to be meta-parameters $\eta$, e.g., bootstrapping, discount factor and learning rate. In the inner loop, the agent is optimized with a fixed $\eta$, taking gradient steps to minimize a (typically fixed) loss function. In the outer loop, $\eta$ is optimized by taking gradient steps to minimize an outer loss function. Each specific choice of the inner and outer loss function defines a new meta-gradient algorithm.

In the original meta-gradient paper, \shortciteA{metagradients}, used IMPALA \shortcite{impala} and set $\eta=\{\lambda, \gamma\}$, the bootstrapping hyperparameter and discount factor. When evaluated on the full suite of Atari games, the meta-gradient version improved over the baseline agent by between $30\%$ and $80\%$ depending on the evaluation protocol used. This work was extended to all differentiable IMPALA hyperparameters with \emph{Self-Tuning Actor-Critic} (STAC, \shortciteA{stac}). STAC introduced a new loss function to handle varying hyperparameters, which allowed it to further boost performance on the same set of Atari games, as well as robotic benchmarks. In addition, \shortciteA{stac} also introduced a means to integrate meta-gradients with auxiliary tasks, producing STACX, which saw further gains. Interestingly, the agents learned non-trivial hyperparameter schedules which could not have been tuned by hand as they were not smoothly varying or static schedules usually used in manual tuning. More recently, \shortciteA{flennerhag2022bootstrapped} proposed \emph{Bootstrapped Meta-Learning}, which first bootstraps a target from the meta-learner, then seeks to match it in a meaningful space, making it possible to extend the meta-learning horizon. This results in state-of-the-art performance for model free agents on the Atari benchmark, by tuning multiple hyperparameters on the fly.

Meta-gradients have also been used to discover auxiliary tasks \shortcite{veeriah2019discovery}, discover options \shortcite{veeriah2021options} and learn RL objectives online, demonstrating strong asymptotic performance \shortcite{frodo}. Importantly, \shortcite{frodo} showed that it was more effective to meta-learn the \emph{target} rather than the full update rule, with the FRODO algorithm capable of producing strong asymptotic performance. Finally, in the field of NAS, differentiable architecture search can also be viewed as a meta-gradient approach. The DARTS algorithm \shortcite{darts} and its successors have become widely popular as well in the NAS community, due the effective use of gradient-based search. Recently this approach has been shown to be effective in RL, finding effective architectures in challenging environments \shortcite{rl_darts,visionary_akinola21}.

\paragraph{Open Problems} Meta-gradient methods have two well-known weaknesses. First, they typically rely on their meta-parameters being initialized well, which itself is a hyperparameter optimization problem. In addition, current meta-gradient methods cannot tune non-differentiable hyperparameters, e.g., the choice of optimizer or the activation function. Nonetheless, they remain one of the most efficient approaches, and offer the potential to improve upon existing strong algorithms with known hyperparameters.

\subsection{Blackbox Online Tuning}
\label{sec:blackbox_online_tuning}
The strength of both PBT and meta-gradients is the ability to adapt hyperparameters on the fly. However, these are not the only approaches to do so. Indeed, a variety of other methods have been considered, from blackbox methods to online learning inspired approaches. This section focuses on single agent approaches to adapt on the fly in settings where the hyperparameters are not (necessarily) differentiable.

Adaptive methods for selecting hyperparameters have been prominent since the 1990s. \shortciteA{Sutton94onstep-size} proposed three alternative methods for adaptive weighting schemes in TD algorithms. \shortciteA{kearnssingh} derived upper bounds on the error of temporal-difference algorithms, and used these bounds to derive schedules for $\lambda$. \shortciteA{tdbma} used Bayesian Model Averaging to select the $\lambda$ bootstrapping hyperparameter for TD methods. More recently, \shortciteA{white2016greedy} proposed $\lambda$-greedy to adapt $\lambda$ as a function of the state and achieve an approximately optimal bias-variance trade-off. \shortciteA{hoof} proposed HOOF which uses random search with off-policy data to periodically select new hyperparameters for policy gradient algorithms.

Several algorithms make use of bandits to adapt hyperparameters. In a distributed setting, \shortciteA{schaul2020adapting} proposed to adapt several behavioral hyperparameters, such as the degree of stochasticity of an agent, using a notion of learning progress as feedback. This idea inspired Agent57 \shortcite{agent57} which adaptively selects from several exploration policies to become the first algorithm to achieve human level performance on all 57 games in the Arcade Learning Environment \shortcite{ale}. Other methods have had success using bandits to select the degree of exploration \shortcite{rp1}, the degree of optimism for off-policy methods \shortcite{top} or the amount of diversity to add to a population of agents \shortcite{dvd}. Finally, \shortciteA{adaptive_td_carlos_neurips_2019} considered adaptively switching between Temporal Difference (TD) learning and Monte Carlo (MC) policy evaluation using learned confidence intervals that detect biases of TD estimates.

\paragraph{Open Problems} One of the challenges of these approaches is the limited scope for the search space, since the bandit algorithm must be able to explore all arms. In addition, most bandit algorithms assume the arms to be independent which may lead to decreased efficiency by removing information which is known to the algorithm designer.

\subsection{Learning Reinforcement Learning Algorithms}
\label{sec:discovering_algs}
Initial approaches to learning RL algorithms employed the learning to learn formulation. \shortciteA{wang-l2rl-arxiv16} and \shortciteA{duan-rl2-arxiv16}, both, use a meta-learner that updates the weight of an agent equipped with an RNN to learn over a distribution of interrelated RL tasks. The former's main interest was in structured task distributions (e.g., dependent bandits) while the latter focussed on unstructured task distributions (e.g., uniformly distributed bandit problems and random MDPs). \shortciteA{wang-l2rl-arxiv16} even showed generalization, to some extent, beyond the exact training distribution encountered. Related to these, \shortciteA{pearl_rakelly_icml19} introduced PEARL, which adapts to new tasks by performing inference over a latent context variable on which the policy is conditioned. They use Variational Autoencoders (VAEs) \shortcite{vae_kingma_iclr_14} to perform such inference. \shortciteA{varibad_zintgraf_iclr20} introduced variBAD, which uses meta-learning to utilise knowledge obtained in related tasks and perform approximate inference in unknown environments to approximate Bayes-optimal behaviour. They also used VAEs perform such inference but used an RNN as the encoder.

Recently however, there has been increased interest in learning to learn or Meta-RL, which aims to automate the design of RL loss functions. A key insight is that a loss function can be viewed as a parameterizable object that can be learned from data. Instead of using a fixed loss function $\mathcal{L}(\theta)$, one can construct a family of loss functions $\mathcal{L}(\theta;\zeta)$ parameterized by $\zeta$, and seek to maximize the expected reward $J(\theta; \zeta)$ via optimizing the surrogate loss $\mathcal{L}(\theta; \zeta)$. In what follows, we separate $\zeta$ into two cases: 1) neural loss functions where $\zeta$ encodes neural network parameters, and 2) symbolic loss functions where $\zeta$ encodes computational graphs.

\paragraph{Neural loss functions:} In this case, the loss function is a neural network with parameters $\zeta$ which may be optimized via ES or gradient based methods. For example, in Evolved Policy Gradient \shortcite{evolvedpg}, the inner loop uses gradient descent to optimize the agent's policy against a loss function provided by the outer loop. The policy's performance is used by the outer loop to evolve the loss parameterization, using ES.

Gradient based methods are closely related to meta-gradient methods, as they use second-order derivatives to optimize $\zeta$. \shortciteA{ml3} described a general framework for meta learning with learned loss functions. Most meta-gradient RL algorithms, as mentioned in Section \ref{sec:metagradients}, start with a human designed loss function $\mathcal{L}(\cdot)$ and modify it with parameterization $\mathcal{L}(\theta; \zeta)$ to allow inner loop and outer loop procedures. In the inner loop, gradient descent via $\nabla_{\theta} \mathcal{L}(\theta; \zeta)$ obtains $\theta^{*}(\zeta)$, which can be viewed as a function of $\zeta$. In the outer loop, the quality of $\zeta$ is then measured by $\mathcal{L}(\theta^{*}(\zeta); \zeta)$, and we can optimize $\zeta$ via gradient descent using $\nabla_{\zeta}\mathcal{L}(\theta^{*}(\zeta); \zeta)$. This approach requires computing second-order information, i.e. $\nabla_{\zeta} \nabla_{\theta}$, although usually only Jacobian-Vector products are needed for the pipeline, which are readily available in popular autodifferentation libraries (Jax \shortcite{jax2018github}, Tensorflow \shortcite{tensorflow2015-whitepaper}, Pytorch \shortcite{pytorch_neurips_2019}).

Examples which employ the above technique include MetaGenRL \shortcite{learned_objectives_meta_rl_kirsch_iclr20}, which is an extension to the DDPG actor-critic framework where the critic is trained to minimize the TD-error as usual. Meta-learning is applied to the policy update. In the original DDPG, an additional policy parameter $\phi$ is trained to maximize the value function $Q_{\theta}(s_{t}, \pi_{\phi}(s_{t}))$. In MetaGenRL, this is used as the outer loss $Q_{\theta}(s_{t}, \pi_{\phi^{*}(\zeta)}(s_{t}))$, and the inner loss $\mathcal{L}(\phi; \zeta)$ is modeled as an LSTM, which can be transferred to different tasks.

In Learned Policy Gradient (LPG) \shortcite{learnedPG}, instead of using a neural meta loss function end-to-end, an LSTM network is used to provide target policy $\hat{\pi}$ and target value $\hat{y}$. The meta loss and the task loss are defined manually. \shortciteA{learnedPG} show that a learned update rule can also be transferred between qualitatively different tasks. More recently \shortciteA{symmetries_bbml2021} showed that incorporating symmetries could lead to improved transfer to unseen action and observation spaces, tasks, and environments.

\paragraph{Symbolic loss functions:} In this case, the loss function is represented as a symbolic expression consisting of predefined primitives, akin to genetic programming. \shortciteA{meta-curiosity} used a Domain Specific Language (DSL) to represent a curiosity module as a directed acyclic graph (DAG). The curiosity module provides intrinsic rewards, which are added to the loss function and can be optimized by PPO. The search space contains 52,000 valid programs and they are evaluated exhaustively with pruning techniques, such as training in a cheap environment and predicting algorithm performance ("meta-meta-RL"). \shortciteA{evolving_algos} proposed a DSL to represent the entire loss function as a DAG, and this loss function is used to train value-based agents. Unlike \shortciteA{meta-curiosity}, the DAG is optimized using an evolutionary algorithm called Regularized Evolution \shortcite{regevo}.

%Learned Policy Gradient \shortcite{learnedPG}, learned a policy gradient algorithm on a small task and then showed it could perform well on Atari games. It has also been shown to be possible to \emph{evolve} reinforcement learning algorithms \shortcite{evolving_algos}.

%Here we can included algorithms in the meta-learning space. We do not want to focus on gradent based meta learning (E.g. MAML) because it is not about learning parts of the RL algorithm, just learning useful parameters. However, other approaches like FRODO \shortcite{frodo}

%\acom{I think RL$^2$ and L2RL should also fit into the last class. Added the refs \shortcite{duan-rl2-arxiv16,wang-l2rl-arxiv16}. Need to double check the details.}

%\ycom{If we focus on meta loss functions, then RL$^2$ doesn't fall into this category; it uses cross-episode RNN state to represent a "fast" RL algorithm, which is quite implicit. Maybe we can still mention RL$^2$ breifly in the end -- as edge cases. }

Finally, regarding the categorical choice of an algorithm for RL, \shortciteA{rl_as_laroche_iclr18} used Algorithm Selection for RL where, given an episodic task and a portfolio of off-policy RL algorithms, a UCB bandit-style meta-algorithm selects which RL algorithm is in control during the next episode so as to maximize the expected return. They evaluated their approach, among others, on an Atari game, where it improved the performance by a wide margin.

\paragraph{Open Problems} In general, it is difficult to understand analytical properties of neural loss functions. Developing tools to empirically study these loss functions becomes crucial for explaining why they work and for understanding their generalization capabilities. For symbolic loss functions, the search space (of all possible graphs) and the search algorithm (e.g., regularized evolution) play a fundamental role, yet it is far from clear what the optimal design choices are.

\subsection{Environment Design}
\label{sec:envdesign}

Environment design is an important component for automating the learning in RL agents. From curriculum learning \shortcite{jiang2020prioritized,space_eimer_icml_21,spdrl_klink_nips_20,teacher_student_curri_matiisen_ieee_20,auto_curricula_sukhbaatar_iclr_18} to synthetic environment learning \shortcite{synth_envs_ferreira_metalearn_21a} and generation \shortcite{evolving_mario_gan_gecco_18} to combining curriculum learning with environment generation \shortcite{gpn_bontrager_arxiv_20,poet_wang_gecco_19,enhanced_poet_wang_icml_20,paired_dennis_nips_20}, the objective here is to speed up the learning for the RL agents through environment design. %In this section, we will explore literature that performs automated environment design (Section \ref{sec:auto_envdesign}) and describe papers that are manually designed but can be used to perform AutoRL (Section \ref{sec:man_envdesign_for_autorl}).

%We begin with automated environment design in the next sub-section and then proceed to describe manually designed environments for AutoRL in the sub-section after.
% Model-Based RL (MBRL) which learns a model of the environment and is a core traditional RL technique can be viewed through the lens of speeding up learning (in terms of samples) on the real environment through environment design.
%  open-ended learning environments
% round robin paper too for curriculum?
% \rr{What about reward shaping? It learns a component of the env. R. Like MBRL it could also be viewed through the AutoRL lens but I don't think we'll include either of these as AutoRL.}

\begin{figure}[h]
    \center
    \includegraphics[keepaspectratio, width=0.8\textwidth]{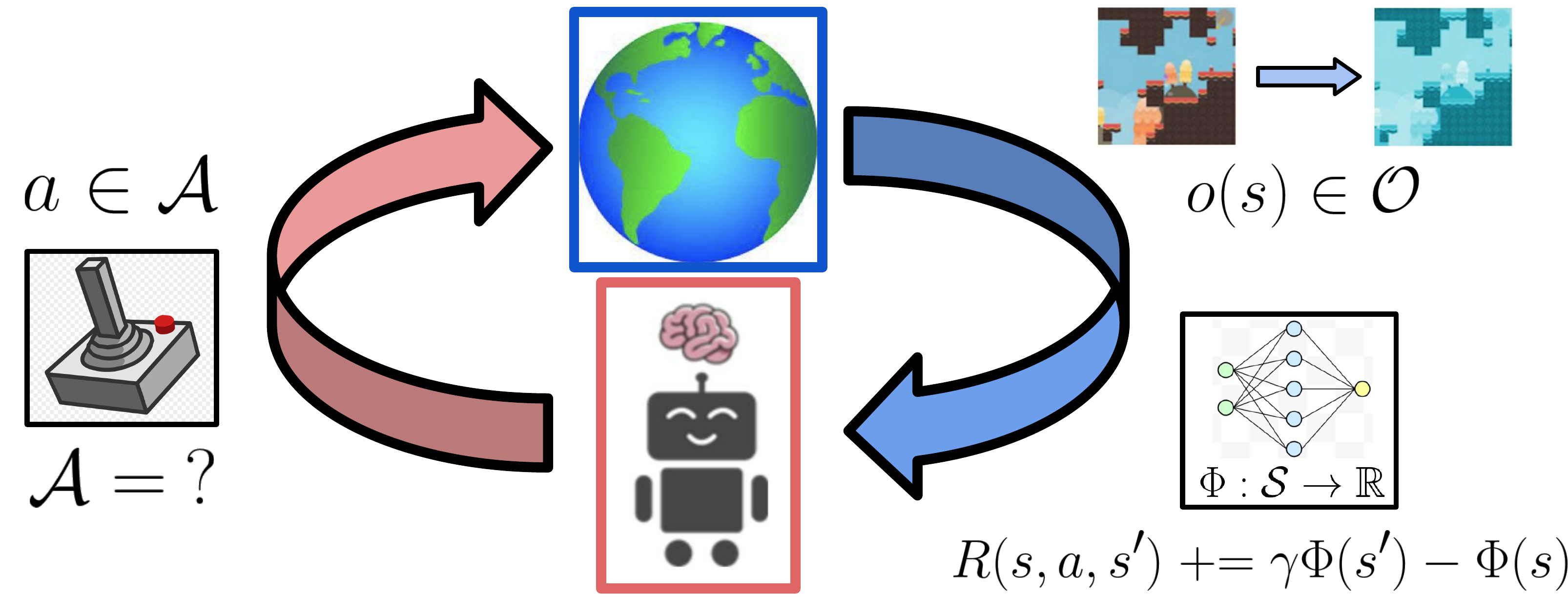}
\caption{Representation of a typical policy + environment feedback loop, with controllable components such as the reward function $R$ from \shortcite{reward_shaping_ng_icml_99}, action space $\mathcal{A}$, and observation space $\mathcal{O}$ from \shortcite{ucb_drac}.}
\label{fig:environment_design}
\end{figure}

% \subsubsection{Automated Environment Design}
% \label{sec:auto_envdesign}
% \textbf{Automated Environment Design}
We organize algorithms performing Automated Environment Design according to the component of the environment (assumed to be a POMDP as defined in Section \ref{sec:preliminaries}) that they try to automatically learn. This organization can also be seen in Figure \ref{fig:environment_design}.

% \paragraph{Transition function, $P$:}
% \shortciteA{synth_envs_ferreira_metalearn_21a} proposed a method that learns a given target environment's transition function in order to train agents more efficiently. They framed this as a bi-level optimization problem and optimized the learnt environment (an NN) in an outer loop to maximize the reward of the agent in the inner loop. They showed that their \textit{synthetic environments} can not only serve as a more efficient proxy for expensive target environments, but that training on them can also reduce the number of training steps.

\paragraph{Reward function, $R$:} \shortciteA{learned_intrinsic_rewards_zheng_neurips_18} presented a bi-level optimization approach that learns reward shaping for policy gradient algorithms using a parameterized intrinsic reward function. In an inner loop, the parameters of the agent are updated using gradient descent to increase the sum of intrinsic and extrinsic rewards, while in the outer loop, the parameters of the intrinsic reward are updated using gradient descent to increase only the extrinsic rewards.
\shortciteA{learn_reward_shaping_neurips_20} also presented a bi-level optimization approach which they term BiPaRS (and its variants). They use a PPO \shortcite{ppo_schulman_arxiv_17} agent in the inner loop. Furthermore, they use meta-gradients with respect to the (state-action dependent) parameters of \textit{user-defined} reward shaping functions to learn helpful reward functions in the outer loop. \shortciteA{reward_shaping_via_meta_learning_arxiv_19} proposed to meta-learn a potential function \shortcite{reward_shaping_ng_icml_99} prior over a distribution of tasks. Inspired by MAML \shortcite{maml_finn_icml17}, they try to adapt the meta-learned prior towards the optimal shaping function (which they derive to be equal to the optimal state-value function $V$).  \shortciteA{autonomous_shaping_konidaris_icml_06} introduced a function that preserves value information between tasks and acts as the agent’s internal shaping reward. The input to this function is the part of the state space whose meaning does \textit{not} change across tasks. This function can be used as the initial estimate for newly observed states and is further refined based on parts of the state space whose meaning \textit{does} change across tasks. \shortciteA{multi_task_evolutionary_shaping_snel_gecco_10} used an evolutionary feature selection method and showed that this method chooses representations that are suitable for shaping and value functions. Their algorithm can find a shaping function in multi-task RL without pre-specifying a separate representation. \shortciteA{automatic_shaping_and_decomposition_marthi_icml_2007} proposed automatically learning a shaped reward and a decomposed reward. Given a set of state and temporal abstractions, they create an abstracted MDP over the original MDP and learn the reward shaping using collected reward samples based on a mathematical formulation they derive.

\shortciteA{faust2019evolving} investigate the effect of learning a multi-objective intrinsic reward over several RL algorithms and benchmarks with an evolutionary search. The results show that a) the utility of automated reward search correlates with the difficulty of the task, b) using simple sparse task objectives (e.g. distance travelled in MuJoCo tasks) as a fitness function leads close to identical results to using the default yet complex hand-tuned MuJoCo rewards, and c) reward search is more likely to produce better policies that the HP search on a fixed training budget and with reasonable hyperparameters.
%The study concludes with the best practices aimed at the practitioners: when training RL on complex tasks, first come up with reasonable hyper-parameters, then automate the inartistic reward search. Automate the intrestic reward search by formulating the reward as a parametrized features relevant to the problem (energy, safety, progress to the goal etc...), and use simple task objective (maximize the distance, reach a goal etc.) as a fitness function. Then search for the inartistic reward and corresponding policy with an evolutionary search.

\paragraph{Action Space, $\mathcal{A}$:} Simplifying an environment's action space can make training significantly easier and faster. The two main approaches used to accomplish this are repeating single actions in order to reduce the number of overall decision points or to construct a new action space entirely, composed of combinations of basic actions as macro actions or options.
While handcrafting has been common in action space augmentation, \shortciteA{learning_to_repeat} learned the number of repetitions for a given action alongside the action selection. \shortciteA{temporl} presented an extension of the idea by conditioning the amount of repetition on the predicted action itself.
In automatic macro action discovery, \shortciteA{macro_actions} showed that it is possible to combine actions into suitable macro actions by defining sub-goals to be reached in the environment; this is achieved by partitioning its transition graph.
Options are similar to macro actions, but instead of executing a macro action once, each option has its own intra-option policy that is followed until a termination function defers back to the agent \shortcite{options}. \shortciteA{ac_options} jointly learned both a policy across these options as well as the options themselves. \shortciteA{robust_options} extended the idea to also learn options that are robust to model misspecification.

\paragraph{Observation Space, $\mathcal{O}$:} \shortciteA{ucb_drac} proposed using a UCB bandit to select an image transformation (e.g. crop, rotate, flip) and applying it to the observations. The augmented and original observations are passed to a regularized actor-critic agent which uses them to learn a policy and value function which are invariant to the transformation.

% \subsubsection{Manual Environment Design for AutoRL}
% \label{sec:man_envdesign_for_autorl}
% \textbf{Environment Design for AutoRL}

\paragraph{Multiple Components, Unsupervised:} We discuss here approaches that change multiple components in the environment or the whole environment itself. Most notable here are curriculum learning approaches that usually modify the state space, $S$, and the initial state distribution, $\rho$. As a result of modifying $S$, they naturally also change the $P$ and $R$ since these are functions with $S$ a component of their domains. We begin with unsupervised methods seeking a generally robust agent, before moving to supervised approaches which typically have a target goal or task.

\shortciteA{poet_wang_gecco_19} as well as \shortciteA{enhanced_poet_wang_icml_20} proposed POET that generates new environments according to an environment encoding. The environments are automatically generated, either with random mutations \shortcite{poet_wang_gecco_19} or if the new environments create a significantly different ranking of existing agents \shortcite{enhanced_poet_wang_icml_20}. These methods of generating new environments are assumed to create a diversity of environments on which to train agents in a curriculum. \shortciteA{quadrup_poetlike} applied the principle to controlling quadrupedal robots and found the robustness to increase significantly.

%\footnote{We categorize this under modification of the state space $S$ because the action space $A$, transition function $P$, reward function $R$ do not change and the observation space $\mathcal{O}$ and observation function $\Omega$ are only affected due to the change in the state space.}

\shortciteA{paired_dennis_nips_20} proposed PAIRED that also automatically generates new environments to create a curriculum. They used a fixed environment encoding. In contrast to \shortciteA{poet_wang_gecco_19,enhanced_poet_wang_icml_20}, environments are chosen to maximize \emph{regret}, defined as the difference in performance between the protagonist agent and an additional \textit{antagonist} agent. This means the adversary is encouraged to propose the simplest environments the protagonist cannot currently solve, while provides theoretical guarantees that (at equilibrium) the protagonist follows a minimax regret strategy. Extending PAIRED, \shortciteA{gur2021code} proposed Compositional Design of Environments (CoDE) for compostional tasks. CoDE's environment generative model builds a series of compositional tasks and environments tailored to the RL agent's skill level and makes use of a population of agents, making it possible to train agents to navigate the web. Extending the theoretical framework from PAIRED in a different direction, \shortciteA{jiang2021robustplr} showed that rather than learn to \emph{generate} new environments, it is also effective to \emph{curate} randomly sampled ones using \emph{Prioritized Level Replay} (PLR, \shortciteA{jiang2020prioritized}). This approach maintains the theoretical properties of PAIRED while demonstrating stronger empirical performance.

%and observing their actual performance, in addition to regret. CoDE then adds an explicit difficulty loss, which encourages the generative model to tune the difficulty (how many sub-tasks) based on the observed competency of the agent, and use regret for fine-tuning of the environment and selection of which design elements to use (sub-task selection).

%The resulting agent is able to generalize to space of $10^31$ tasks, induced with only 1500 sub-tasks.

%The regret is calculated as the difference between the performance of the protagonist and antagonist on sampled trajectories in sampled environments.

\shortciteA{gpn_bontrager_arxiv_20} also employed an environment generator (in addition to human created environments) and an actor-critic agent which work both cooperatively and adversarially to create a curriculum for the agent by selecting environments minimizing the agent's expected value.

% They used automatically generated environments and human created environments in an equal proportion.
% The generator network tries to create environments that provide an appropriate level of challenge for the agent based on estimates by the critic.
%. So, initially when the agent has negative rewards, this results in the generator creating environments that would be easy for the agent but once the agent starts obtaining positive rewards the generator creates environments that still have 0 utility for the agent and thus makes it harder for the agent to do well.

\paragraph{Multiple Components, Supervised:} If the agent's goal is solving a specific task instead of being generally more robust, it can be supported by continually progressing a simple version of the task towards this goal in a curriculum \shortcite{narvekar-cl-survey-jmlr20}.
Progression is usually tied to how well the agent currently performs in order to keep the environment difficulty at an appropriate level. A well-known example of this is \shortciteA{openai-rubikscube} who showed it is possible to use a robot hand to solve the Rubik's cube, by starting from an almost solved cube and gradually increasing the starting position complexity of both hand and cube as soon as the agent can solve the current state sufficiently well.
\shortciteA{spdrl_klink_nips_20} applied the same principle to several physics simulation tasks, using value estimations as the progression criterion.
Value estimation approaches in general have been successful for RL because their approximation of an environment's challenge level is cheap to compute \shortcite{jian-spl-aaai15,zhang-vds-neurips20,space_eimer_icml_21}.
They are an important class in the starting state curriculum generation taxonomy proposed by \shortciteA{woehlke-spl-aamas20}.
Student-teacher curriculum learning approaches can also create new task variations, e.g. by using GANs, although they only gradually increase the complexity of their distribution as the agent improves during training \shortcite{goal_gan,teacher_student_curri_matiisen_ieee_20,turchetta-neurips20}. We can even induce a difficulty curve into environments that are immutable by using Self-Play to challenge agents with ever more difficult opponents \shortcite{silva-ieee19}. A canonical approach is \emph{Asymmetric Self-Play} (ASP, \shortciteA{auto_curricula_sukhbaatar_iclr_18}) which proposes to use two agents: ``Alice'', who proposes new tasks by doing a sequence of actions and ``Bob'' who must undo or repeat them. ASP was also shown to be highly effective for challenging robotic manipulation tasks \shortcite{openai2021asymmetric}. Finally AMIGo \shortcite{campero2021learning} and APT-Gen \shortciteA{fang2021adaptive} both consider settings with a fixed task, whereby the goal becomes incrementally harder as the student becomes increasingly capable.
%such-gtn-pmlr20,

%Curriculum learning can also be applied for open-ended learning as shown in \shortcite{poet_wang_gecco_19,enhanced_poet_wang_icml_20}. Instead of moving towards a specific task configuration, the general environment complexity is simply increased in time with the agent's increased abilities.

%If the environment parameters are not well enough understood to manually generate challenging variations of a task, methods like \shortcite{paired_dennis_nips_20} can generate valid distributions across a given environment.
%This could be compared to augmenting the agent's observations as in \shortcite{ucb_drac}.

\shortciteA{synth_envs_ferreira_metalearn_21a} proposed a method that learns a given target environment's transition and reward functions in order to train agents more efficiently. They framed this as a bi-level optimization problem and optimized the learnt environment (an NN) in an outer loop to maximize the reward of the agent in the inner loop. They showed that their \textit{synthetic environments} can not only serve as a more efficient proxy for expensive target environments, but that training on them can also reduce the number of training steps.
%Ferreira et al. \shortcite{synth_envs_ferreira_metalearn_21a} also learned a given target environment's using a bi-level optimization framework.

% We describe \shortcite{autorl_learn_reward_nn_arch_chiang_ral19, Jaderberg859} in section \ref{sec:hybrid_approaches} as they employ not just environment design.

%While we have presented approaches that automatically learn $S$, $A$, $P$ and $R$, to the best of our knowledge there is no literature that tries to automatically learn $\mathcal{O}$ or $\Omega$. It might be useful to learn what observations increase the performance on an environment most while keeping the underlying state space fixed. For example, partial observability has been shown to improve generalization performance in grid-worlds \shortcite{paired_dennis_nips_20,transforms_zero_shot_ye_cog_20}.

\paragraph{Open Problems} There are many different approaches for environment design and it is unclear which of them leads to the biggest gains in performance. This raises the possibility of: a) employing several of these methods in a single approach and studying the impact on performance and to what extent the gains are complementary; and b) evaluating the approaches on a shared benchmark. Unfortunately, there is currently also a lack of unified frameworks and shared benchmarks. To further progress towards fully automated progression through environments, more efforts like \shortciteA{romac-tma-icml21} would be helpful in encouraging closer cooperation and better comparability.

% However, differentiable architecture search has also been shown to have frequent catastrophic failure modes~\shortcite{Zela2020Understanding}, and multi-fidelity Bayesian optimization is a more robust approach that has already been used to optimize complex neural architectures for RL including both convolutional and recurrent layers~\shortcite{runge2018learning}.

\subsection{Hybrid Approaches}\label{sec:hybrid_approaches}
%\textcolor{red}{All:}
%\textcolor{red}{Please put methods which are hybrid here}

Inevitably some methods do not fall into a single category. Indeed, many methods seek to exploit the strengths of different approaches, which we refer to as a \emph{hybrid} method. In this section we define these hybrid methods as those which use more than one class of approaches from Table \ref{table:autorl_summary}.

%BOHB \shortcite{falkner2018bohb} combined Hyperband with BO to determine how many configurations to evaluate with which budget, but it replaces the random selection of configurations at the beginning of each Hyperband iteration with BO, where the model is trained on the configurations that were evaluated so far. BOHB can also efficiently and effectively take advantage of parallel resources. BOHB was used to optimize the hyperparameters of PPO on the cartpole swing-up task from Gym and the problem of RNA design from \shortcite{runge2018learning}.
%Its successor SMAC3~\shortcite{LinEgg2021} replaces the KDE surrogate model by a more efficient random forest model. In each iteration BOHB evaluates multiple configurations, which can independently run on multiple workers.

%DEHB \shortcite{awad2021dehb} combined Hyperband with Differential Evolution (DE) to optimize the hyperparameters of the PPO algorithm 5x faster than random search (see Figure 9 in the paper) on an AutoRL task and up to three orders of magnitudes on other hyperparameter optimization tasks.

\shortciteA{autorl_learn_reward_nn_arch_chiang_ral19} combined evolutionary search with reward shaping to automatically select a reward function and neural architecture for navigation task. They learn a more robust policy and a hyperparameter configuration which generalizes better to unseen environments.
% Despite the high computing cost, The previous two papers are hybrid approaches according to our taxonomy as they optimize both, the environment and HPs or NN architectures.

\shortciteA{Jaderberg859} achieved human-level performance in the game Quake III Arena in Capture the Flag mode, using only pixels and game points scored as input. To do this, their \emph{FTW} agent learned reward shaping coefficients and hyperparameters jointly in an outer optimization loop. They use a bi-level optimization process with PBT, with the inner loop optimizing an IMPALA \shortcite{impala} RL agent. The outer loop is evolutionary and maintains a population of such independent RL agents which are trained concurrently from thousands of parallel matches on randomly generated environments. Each agent learns its own internal reward signal and rich representation of the world. More recently, another large-scale project saw the use of environment design, through a form of guided domain randomization, and PBT to produce ``generally capable'' agents in a large simulation environment \shortcite{xland2021}. This work attempted to achieve a more open-ended system, whereby an agent could learn to play a wide variety of games, resulting in multiple innovations, such as generational training which allowed agents to transfer behaviors across different reward functions in a PBT setup.
%\rr{I get the feeling all reward-shaping ones will end up being hybrid unless we include them in earlier sub-sections in, e.g., evolutionary/meta-gradient/etc. approaches as approaches that affect the R and maybe not have an env design section.}

The Population Based Bandits (PB2 \shortciteA{pb2}) approach seeks to combine ideas from both PBT and BO. It formulates the explore step of PBT as a batch GP-bandit optimization problem and uses an upper confidence bound (UCB) acquisition function to select new configurations. In a series of RL problems it was shown to be more sample efficient than PBT, but remains untested in larger problems or with more than a handful of hyperparameters. In addition, PB2 was recently extended to deal with mixed input hyperparameters (continuous and categorical) \shortcite{parkerholder2021tuning}, but there has been very little work exploring kernel choice, improving the time-varying mechanism or further extending, for example, to architectures.

\paragraph{Open Problems} In addition to the open problems for each of the individual sub-sections above, hybrid approaches face the additional open problems of how best to combine the individual approaches. Since the compute requirement of combining many methods and searching for the best combination among these could be extremely large, hybridizing approaches would require being efficiently able to prune the search space of such combinations.

\section{Benchmarks} \label{sec:benchmarks} % from rebuttal discussion

When considering different AutoRL methods, we must inevitably ask how to compare them to one another. There are no established standardized benchmarks for evaluating AutoRL methods as of yet. Instead, many works thus far have tuned components of a baseline RL method and re-used its evaluation environments, typically continuous control tasks from OpenAI Gym \shortcite{gym} or discrete environments from the Arcade Learning Environment \shortcite{ale}.

There is, however, an increasing emphasis on using environments that test generalization. \shortciteA{whiteson2009generalized} were one of the first to propose testing agents on distributions of environments to quantify generalization in RL. The popular modern benchmark of OpenAI Procgen \shortcite{procgen} is based on procedurally generated environment variations and has been used to evaluate AutoRL methods \shortcite{rl_darts,parkerholder2021tuning}. Other procedurally generated environments include CoinRun \shortcite{pmlr-v97-cobbe19a}, MiniGrid \shortcite{gym_minigrid}, NetHack \shortcite{nle}, MiniHack \shortcite{samvelyan2021minihack}, Griddly \shortcite{griddly} or MineRL \shortcite{guss2019minerlcomp}.

Instead of using procedural generation, \shortciteA{mdp_playground_rajan_neurips_workshop_19} as well as \shortciteA{osband2020bsuite} provide simple and configurable environments with a latent causal structure. These low-level benchmarks may be used to perform experiments at a small cost at the expense of the complexity of the environments. CARL~\shortcite{BenEim2021a} provides similar freedom in defining environment distributions via potentially observable context features, but for more complex domains like physics simulation environments.

In addition, it can be worthwhile to consider AutoRL in more targeted, compositional tasks. \shortciteA{meta_world_yu_corl_2019} introduced Meta-World, a benchmark that proposes 50 distinct robotic manipulation tasks and their variations to enable such learning. Alchemy~\shortcite{alchemy_wang_arxiv_21} is another complex benchmark that also proposes a meta-distribution for RL agents which can be used to perform meta-RL to determine an underlying latent causal structure.

Learning over such distributions can not only make RL agents robust to variation but also allow them to perform few-shot learning on a distribution of environments.
AutoRL for generalization \shortcite{hyp_dep_gen} remains an understudied problem, where it becomes a challenge to define an appropriate task for the outer loop.

\paragraph{Open Problems} While RL benchmarks keep evolving, none of the tasks above are specific to AutoRL and thus AutoRL methods are still tested on problems with various degrees of complexity and different needs for generalization. Complicating comparability even further, AutoRL includes additional factors like configuration spaces for hyperparameters or architectures that are not standardized even across methods evaluating on the same RL benchmarks or environments. Thus specific benchmarking protocols that not only control for RL-specific environment factors but also the AutoRL setting are necessary to enable reliable comparison of different AutoRL methods. Finally, the Brax physics engine environments~\shortcite{brax} provide massively parallel simulation with a single GPU, which may make it possible to make rapid progress AutoRL methods.

\begin{comment}
Small environment (e.g. MDP Playground) with differing degrees of hyperperarameter tuning. For example: small-differentiable, large-differentiable, small-mixed, large-mixed etc.

Discuss comparison between PBT/parallel methods vs. single agent processes (such as meta-gradients).

\vcom{Vu : do we need experiments? if so, we need a list of baselines and environments}

\vcom{Vu : shall we build a benchmark repo and release together with the paper? this will make our paper stronger and more useful.}
\rr{Yes, I think so.}

\end{comment}

\section{Future Directions} \label{sec:future_directions}

In this section, we highlight a few specific areas which we believe will be particularly fruitful avenues for future work.

In this survey we emphasized the success of methods for AutoRL that dynamically adapt configurations during training. However, to the best of our knowledge, the non-stationarity of hyperparameters of many modern state-of-the-art algorithms has not been studied extensively. Thus, it is often not clear which hyperparameters need to be optimized dynamically and which are best optimized statically. Furthermore, it remains largely an open question if the impact of hyperparameters remains the same across environments or if different hyperparameters are important for different tasks~\shortcite{EimBen2021a}, and why that may be the case. Methods to analyze such effects have so far only been proposed for static configuration procedures \shortcite{hutter-icml14a,fawcett-heu16a,biedenkapp-aaai17,biedenkapp-lion18a,Rijn-KDD18} and have not yet found wide-spread use in RL.

%\subsection{AutoRL's Relationship with Meta-Algorithmic Frameworks}
Classical hyperparameter optimization \shortcite{automl_book} for (un-)supervised learning considers only individual datasets when searching for well performing configurations. Similarly, when optimizing the hyperparameters of an RL agent commonly only individual environments are considered. In fact, nearly all of the discussed optimization methods throughout this paper consider only individual environments.
Given the sensitivity of RL agents discussed throughout this paper \shortcite{deeprlmatters,andrychowicz2021what,Engstrom2020Implementation}, it is to be expected that the discovered settings are not transferable to other environments. Recently however there is increasing interest in training RL agents that are capable of handling multiple (homogeneous) environments. While agents have typically used single configurations across all Atari games \shortcite{dqn}, it remains a challenge to find robust configurations that work in a variety of settings \shortcite{EimBen2021a}. AutoRL approaches that tune RL agents across environments fall into the class of algorithm configuration \shortcite{HutHooLeyStu09a} methods, which seek to find better parameters to improve general algorithm performance. Such methods are capable of finding well performing and robust hyperparameters for a set of environments \shortcite{eggensperger-jair19a}. Recently, evaluation protocols were proposed that consider the performance of RL algorithms across a set of environments \shortcite{pmlr-v119-jordan20a,rl_generalization_survey,crossenvhyperopt-2021}, which may prove useful for future work in this space. Meanwhile, it also raises the problem of AutoRL-specific benchmarking, which considers different metrics such as performance improvement, data efficiency and generalization capability of the AutoRL method.

Related to the problem of algorithm configuration, algorithm selection \shortcite{rice1976algorithm} can be used to choose which RL algorithm to use for learning \shortcite{DBLP:conf/iclr/LarocheF18}. To the best of our knowledge, so far no AutoRL approach has explored the intersection of algorithm selection and algorithm configuration known as per-instance algorithm configuration (PIAC, \shortciteA{xu-aaai10a,kadioglu-ecai10}).
In this framework, a selector can choose a configuration from a portfolio of well performing configurations and decides which of these to use for the environment at hand.

One important area for research is to provide a more rigorous understanding of the impact of design choices on performance. If we can understand how each component interacts with others then we can either select more amenable combinations or design search spaces to account for this dependency. This could come from empirical investigations or theoretical analysis. Indeed, recent works such as \shortciteA{Fedus2020RevisitingFO}, \shortciteA{obando20revisiting} and \shortciteA{andrychowicz2021what} have provided foundations in this area, but significant work remains to be done. Furthermore, hyperparameter importance methods and analysis tools from the AutoML community \shortcite{hutter-icml14a,fawcett-heu16a,biedenkapp-aaai17,biedenkapp-lion18a,Rijn-KDD18} have not yet been explored in AutoRL. Increasing focus in this area is likely to have profound knock-on effects on other areas of RL, especially AutoRL.

Another important design choice that impacts performance of RL agents is that of \textit{on-} vs \textit{off-policy} RL. It has significant implications for algorithms (e.g., V-trace for IMPALA \shortcite{impala}). While some AutoRL approaches such as SEARL can only work in the off-policy setting, making the choice of whether to use on- or off-policy algorithms could itself be considered a hyperparameter in general and studied further under AutoRL.

Another recent area of research is \emph{offline} RL, where agents must generalize to an online simulator or real world environment from a static dataset of experiences. Research in this space remains nascent, however, it already contains a large diversity of methods, while also introducing new challenges. For example, it is challenging to get real-world policy returns for an AutoRL algorithm, so often we must make use of off-policy evaluation \shortcite{precup2000ope}. It is possible that in this setting there may be scope to use more ``traditional'' AutoML methods, or it may require completely new approaches.

In addition to the discussion so far, the majority of the work in this survey addresses \emph{single agent} RL. However, many real-world systems are in fact inherently multi-agent \cite{jakob2018a}. Consider, for example, self-driving cars which need to cooperate with others to safely drive on the road. When using RL in this type of problem, there is an additional challenge of how to parameterize the different agents: whether to train in a centralized or decentralized manner, while also designing reward functions and algorithms to capture the impact of individual agent actions.

Another important sub-field of RL that was beyond the scope of the survey is Multi-Objective Reinforcement Learning (MORL). While MORL aims at optimizing a vector reward from an environment, various metrics might also need optimizing in the outer loop as mentioned in Section \ref{subsec:eval_methods}. For example, minimizing the memory usage of an algorithm and/or the wall-clock time, which may interfere with purely optimizing the maximum reward from the environment, may be important considerations in choosing the algorithm. Even multi-task RL where performance on different environments could be traded off with each other could be considered an instance of MORL. Such Multi-Objective Optimization could explicitly consider the Pareto front in the outer loop optimization or could even resort to heuristic design choices to constrain the design space of an AutoRL pipeline.

\section{Conclusion} \label{sec:conclusion}

This paper has introduced AutoRL by discussing a wide variety of methods to automate the RL training pipeline. Indeed, unlike supervised learning, which is usually an open-loop one-step process, RL is a complete closed-loop \textit{system}. As such, it is likely each of the components discussed has an influence on others, and if we want to train our agents end-to-end as part of a broader system, it ultimately will require a holistic solution. The challenge is compounded in RL since evaluations in RL are almost always necessarily stochastic and (potentially much more) noisy than in supervised learning, due to various sources (e.g. policy, environment), which can be a challenge for any form of automatic tuning. However, we have presented a variety of promising directions in this survey that can help overcome the challenge and which will likely provide improvements in the coming years. It is evident that AutoRL is maturing as a field and exciting possibilities lay ahead.

\section*{Acknowledgements}
Jack Parker-Holder, Raghu Rajan, and Xingyou Song are listed as first authors in alphabetical ordering of this work. Aleksandra Faust, Frank Hutter, and Marius Lindauer are listed as last authors in alphabetical ordering for their advising throughout this work.

We would like to thank Jie Tan for providing feedback on the survey, as well as Sagi Perel and Daniel Golovin for valuable discussions. Frank, André and Raghu acknowledge Robert Bosch GmbH for financial support.

%\vskip 0.2in
\newpage
\bibliography{main}

\begin{thebibliography}{}

\bibitem[\protect\BCAY{Abadi, Agarwal, Barham, Brevdo, Chen, Citro, Corrado,
  Davis, Dean, Devin, Ghemawat, Goodfellow, Harp, Irving, Isard, Jia,
  Jozefowicz, Kaiser, Kudlur, Levenberg, Man\'{e}, Monga, Moore, Murray, Olah,
  Schuster, Shlens, Steiner, Sutskever, Talwar, Tucker, Vanhoucke, Vasudevan,
  Vi\'{e}gas, Vinyals, Warden, Wattenberg, Wicke, Yu,\ \BBA\ Zheng}{Abadi
  et~al.}{2015}]{tensorflow2015-whitepaper}
Abadi, M., Agarwal, A., Barham, P., Brevdo, E., Chen, Z., Citro, C., Corrado,
  G.~S., Davis, A., Dean, J., Devin, M., Ghemawat, S., Goodfellow, I., Harp,
  A., Irving, G., Isard, M., Jia, Y., Jozefowicz, R., Kaiser, L., Kudlur, M.,
  Levenberg, J., Man\'{e}, D., Monga, R., Moore, S., Murray, D., Olah, C.,
  Schuster, M., Shlens, J., Steiner, B., Sutskever, I., Talwar, K., Tucker, P.,
  Vanhoucke, V., Vasudevan, V., Vi\'{e}gas, F., Vinyals, O., Warden, P.,
  Wattenberg, M., Wicke, M., Yu, Y., \BBA\ Zheng, X. \BBOP2015\BBCP.
\newblock \BBOQ {TensorFlow}: Large-scale machine learning on heterogeneous
  systems\BBCQ.
\newblock Software available from tensorflow.org.

\bibitem[\protect\BCAY{Agarwal, Schwarzer, Castro, Courville,\ \BBA\
  Bellemare}{Agarwal et~al.}{2021}]{rliable}
Agarwal, R., Schwarzer, M., Castro, P.~S., Courville, A.~C., \BBA\ Bellemare,
  M.~G. \BBOP2021\BBCP.
\newblock \BBOQ Deep reinforcement learning at the edge of the statistical
  precipice\BBCQ\
\newblock {\Bem CoRR}, {\Bem abs/2108.13264}.

\bibitem[\protect\BCAY{Akinola, Angelova, Lu, Chebotar, Kalashnikov, Varley,
  Ibarz,\ \BBA\ Ryoo}{Akinola et~al.}{2021}]{visionary_akinola21}
Akinola, I., Angelova, A., Lu, Y., Chebotar, Y., Kalashnikov, D., Varley, J.,
  Ibarz, J., \BBA\ Ryoo, M.~S. \BBOP2021\BBCP.
\newblock \BBOQ Visionary: Vision architecture discovery for robot
  learning\BBCQ\
\newblock In {\Bem {IEEE} International Conference on Robotics and Automation,
  {ICRA} 2021, Xi'an, China, May 30 - June 5, 2021}, \BPGS\ 10779--10785.
  {IEEE}.

\bibitem[\protect\BCAY{Alet, Schneider, Lozano{-}P{\'{e}}rez,\ \BBA\
  Kaelbling}{Alet et~al.}{2020}]{meta-curiosity}
Alet, F., Schneider, M.~F., Lozano{-}P{\'{e}}rez, T., \BBA\ Kaelbling, L.~P.
  \BBOP2020\BBCP.
\newblock \BBOQ Meta-learning curiosity algorithms\BBCQ\
\newblock In {\Bem 8th International Conference on Learning Representations,
  {ICLR}, Addis Ababa, Ethiopia, April 26-30}. OpenReview.net.

\bibitem[\protect\BCAY{Andrychowicz, Crow, Ray, Schneider, Fong, Welinder,
  McGrew, Tobin, Abbeel,\ \BBA\ Zaremba}{Andrychowicz et~al.}{2017}]{HER}
Andrychowicz, M., Crow, D., Ray, A., Schneider, J., Fong, R., Welinder, P.,
  McGrew, B., Tobin, J., Abbeel, P., \BBA\ Zaremba, W. \BBOP2017\BBCP.
\newblock \BBOQ Hindsight experience replay\BBCQ\
\newblock In {\Bem Advances in Neural Information Processing Systems 30: Annual
  Conference on Neural Information Processing Systems, December 4-9, Long
  Beach, CA, {USA}}, \BPGS\ 5048--5058.

\bibitem[\protect\BCAY{Andrychowicz, Denil, Colmenarejo, Hoffman, Pfau,
  Schaul,\ \BBA\ de~Freitas}{Andrychowicz
  et~al.}{2016}]{l2lbGDbGD_andrychowicz_neurips_16}
Andrychowicz, M., Denil, M., Colmenarejo, S.~G., Hoffman, M.~W., Pfau, D.,
  Schaul, T., \BBA\ de~Freitas, N. \BBOP2016\BBCP.
\newblock \BBOQ Learning to learn by gradient descent by gradient descent\BBCQ\
\newblock In Lee, D.~D., Sugiyama, M., von Luxburg, U., Guyon, I., \BBA\
  Garnett, R.\BEDS, {\Bem Advances in Neural Information Processing Systems 29:
  Annual Conference on Neural Information Processing Systems 2016, December
  5-10, 2016, Barcelona, Spain}, \BPGS\ 3981--3989.

\bibitem[\protect\BCAY{Andrychowicz, Raichuk, Sta{\'n}czyk, Orsini, Girgin,
  Marinier, Hussenot, Geist, Pietquin, Michalski, Gelly,\ \BBA\
  Bachem}{Andrychowicz et~al.}{2021}]{andrychowicz2021what}
Andrychowicz, M., Raichuk, A., Sta{\'n}czyk, P., Orsini, M., Girgin, S.,
  Marinier, R., Hussenot, L., Geist, M., Pietquin, O., Michalski, M., Gelly,
  S., \BBA\ Bachem, O. \BBOP2021\BBCP.
\newblock \BBOQ What matters for on-policy deep actor-critic methods? a
  large-scale study\BBCQ\
\newblock In {\Bem International Conference on Learning Representations}.

\bibitem[\protect\BCAY{Arora, Du, Hu, Li, Salakhutdinov,\ \BBA\ Wang}{Arora
  et~al.}{2019}]{cntk}
Arora, S., Du, S.~S., Hu, W., Li, Z., Salakhutdinov, R., \BBA\ Wang, R.
  \BBOP2019\BBCP.
\newblock \BBOQ On exact computation with an infinitely wide neural net\BBCQ\
\newblock In {\Bem Advances in Neural Information Processing Systems 32:
  NeurIPS, December 8-14, Vancouver, BC, Canada}, \BPGS\ 8139--8148.

\bibitem[\protect\BCAY{Ashraf, Mostafa, Sakr,\ \BBA\ Rashad}{Ashraf
  et~al.}{2021}]{WhaleOptforRL}
Ashraf, N.~M., Mostafa, R.~R., Sakr, R.~H., \BBA\ Rashad, M.~Z. \BBOP2021\BBCP.
\newblock \BBOQ Optimizing hyperparameters of deep reinforcement learning for
  autonomous driving based on whale optimization algorithm\BBCQ\
\newblock {\Bem PLOS ONE}, {\Bem 16\/}(6), 1--24.

\bibitem[\protect\BCAY{Awad, Mallik,\ \BBA\ Hutter}{Awad
  et~al.}{2021}]{awad2021dehb}
Awad, N., Mallik, N., \BBA\ Hutter, F. \BBOP2021\BBCP.
\newblock \BBOQ Dehb: Evolutionary hyberband for scalable, robust and efficient
  hyperparameter optimization\BBCQ\
\newblock In {\Bem Proceedings of the Thirtieth International Joint Conference
  on Artificial Intelligence}.

\bibitem[\protect\BCAY{B\"{a}ck}{B\"{a}ck}{1998}]{SurveyOfEA}
B\"{a}ck, T. \BBOP1998\BBCP.
\newblock \BBOQ An overview of parameter control methods by self-adaptation in
  evolutionary algorithms\BBCQ\
\newblock {\Bem Fundam. Inf.}, {\Bem 35\/}(1–4), 51–66.

\bibitem[\protect\BCAY{Bacon, Harb,\ \BBA\ Precup}{Bacon
  et~al.}{2017}]{ac_options}
Bacon, P., Harb, J., \BBA\ Precup, D. \BBOP2017\BBCP.
\newblock \BBOQ The option-critic architecture\BBCQ\
\newblock In {\Bem Proceedings of the Thirty-First {AAAI} Conference on
  Artificial Intelligence, February 4-9, San Francisco, California, {USA}},
  \BPGS\ 1726--1734. {AAAI} Press.

\bibitem[\protect\BCAY{Badia, Piot, Kapturowski, Sprechmann, Vitvitskyi, Guo,\
  \BBA\ Blundell}{Badia et~al.}{2020}]{agent57}
Badia, A.~P., Piot, B., Kapturowski, S., Sprechmann, P., Vitvitskyi, A., Guo,
  Z.~D., \BBA\ Blundell, C. \BBOP2020\BBCP.
\newblock \BBOQ Agent57: Outperforming the atari human benchmark\BBCQ\
\newblock In {\Bem Proceedings of the 37th International Conference on Machine
  Learning, {ICML}, 13-18 July, Virtual Event}, \lowercase{\BVOL}\ 119 of {\Bem
  Proceedings of Machine Learning Research}, \BPGS\ 507--517. {PMLR}.

\bibitem[\protect\BCAY{Balandat, Karrer, Jiang, Daulton, Letham, Wilson,\ \BBA\
  Bakshy}{Balandat et~al.}{2020}]{botorch}
Balandat, M., Karrer, B., Jiang, D.~R., Daulton, S., Letham, B., Wilson, A.~G.,
  \BBA\ Bakshy, E. \BBOP2020\BBCP.
\newblock \BBOQ Botorch: {A} framework for efficient monte-carlo bayesian
  optimization\BBCQ\
\newblock In {\Bem Advances in Neural Information Processing Systems 33:
  NeurIPS, December 6-12, virtual}.

\bibitem[\protect\BCAY{Ball, Parker-Holder, Pacchiano, Choromanski,\ \BBA\
  Roberts}{Ball et~al.}{2020}]{rp1}
Ball, P., Parker-Holder, J., Pacchiano, A., Choromanski, K., \BBA\ Roberts, S.
  \BBOP2020\BBCP.
\newblock \BBOQ Ready policy one: World building through active learning\BBCQ\
\newblock In {\Bem Proceedings of the 37th International Conference on Machine
  Learning, {ICML}}.

\bibitem[\protect\BCAY{Bamford, Huang,\ \BBA\ Lucas}{Bamford
  et~al.}{2020}]{griddly}
Bamford, C., Huang, S., \BBA\ Lucas, S.~M. \BBOP2020\BBCP.
\newblock \BBOQ Griddly: {A} platform for {AI} research in games\BBCQ\
\newblock {\Bem CoRR}, {\Bem abs/2011.06363}.

\bibitem[\protect\BCAY{Barto\ \BBA\ Sutton}{Barto\ \BBA\
  Sutton}{1981}]{barto1981goal}
Barto, A.~G.\BBACOMMA\  \BBA\ Sutton, R.~S. \BBOP1981\BBCP.
\newblock \BBOQ Goal seeking components for adaptive intelligence: An initial
  assessment.\BBCQ\
\newblock \BTR, Massachusetts Univ Amherst Dept of Computer and Information
  Science.

\bibitem[\protect\BCAY{Bas{-}Serrano, Curi, Krause,\ \BBA\ Neu}{Bas{-}Serrano
  et~al.}{2021}]{logistic-q}
Bas{-}Serrano, J., Curi, S., Krause, A., \BBA\ Neu, G. \BBOP2021\BBCP.
\newblock \BBOQ Logistic q-learning\BBCQ\
\newblock In {\Bem The 24th International Conference on Artificial Intelligence
  and Statistics, {AISTATS}, April 13-15, Virtual Event}, \lowercase{\BVOL}\
  130 of {\Bem Proceedings of Machine Learning Research}, \BPGS\ 3610--3618.
  {PMLR}.

\bibitem[\protect\BCAY{Bechtle, Molchanov, Chebotar, Grefenstette, Righetti,
  Sukhatme,\ \BBA\ Meier}{Bechtle et~al.}{2020}]{ml3}
Bechtle, S., Molchanov, A., Chebotar, Y., Grefenstette, E., Righetti, L.,
  Sukhatme, G.~S., \BBA\ Meier, F. \BBOP2020\BBCP.
\newblock \BBOQ Meta learning via learned loss\BBCQ\
\newblock In {\Bem 25th International Conference on Pattern Recognition,
  {ICPR}, Virtual Event / Milan, Italy, January 10-15, 2021}, \BPGS\
  4161--4168. {IEEE}.

\bibitem[\protect\BCAY{Bellemare, Candido, Castro, Gong, Machado, Moitra,
  Ponda,\ \BBA\ Wang}{Bellemare et~al.}{2020}]{loon}
Bellemare, M., Candido, S., Castro, P., Gong, J., Machado, M., Moitra, S.,
  Ponda, S., \BBA\ Wang, Z. \BBOP2020\BBCP.
\newblock \BBOQ Autonomous navigation of stratospheric balloons using
  reinforcement learning\BBCQ\
\newblock {\Bem Nature}, {\Bem 588}, 77--82.

\bibitem[\protect\BCAY{Bellemare, Dabney,\ \BBA\ Munos}{Bellemare
  et~al.}{2017}]{distributional_og}
Bellemare, M.~G., Dabney, W., \BBA\ Munos, R. \BBOP2017\BBCP.
\newblock \BBOQ A distributional perspective on reinforcement learning\BBCQ\
\newblock In {\Bem Proceedings of the 34th International Conference on Machine
  Learning - Volume 70}, ICML'17, \BPG\ 449–458. JMLR.org.

\bibitem[\protect\BCAY{Bellemare, Naddaf, Veness,\ \BBA\ Bowling}{Bellemare
  et~al.}{2012}]{ale}
Bellemare, M.~G., Naddaf, Y., Veness, J., \BBA\ Bowling, M. \BBOP2012\BBCP.
\newblock \BBOQ The {A}rcade {L}earning {E}nvironment: {A}n {E}valuation
  {P}latform for {G}eneral {A}gents\BBCQ\
\newblock {\Bem CoRR}, {\Bem abs/1207.4708}.

\bibitem[\protect\BCAY{Benjamins, Eimer, Schubert, Biedenkapp, Rosenhahn,
  Hutter,\ \BBA\ Lindauer}{Benjamins et~al.}{2021}]{BenEim2021a}
Benjamins, C., Eimer, T., Schubert, F., Biedenkapp, A., Rosenhahn, B., Hutter,
  F., \BBA\ Lindauer, M. \BBOP2021\BBCP.
\newblock \BBOQ Carl: A benchmark for contextual and adaptive reinforcement
  learning\BBCQ\
\newblock In {\Bem NeurIPS 2021 Workshop on Ecological Theory of Reinforcement
  Learning}.

\bibitem[\protect\BCAY{Bergstra\ \BBA\ Bengio}{Bergstra\ \BBA\
  Bengio}{2012}]{randomsearch}
Bergstra, J.\BBACOMMA\  \BBA\ Bengio, Y. \BBOP2012\BBCP.
\newblock \BBOQ Random search for hyper-parameter optimization\BBCQ\
\newblock In {\Bem Journal of Machine Learning Research}.

\bibitem[\protect\BCAY{Berner, Brockman, Chan, Cheung, Debiak, Dennison, Farhi,
  Fischer, Hashme, Hesse, J{\'{o}}zefowicz, Gray, Olsson, Pachocki, Petrov,
  de~Oliveira~Pinto, Raiman, Salimans, Schlatter, Schneider, Sidor, Sutskever,
  Tang, Wolski,\ \BBA\ Zhang}{Berner et~al.}{2019}]{dota}
Berner, C., Brockman, G., Chan, B., Cheung, V., Debiak, P., Dennison, C.,
  Farhi, D., Fischer, Q., Hashme, S., Hesse, C., J{\'{o}}zefowicz, R., Gray,
  S., Olsson, C., Pachocki, J., Petrov, M., de~Oliveira~Pinto, H.~P., Raiman,
  J., Salimans, T., Schlatter, J., Schneider, J., Sidor, S., Sutskever, I.,
  Tang, J., Wolski, F., \BBA\ Zhang, S. \BBOP2019\BBCP.
\newblock \BBOQ Dota 2 with large scale deep reinforcement learning\BBCQ\
\newblock {\Bem CoRR}, {\Bem abs/1912.06680}.

\bibitem[\protect\BCAY{Bertsekas\ \BBA\ Tsitsiklis}{Bertsekas\ \BBA\
  Tsitsiklis}{1996}]{neurodynamic}
Bertsekas, D.\BBACOMMA\  \BBA\ Tsitsiklis, J. \BBOP1996\BBCP.
\newblock {\Bem Neuro-Dynamic Programming}, \lowercase{\BVOL}~27.

\bibitem[\protect\BCAY{Biedenkapp, Lindauer, Eggensperger, Fawcett, Hoos,\
  \BBA\ Hutter}{Biedenkapp et~al.}{2017}]{biedenkapp-aaai17}
Biedenkapp, A., Lindauer, M., Eggensperger, K., Fawcett, C., Hoos, H.~H., \BBA\
  Hutter, F. \BBOP2017\BBCP.
\newblock \BBOQ Efficient parameter importance analysis via ablation with
  surrogates\BBCQ\
\newblock In {\Bem Proceedings of the Conference on Artificial Intelligence
  ({AAAI}'17)}, \BPGS\ 773--779. {AAAI} Press.

\bibitem[\protect\BCAY{Biedenkapp, Marben, Lindauer,\ \BBA\ Hutter}{Biedenkapp
  et~al.}{2018}]{biedenkapp-lion18a}
Biedenkapp, A., Marben, J., Lindauer, M., \BBA\ Hutter, F. \BBOP2018\BBCP.
\newblock \BBOQ {CAVE}: Configuration assessment, visualization and
  evaluation\BBCQ\
\newblock In {\Bem Proceedings of the International Conference on Learning and
  Intelligent Optimization ({LION})}. Springer.

\bibitem[\protect\BCAY{Biedenkapp, Rajan, Hutter,\ \BBA\ Lindauer}{Biedenkapp
  et~al.}{2021}]{temporl}
Biedenkapp, A., Rajan, R., Hutter, F., \BBA\ Lindauer, M. \BBOP2021\BBCP.
\newblock \BBOQ Tempo{RL}: Learning when to act\BBCQ\
\newblock In {\Bem Proceedings of the 38th International Conference on Machine
  Learning ({ICML}'21)}, \lowercase{\BVOL}\ 139 of {\Bem Proceedings of Machine
  Learning Research}, \BPGS\ 914--924. PMLR.

\bibitem[\protect\BCAY{Bontrager\ \BBA\ Togelius}{Bontrager\ \BBA\
  Togelius}{2020}]{gpn_bontrager_arxiv_20}
Bontrager, P.\BBACOMMA\  \BBA\ Togelius, J. \BBOP2020\BBCP.
\newblock \BBOQ Fully differentiable procedural content generation through
  generative playing networks\BBCQ\
\newblock {\Bem CoRR}, {\Bem abs/2002.05259}.

\bibitem[\protect\BCAY{Bradbury, Frostig, Hawkins, Johnson, Leary, Maclaurin,
  Necula, Paszke, Vander{P}las, Wanderman-{M}ilne,\ \BBA\ Zhang}{Bradbury
  et~al.}{2018}]{jax2018github}
Bradbury, J., Frostig, R., Hawkins, P., Johnson, M.~J., Leary, C., Maclaurin,
  D., Necula, G., Paszke, A., Vander{P}las, J., Wanderman-{M}ilne, S., \BBA\
  Zhang, Q. \BBOP2018\BBCP.
\newblock \BBOQ {JAX}: composable transformations of {P}ython+{N}um{P}y
  programs\BBCQ.

\bibitem[\protect\BCAY{Brochu, Cora,\ \BBA\ de~Freitas}{Brochu
  et~al.}{2010}]{bayesopt_nando}
Brochu, E., Cora, V.~M., \BBA\ de~Freitas, N. \BBOP2010\BBCP.
\newblock \BBOQ A tutorial on {B}ayesian optimization of expensive cost
  functions, with application to active user modeling and hierarchical
  reinforcement learning\BBCQ\
\newblock {\Bem CoRR}, {\Bem abs/1012.2599}.

\bibitem[\protect\BCAY{Brockman, Cheung, Pettersson, Schneider, Schulman,
  Tang,\ \BBA\ Zaremba}{Brockman et~al.}{2016}]{gym}
Brockman, G., Cheung, V., Pettersson, L., Schneider, J., Schulman, J., Tang,
  J., \BBA\ Zaremba, W. \BBOP2016\BBCP.
\newblock \BBOQ Openai gym\BBCQ.

\bibitem[\protect\BCAY{Browne, Powley, Whitehouse, Lucas, Cowling, Rohlfshagen,
  Tavener, Liebana, Samothrakis,\ \BBA\ Colton}{Browne
  et~al.}{2012}]{mcts_browne_ieee_12}
Browne, C., Powley, E.~J., Whitehouse, D., Lucas, S.~M., Cowling, P.~I.,
  Rohlfshagen, P., Tavener, S., Liebana, D.~P., Samothrakis, S., \BBA\ Colton,
  S. \BBOP2012\BBCP.
\newblock \BBOQ A survey of monte carlo tree search methods\BBCQ\
\newblock {\Bem {IEEE} Trans. Comput. Intell. {AI} Games}, {\Bem 4\/}(1),
  1--43.

\bibitem[\protect\BCAY{Campero, Raileanu, Kuttler, Tenenbaum, Rockt{\"a}schel,\
  \BBA\ Grefenstette}{Campero et~al.}{2021}]{campero2021learning}
Campero, A., Raileanu, R., Kuttler, H., Tenenbaum, J.~B., Rockt{\"a}schel, T.,
  \BBA\ Grefenstette, E. \BBOP2021\BBCP.
\newblock \BBOQ Learning with {\{}amig{\}}o: Adversarially motivated intrinsic
  goals\BBCQ\
\newblock In {\Bem International Conference on Learning Representations}.

\bibitem[\protect\BCAY{Chen, Hoffman, Colmenarejo, Denil, Lillicrap,
  Botvinick,\ \BBA\ de~Freitas}{Chen et~al.}{2017}]{l2lwGDbGD_chen_icml_17}
Chen, Y., Hoffman, M.~W., Colmenarejo, S.~G., Denil, M., Lillicrap, T.~P.,
  Botvinick, M., \BBA\ de~Freitas, N. \BBOP2017\BBCP.
\newblock \BBOQ Learning to learn without gradient descent by gradient
  descent\BBCQ\
\newblock In Precup, D.\BBACOMMA\  \BBA\ Teh, Y.~W.\BEDS, {\Bem Proceedings of
  the 34th International Conference on Machine Learning, {ICML} 2017, Sydney,
  NSW, Australia, 6-11 August 2017}, \lowercase{\BVOL}~70 of {\Bem Proceedings
  of Machine Learning Research}, \BPGS\ 748--756. {PMLR}.

\bibitem[\protect\BCAY{Chen, Huang, Wang, Antonoglou, Schrittwieser, Silver,\
  \BBA\ de~Freitas}{Chen et~al.}{2018}]{bo_alphago}
Chen, Y., Huang, A., Wang, Z., Antonoglou, I., Schrittwieser, J., Silver, D.,
  \BBA\ de~Freitas, N. \BBOP2018\BBCP.
\newblock \BBOQ {B}ayesian optimization in {AlphaGo}\BBCQ\
\newblock {\Bem CoRR}, {\Bem abs/1812.06855}.

\bibitem[\protect\BCAY{Chevalier-Boisvert, Willems,\ \BBA\
  Pal}{Chevalier-Boisvert et~al.}{2018}]{gym_minigrid}
Chevalier-Boisvert, M., Willems, L., \BBA\ Pal, S. \BBOP2018\BBCP.
\newblock \BBOQ Minimalistic gridworld environment for openai gym\BBCQ.

\bibitem[\protect\BCAY{Chiang, Faust, Fiser,\ \BBA\ Francis}{Chiang
  et~al.}{2019}]{autorl_learn_reward_nn_arch_chiang_ral19}
Chiang, H.~L., Faust, A., Fiser, M., \BBA\ Francis, A.~G. \BBOP2019\BBCP.
\newblock \BBOQ Learning navigation behaviors end-to-end with autorl\BBCQ\
\newblock {\Bem {IEEE} Robotics Autom. Lett.}, {\Bem 4\/}(2), 2007--2014.

\bibitem[\protect\BCAY{Cho, Kim, Lee, Choi, Lee,\ \BBA\ Rhee}{Cho
  et~al.}{2020}]{expected_time_for_performance}
Cho, H., Kim, Y., Lee, E., Choi, D., Lee, Y., \BBA\ Rhee, W. \BBOP2020\BBCP.
\newblock \BBOQ Basic enhancement strategies when using bayesian optimization
  for hyperparameter tuning of deep neural networks\BBCQ\
\newblock {\Bem {IEEE} Access}, {\Bem 8}, 52588--52608.

\bibitem[\protect\BCAY{Chrabąszcz, Loshchilov,\ \BBA\ Hutter}{Chrabąszcz
  et~al.}{2018}]{chrabaszcz-ijcai18a}
Chrabąszcz, P., Loshchilov, I., \BBA\ Hutter, F. \BBOP2018\BBCP.
\newblock \BBOQ Back to basics: Benchmarking canonical evolution strategies for
  playing atari\BBCQ\
\newblock In {\Bem Proceedings of the Twenty-Seventh International Joint
  Conference on Artificial Intelligence, IJCAI}, \BPGS\ 1419--1426.
  International Joint Conferences on Artificial Intelligence Organization.

\bibitem[\protect\BCAY{Chua, Calandra, McAllister,\ \BBA\ Levine}{Chua
  et~al.}{2018}]{pets}
Chua, K., Calandra, R., McAllister, R., \BBA\ Levine, S. \BBOP2018\BBCP.
\newblock \BBOQ Deep reinforcement learning in a handful of trials using
  probabilistic dynamics models\BBCQ\
\newblock In {\Bem Advances in Neural Information Processing Systems 31},
  \BPGS\ 4754--4765.

\bibitem[\protect\BCAY{Clune, Beckmann, Ofria,\ \BBA\ Pennock}{Clune
  et~al.}{2009}]{hyperneat2009gaits}
Clune, J., Beckmann, B.~E., Ofria, C., \BBA\ Pennock, R.~T. \BBOP2009\BBCP.
\newblock \BBOQ Evolving coordinated quadruped gaits with the hyperneat
  generative encoding\BBCQ\
\newblock In {\Bem Proceedings of the Eleventh Conference on Congress on
  Evolutionary Computation}, CEC'09, \BPG\ 2764–2771. IEEE Press.

\bibitem[\protect\BCAY{Clune, Misevic, Ofria, Lenski, Elena,\ \BBA\
  Sanjuán}{Clune et~al.}{2008}]{Clune2008EA}
Clune, J., Misevic, D., Ofria, C., Lenski, R.~E., Elena, S.~F., \BBA\ Sanjuán,
  R. \BBOP2008\BBCP.
\newblock \BBOQ Natural selection fails to optimize mutation rates for
  long-term adaptation on rugged fitness landscapes\BBCQ\
\newblock {\Bem PLOS Computational Biology}, {\Bem 4}, 1--8.

\bibitem[\protect\BCAY{Co-Reyes, Miao, Peng, Le, Levine, Lee,\ \BBA\
  Faust}{Co-Reyes et~al.}{2021}]{evolving_algos}
Co-Reyes, J.~D., Miao, Y., Peng, D., Le, Q.~V., Levine, S., Lee, H., \BBA\
  Faust, A. \BBOP2021\BBCP.
\newblock \BBOQ Evolving reinforcement learning algorithms\BBCQ\
\newblock In {\Bem International Conference on Learning Representations}.

\bibitem[\protect\BCAY{Cobbe, Hesse, Hilton,\ \BBA\ Schulman}{Cobbe
  et~al.}{2020}]{procgen}
Cobbe, K., Hesse, C., Hilton, J., \BBA\ Schulman, J. \BBOP2020\BBCP.
\newblock \BBOQ Leveraging procedural generation to benchmark reinforcement
  learning\BBCQ\
\newblock In {\Bem Proceedings of the 37th International Conference on Machine
  Learning, {ICML}, 13-18 July, Virtual Event}, \lowercase{\BVOL}\ 119 of {\Bem
  Proceedings of Machine Learning Research}, \BPGS\ 2048--2056. {PMLR}.

\bibitem[\protect\BCAY{Cobbe, Klimov, Hesse, Kim,\ \BBA\ Schulman}{Cobbe
  et~al.}{2019a}]{pmlr-v97-cobbe19a}
Cobbe, K., Klimov, O., Hesse, C., Kim, T., \BBA\ Schulman, J. \BBOP2019a\BBCP.
\newblock \BBOQ Quantifying generalization in reinforcement learning\BBCQ\
\newblock In Chaudhuri, K.\BBACOMMA\  \BBA\ Salakhutdinov, R.\BEDS, {\Bem
  Proceedings of the 36th International Conference on Machine Learning},
  \lowercase{\BVOL}~97 of {\Bem Proceedings of Machine Learning Research},
  \BPGS\ 1282--1289. PMLR.

\bibitem[\protect\BCAY{Cobbe, Klimov, Hesse, Kim,\ \BBA\ Schulman}{Cobbe
  et~al.}{2019b}]{coinrun}
Cobbe, K., Klimov, O., Hesse, C., Kim, T., \BBA\ Schulman, J. \BBOP2019b\BBCP.
\newblock \BBOQ Quantifying generalization in reinforcement learning\BBCQ\
\newblock In {\Bem Proceedings of the 36th International Conference on Machine
  Learning, {ICML}, 9-15 June, Long Beach, California, {USA}},
  \lowercase{\BVOL}~97 of {\Bem Proceedings of Machine Learning Research},
  \BPGS\ 1282--1289. {PMLR}.

\bibitem[\protect\BCAY{Colas, Sigaud,\ \BBA\ Oudeyer}{Colas
  et~al.}{2018}]{how_many_rl_seeds}
Colas, C., Sigaud, O., \BBA\ Oudeyer, P. \BBOP2018\BBCP.
\newblock \BBOQ How many random seeds? statistical power analysis in deep
  reinforcement learning experiments\BBCQ\
\newblock {\Bem CoRR}, {\Bem abs/1806.08295}.

\bibitem[\protect\BCAY{Cutler, Walsh,\ \BBA\ How}{Cutler
  et~al.}{2014}]{cutler2014multifidelity_rl}
Cutler, M., Walsh, T.~J., \BBA\ How, J.~P. \BBOP2014\BBCP.
\newblock \BBOQ Reinforcement learning with multi-fidelity simulators\BBCQ\
\newblock In {\Bem 2014 IEEE International Conference on Robotics and
  Automation (ICRA)}, \BPGS\ 3888--3895.

\bibitem[\protect\BCAY{da~Silva, Costa,\ \BBA\ Stone}{da~Silva
  et~al.}{2019}]{silva-ieee19}
da~Silva, F.~L., Costa, A. H.~R., \BBA\ Stone, P. \BBOP2019\BBCP.
\newblock \BBOQ Building self-play curricula online by playing with expert
  agents in adversarial games\BBCQ\
\newblock In {\Bem 8th Brazilian Conference on Intelligent Systems, {BRACIS},
  Salvador, Brazil, October 15-18}, \BPGS\ 479--484. {IEEE}.

\bibitem[\protect\BCAY{Dabney, Ostrovski,\ \BBA\ Barreto}{Dabney
  et~al.}{2021}]{dabney_temporally_extended_e_greedy_iclr_21}
Dabney, W., Ostrovski, G., \BBA\ Barreto, A. \BBOP2021\BBCP.
\newblock \BBOQ Temporally-extended {\(\epsilon\)}-greedy exploration\BBCQ\
\newblock In {\Bem 9th International Conference on Learning Representations,
  {ICLR} 2021, Virtual Event, Austria, May 3-7, 2021}. OpenReview.net.

\bibitem[\protect\BCAY{Dalibard\ \BBA\ Jaderberg}{Dalibard\ \BBA\
  Jaderberg}{2021}]{dalibard2021faster}
Dalibard, V.\BBACOMMA\  \BBA\ Jaderberg, M. \BBOP2021\BBCP.
\newblock \BBOQ Faster improvement rate population based training\BBCQ.

\bibitem[\protect\BCAY{Degrave, Felici, Buchli, Neunert, Tracey, Carpanese,
  Ewalds, Hafner, Abdolmaleki, de~las Casas, Donner, Fritz, Galperti, Huber,
  Keeling, Tsimpoukelli, Kay, Merle, Moret, Noury, Pesamosca, Pfau, Sauter,
  Sommariva, Coda, Duval, Fasoli, Kohli, Kavukcuoglu, Hassabis,\ \BBA\
  Riedmiller}{Degrave et~al.}{2022}]{rl_plasma}
Degrave, J., Felici, F., Buchli, J., Neunert, M., Tracey, B., Carpanese, F.,
  Ewalds, T., Hafner, R., Abdolmaleki, A., de~las Casas, D., Donner, C., Fritz,
  L., Galperti, C., Huber, A., Keeling, J., Tsimpoukelli, M., Kay, J., Merle,
  A., Moret, J.-M., Noury, S., Pesamosca, F., Pfau, D., Sauter, O., Sommariva,
  C., Coda, S., Duval, B., Fasoli, A., Kohli, P., Kavukcuoglu, K., Hassabis,
  D., \BBA\ Riedmiller, M. \BBOP2022\BBCP.
\newblock \BBOQ Magnetic control of tokamak plasmas through deep reinforcement
  learning\BBCQ\
\newblock {\Bem Nature}, {\Bem 602}, 414--419.

\bibitem[\protect\BCAY{Dennis, Jaques, Vinitsky, Bayen, Russell, Critch,\ \BBA\
  Levine}{Dennis et~al.}{2020}]{paired_dennis_nips_20}
Dennis, M., Jaques, N., Vinitsky, E., Bayen, A.~M., Russell, S., Critch, A.,
  \BBA\ Levine, S. \BBOP2020\BBCP.
\newblock \BBOQ Emergent complexity and zero-shot transfer via unsupervised
  environment design\BBCQ\
\newblock In {\Bem Advances in Neural Information Processing Systems 33:
  December 6-12, virtual}.

\bibitem[\protect\BCAY{Doerr\ \BBA\ Doerr}{Doerr\ \BBA\
  Doerr}{2020}]{doerr-toec20}
Doerr, B.\BBACOMMA\  \BBA\ Doerr, C. \BBOP2020\BBCP.
\newblock \BBOQ Theory of parameter control for discrete black-box
  optimization: Provable performance gains through dynamic parameter
  choices\BBCQ\
\newblock In {\Bem Theory of Evolutionary Computation}, \BPGS\ 271--321.
  Springer.

\bibitem[\protect\BCAY{Downey\ \BBA\ Sanner}{Downey\ \BBA\
  Sanner}{2010}]{tdbma}
Downey, C.\BBACOMMA\  \BBA\ Sanner, S. \BBOP2010\BBCP.
\newblock \BBOQ Temporal difference bayesian model averaging: A bayesian
  perspective on adapting lambda\BBCQ\
\newblock In {\Bem ICML}, \BPGS\ 311--318.

\bibitem[\protect\BCAY{Duan, Schulman, Chen, Bartlett, Sutskever,\ \BBA\
  Abbeel}{Duan et~al.}{2016}]{duan-rl2-arxiv16}
Duan, Y., Schulman, J., Chen, X., Bartlett, P.~L., Sutskever, I., \BBA\ Abbeel,
  P. \BBOP2016\BBCP.
\newblock \BBOQ Rl{\textdollar}{\^{}}2{\textdollar}: Fast reinforcement
  learning via slow reinforcement learning\BBCQ\
\newblock {\Bem CoRR}, {\Bem abs/1611.02779}.

\bibitem[\protect\BCAY{Eggensperger, Lindauer,\ \BBA\ Hutter}{Eggensperger
  et~al.}{2019}]{eggensperger-jair19a}
Eggensperger, K., Lindauer, M., \BBA\ Hutter, F. \BBOP2019\BBCP.
\newblock \BBOQ Pitfalls and best practices in algorithm configuration\BBCQ\
\newblock {\Bem Journal of Artificial Intelligence Research (JAIR)}, {\Bem 64},
  861--893.

\bibitem[\protect\BCAY{Eimer, Benjamins,\ \BBA\ Lindauer}{Eimer
  et~al.}{2021a}]{EimBen2021a}
Eimer, T., Benjamins, C., \BBA\ Lindauer, M. \BBOP2021a\BBCP.
\newblock \BBOQ Hyperparameters in contextual rl are highly situational\BBCQ\
\newblock In {\Bem NeurIPS 2021 Workshop on Ecological Theory of Reinforcement
  Learning}.

\bibitem[\protect\BCAY{Eimer, Biedenkapp, Hutter,\ \BBA\ Lindauer}{Eimer
  et~al.}{2021b}]{space_eimer_icml_21}
Eimer, T., Biedenkapp, A., Hutter, F., \BBA\ Lindauer, M. \BBOP2021b\BBCP.
\newblock \BBOQ Self-paced context evaluation for contextual reinforcement
  learning\BBCQ\
\newblock In {\Bem Proceedings of the 38th International Conference on Machine
  Learning ({ICML}'21)}, \lowercase{\BVOL}\ 139 of {\Bem Proceedings of Machine
  Learning Research}, \BPGS\ 2948--2958. PMLR.

\bibitem[\protect\BCAY{Elfwing, Uchibe,\ \BBA\ Doya}{Elfwing
  et~al.}{2017}]{elfwing-arxiv17}
Elfwing, S., Uchibe, E., \BBA\ Doya, K. \BBOP2017\BBCP.
\newblock \BBOQ Online meta-learning by parallel algorithm competition\BBCQ\
\newblock {\Bem CoRR}, {\Bem abs/1702.07490}.

\bibitem[\protect\BCAY{Elfwing, Uchibe,\ \BBA\ Doya}{Elfwing
  et~al.}{2018}]{elfwing-gecco18}
Elfwing, S., Uchibe, E., \BBA\ Doya, K. \BBOP2018\BBCP.
\newblock \BBOQ Online meta-learning by parallel algorithm competition\BBCQ\
\newblock In Aguirre, H.~E.\BBACOMMA\  \BBA\ Takadama, K.\BEDS, {\Bem
  Proceedings of the Genetic and Evolutionary Computation Conference, {GECCO}
  2018, Kyoto, Japan, July 15-19, 2018}, \BPGS\ 426--433. {ACM}.

\bibitem[\protect\BCAY{Elsken, Metzen,\ \BBA\ Hutter}{Elsken
  et~al.}{2019}]{ElskenMH19}
Elsken, T., Metzen, J.~H., \BBA\ Hutter, F. \BBOP2019\BBCP.
\newblock \BBOQ Neural architecture search: {A} survey\BBCQ\
\newblock {\Bem J. Mach. Learn. Res.}, {\Bem 20}, 55:1--55:21.

\bibitem[\protect\BCAY{Engstrom, Ilyas, Santurkar, Tsipras, Janoos, Rudolph,\
  \BBA\ Madry}{Engstrom et~al.}{2020}]{Engstrom2020Implementation}
Engstrom, L., Ilyas, A., Santurkar, S., Tsipras, D., Janoos, F., Rudolph, L.,
  \BBA\ Madry, A. \BBOP2020\BBCP.
\newblock \BBOQ Implementation matters in deep {RL:} {A} case study on {PPO}
  and {TRPO}\BBCQ\
\newblock In {\Bem 8th International Conference on Learning Representations,
  {ICLR}, Addis Ababa, Ethiopia, April 26-30}. OpenReview.net.

\bibitem[\protect\BCAY{Eriksson, Capi,\ \BBA\ Doya}{Eriksson
  et~al.}{2003}]{hpoGA}
Eriksson, A., Capi, G., \BBA\ Doya, K. \BBOP2003\BBCP.
\newblock \BBOQ Evolution of meta-parameters in reinforcement learning
  algorithm\BBCQ\
\newblock In {\Bem Proceedings IEEE/RSJ International Conference on Intelligent
  Robots and Systems (IROS) (Cat. No.03CH37453)}, \lowercase{\BVOL}~1, \BPGS\
  412--417 vol.1.

\bibitem[\protect\BCAY{Espeholt, Soyer, Munos, Simonyan, Mnih, Ward, Doron,
  Firoiu, Harley, Dunning, Legg,\ \BBA\ Kavukcuoglu}{Espeholt
  et~al.}{2018}]{impala}
Espeholt, L., Soyer, H., Munos, R., Simonyan, K., Mnih, V., Ward, T., Doron,
  Y., Firoiu, V., Harley, T., Dunning, I., Legg, S., \BBA\ Kavukcuoglu, K.
  \BBOP2018\BBCP.
\newblock \BBOQ {IMPALA:} scalable distributed deep-rl with importance weighted
  actor-learner architectures\BBCQ\
\newblock In {\Bem Proceedings of the 35th International Conference on Machine
  Learning, {ICML}, Stockholmsm{\"{a}}ssan, Stockholm, Sweden, July 10-15,
  2018}, Proceedings of Machine Learning Research, \BPGS\ 1406--1415. {PMLR}.

\bibitem[\protect\BCAY{Falkner, Klein,\ \BBA\ Hutter}{Falkner
  et~al.}{2018}]{falkner2018bohb}
Falkner, S., Klein, A., \BBA\ Hutter, F. \BBOP2018\BBCP.
\newblock \BBOQ Bohb: Robust and efficient hyperparameter optimization at
  scale\BBCQ\
\newblock In {\Bem International Conference on Machine Learning}, \BPGS\
  1437--1446. PMLR.

\bibitem[\protect\BCAY{Fang, Zhu, Savarese,\ \BBA\ Fei-Fei}{Fang
  et~al.}{2021}]{fang2021adaptive}
Fang, K., Zhu, Y., Savarese, S., \BBA\ Fei-Fei, L. \BBOP2021\BBCP.
\newblock \BBOQ Adaptive procedural task generation for hard-exploration
  problems\BBCQ\
\newblock In {\Bem International Conference on Learning Representations}.

\bibitem[\protect\BCAY{Farahani\ \BBA\ Mozayani}{Farahani\ \BBA\
  Mozayani}{2019}]{macro_actions}
Farahani, M.~D.\BBACOMMA\  \BBA\ Mozayani, N. \BBOP2019\BBCP.
\newblock \BBOQ Automatic construction and evaluation of macro-actions in
  reinforcement learning\BBCQ\
\newblock {\Bem Appl. Soft Comput.}, {\Bem 82}.

\bibitem[\protect\BCAY{Farquhar, Gustafson, Lin, Whiteson, Usunier,\ \BBA\
  Synnaeve}{Farquhar et~al.}{2020}]{farquhar2020growing}
Farquhar, G., Gustafson, L., Lin, Z., Whiteson, S., Usunier, N., \BBA\
  Synnaeve, G. \BBOP2020\BBCP.
\newblock \BBOQ Growing action spaces\BBCQ\
\newblock In {\Bem Proceedings of the 37th International Conference on Machine
  Learning, {ICML}, 13-18 July, Virtual Event}, \lowercase{\BVOL}\ 119, \BPGS\
  3040--3051. {PMLR}.

\bibitem[\protect\BCAY{Faust, Francis,\ \BBA\ Mehta}{Faust
  et~al.}{2019}]{faust2019evolving}
Faust, A., Francis, A., \BBA\ Mehta, D. \BBOP2019\BBCP.
\newblock \BBOQ Evolving rewards to automate reinforcement learning\BBCQ\
\newblock In {\Bem AutoML workshop at 7th International Conference on Learning
  Representation}.

\bibitem[\protect\BCAY{Fawcett\ \BBA\ Hoos}{Fawcett\ \BBA\
  Hoos}{2016}]{fawcett-heu16a}
Fawcett, C.\BBACOMMA\  \BBA\ Hoos, H.~H. \BBOP2016\BBCP.
\newblock \BBOQ Analysing differences between algorithm configurations through
  ablation\BBCQ\
\newblock {\Bem Journal of Heuristics}, {\Bem 22\/}(4), 431--458.

\bibitem[\protect\BCAY{Fedus, Ramachandran, Agarwal, Bengio, Larochelle,
  Rowland,\ \BBA\ Dabney}{Fedus et~al.}{2020}]{Fedus2020RevisitingFO}
Fedus, W., Ramachandran, P., Agarwal, R., Bengio, Y., Larochelle, H., Rowland,
  M., \BBA\ Dabney, W. \BBOP2020\BBCP.
\newblock \BBOQ Revisiting fundamentals of experience replay\BBCQ\
\newblock In {\Bem ICML}.

\bibitem[\protect\BCAY{Fernandez\ \BBA\ Caarls}{Fernandez\ \BBA\
  Caarls}{2018}]{hpoeo}
Fernandez, F.~C.\BBACOMMA\  \BBA\ Caarls, W. \BBOP2018\BBCP.
\newblock \BBOQ Parameters tuning and optimization for reinforcement learning
  algorithms using evolutionary computing\BBCQ\
\newblock In {\Bem 2018 International Conference on Information Systems and
  Computer Science (INCISCOS)}, \BPGS\ 301--305.

\bibitem[\protect\BCAY{Ferreira, Nierhoff,\ \BBA\ Hutter}{Ferreira
  et~al.}{2021}]{synth_envs_ferreira_metalearn_21a}
Ferreira, F., Nierhoff, T., \BBA\ Hutter, F. \BBOP2021\BBCP.
\newblock \BBOQ Learning synthetic environments for reinforcement learning with
  evolution strategies\BBCQ.

\bibitem[\protect\BCAY{Finn, Abbeel,\ \BBA\ Levine}{Finn
  et~al.}{2017}]{maml_finn_icml17}
Finn, C., Abbeel, P., \BBA\ Levine, S. \BBOP2017\BBCP.
\newblock \BBOQ Model-agnostic meta-learning for fast adaptation of deep
  networks\BBCQ\
\newblock In {\Bem Proceedings of the 34th International Conference on Machine
  Learning, {ICML}, Sydney, NSW, Australia, 6-11 August}, \lowercase{\BVOL}~70
  of {\Bem Proceedings of Machine Learning Research}, \BPGS\ 1126--1135.
  {PMLR}.

\bibitem[\protect\BCAY{Flennerhag, Schroecker, Zahavy, van Hasselt, Silver,\
  \BBA\ Singh}{Flennerhag et~al.}{2021}]{flennerhag2022bootstrapped}
Flennerhag, S., Schroecker, Y., Zahavy, T., van Hasselt, H., Silver, D., \BBA\
  Singh, S. \BBOP2021\BBCP.
\newblock \BBOQ Bootstrapped meta-learning\BBCQ\
\newblock In {\Bem arxiv}.

\bibitem[\protect\BCAY{Florensa, Held, Geng,\ \BBA\ Abbeel}{Florensa
  et~al.}{2018}]{goal_gan}
Florensa, C., Held, D., Geng, X., \BBA\ Abbeel, P. \BBOP2018\BBCP.
\newblock \BBOQ Automatic goal generation for reinforcement learning
  agents\BBCQ\
\newblock In {\Bem Proceedings of the 35th International Conference on Machine
  Learning, {ICML}, Stockholmsm{\"{a}}ssan, Stockholm, Sweden, July 10-15},
  \lowercase{\BVOL}~80 of {\Bem Proceedings of Machine Learning Research},
  \BPGS\ 1514--1523. {PMLR}.

\bibitem[\protect\BCAY{Foerster}{Foerster}{2018}]{jakob2018a}
Foerster, J.~N. \BBOP2018\BBCP.
\newblock {\Bem Deep multi-agent reinforcement learning}.
\newblock Ph.D.\ thesis, University of Oxford.

\bibitem[\protect\BCAY{Fortunato, Azar, Piot, Menick, Hessel, Osband, Graves,
  Mnih, Munos, Hassabis, Pietquin, Blundell,\ \BBA\ Legg}{Fortunato
  et~al.}{2018}]{fortunato2018noisy}
Fortunato, M., Azar, M.~G., Piot, B., Menick, J., Hessel, M., Osband, I.,
  Graves, A., Mnih, V., Munos, R., Hassabis, D., Pietquin, O., Blundell, C.,
  \BBA\ Legg, S. \BBOP2018\BBCP.
\newblock \BBOQ Noisy networks for exploration\BBCQ\
\newblock In {\Bem International Conference on Learning Representations}.

\bibitem[\protect\BCAY{Franke, Koehler, Biedenkapp,\ \BBA\ Hutter}{Franke
  et~al.}{2021}]{franke2021sampleefficient}
Franke, J.~K., Koehler, G., Biedenkapp, A., \BBA\ Hutter, F. \BBOP2021\BBCP.
\newblock \BBOQ Sample-efficient automated deep reinforcement learning\BBCQ\
\newblock In {\Bem International Conference on Learning Representations}.

\bibitem[\protect\BCAY{Frazier\ \BBA\ Wang}{Frazier\ \BBA\
  Wang}{2015}]{materials_bayesopt_Frazier_2015}
Frazier, P.~I.\BBACOMMA\  \BBA\ Wang, J. \BBOP2015\BBCP.
\newblock \BBOQ Bayesian optimization for materials design\BBCQ\
\newblock {\Bem Springer Series in Materials Science}, {\Bem 225}, 45–75.

\bibitem[\protect\BCAY{Freeman, Frey, Raichuk, Girgin, Mordatch,\ \BBA\
  Bachem}{Freeman et~al.}{2021}]{brax}
Freeman, C.~D., Frey, E., Raichuk, A., Girgin, S., Mordatch, I., \BBA\ Bachem,
  O. \BBOP2021\BBCP.
\newblock \BBOQ Brax - a differentiable physics engine for large scale rigid
  body simulation\BBCQ\
\newblock In {\Bem Thirty-fifth Conference on Neural Information Processing
  Systems Datasets and Benchmarks Track (Round 1)}.

\bibitem[\protect\BCAY{Fukunaga}{Fukunaga}{1998}]{RestartGA}
Fukunaga, A.~S. \BBOP1998\BBCP.
\newblock \BBOQ Restart scheduling for genetic algorithms\BBCQ\
\newblock In {\Bem Parallel Problem Solving from Nature --- PPSN V}, \BPGS\
  357--366, Berlin, Heidelberg. Springer Berlin Heidelberg.

\bibitem[\protect\BCAY{Gaier\ \BBA\ Ha}{Gaier\ \BBA\
  Ha}{2019}]{wanns_gaier_neurips19}
Gaier, A.\BBACOMMA\  \BBA\ Ha, D. \BBOP2019\BBCP.
\newblock \BBOQ Weight agnostic neural networks\BBCQ\
\newblock In {\Bem Advances in Neural Information Processing Systems 32: Annual
  Conference on Neural Information Processing Systems, NeurIPS, December 8-14,
  Vancouver, BC, Canada}, \BPGS\ 5365--5379.

\bibitem[\protect\BCAY{Garcia, Prett,\ \BBA\ Morari}{Garcia
  et~al.}{1989}]{mpc_garcia_journals_automatica_89}
Garcia, C.~E., Prett, D.~M., \BBA\ Morari, M. \BBOP1989\BBCP.
\newblock \BBOQ Model predictive control: Theory and practice - {A}
  survey\BBCQ\
\newblock {\Bem Autom.}, {\Bem 25\/}(3), 335--348.

\bibitem[\protect\BCAY{Gleave, Dennis, Legg, Russell,\ \BBA\ Leike}{Gleave
  et~al.}{2021}]{gleave2021quantifying}
Gleave, A., Dennis, M.~D., Legg, S., Russell, S., \BBA\ Leike, J.
  \BBOP2021\BBCP.
\newblock \BBOQ Quantifying differences in reward functions\BBCQ\
\newblock In {\Bem International Conference on Learning Representations}.

\bibitem[\protect\BCAY{Gloger}{Gloger}{2004}]{Gloger2004SelfadaptiveEA}
Gloger, B. \BBOP2004\BBCP.
\newblock \BBOQ Self-adaptive evolutionary algorithms\BBCQ.

\bibitem[\protect\BCAY{Golovin, Solnik, Moitra, Kochanski, Karro,\ \BBA\
  Sculley}{Golovin et~al.}{2017}]{vizier}
Golovin, D., Solnik, B., Moitra, S., Kochanski, G., Karro, J., \BBA\ Sculley,
  D. \BBOP2017\BBCP.
\newblock \BBOQ Google vizier: {A} service for black-box optimization\BBCQ\
\newblock In {\Bem Proceedings of the 23rd {ACM} {SIGKDD} International
  Conference on Knowledge Discovery and Data Mining, Halifax, NS, Canada,
  August 13 - 17}, \BPGS\ 1487--1495. {ACM}.

\bibitem[\protect\BCAY{Gomez, Schmidhuber,\ \BBA\ Miikkulainen}{Gomez
  et~al.}{2006}]{neuroevolutioncontrol}
Gomez, F., Schmidhuber, J., \BBA\ Miikkulainen, R. \BBOP2006\BBCP.
\newblock \BBOQ Efficient non-linear control through neuroevolution\BBCQ\
\newblock In {\Bem Machine Learning: ECML}, \BPGS\ 654--662, Berlin,
  Heidelberg. Springer Berlin Heidelberg.

\bibitem[\protect\BCAY{Griffiths\ \BBA\ Hern{\'a}ndez-Lobato}{Griffiths\ \BBA\
  Hern{\'a}ndez-Lobato}{2020}]{griffiths2020constrained}
Griffiths, R.-R.\BBACOMMA\  \BBA\ Hern{\'a}ndez-Lobato, J.~M. \BBOP2020\BBCP.
\newblock \BBOQ Constrained bayesian optimization for automatic chemical design
  using variational autoencoders\BBCQ\
\newblock {\Bem Chemical science}, {\Bem 11\/}(2), 577--586.

\bibitem[\protect\BCAY{Guo}{Guo}{2020}]{optimization_picture}
Guo, S. \BBOP2020\BBCP.
\newblock \BBOQ An introduction to surrogate optimization: Intuition,
  illustration, case study, and the code\BBCQ.

\bibitem[\protect\BCAY{Gur, Jaques, Miao, Choi, Tiwari, Lee,\ \BBA\ Faust}{Gur
  et~al.}{2021}]{gur2021code}
Gur, I., Jaques, N., Miao, Y., Choi, J., Tiwari, M., Lee, H., \BBA\ Faust, A.
  \BBOP2021\BBCP.
\newblock \BBOQ Environment generation for zero-shot compositional
  reinforcement learning\BBCQ\
\newblock In {\Bem Advances in Neural Information Processing Systems}.

\bibitem[\protect\BCAY{Guss, Codel, Hofmann, Houghton, Kuno, Milani, Mohanty,
  Liebana, Salakhutdinov, Topin, et~al.}{Guss
  et~al.}{2019}]{guss2019minerlcomp}
Guss, W.~H., Codel, C., Hofmann, K., Houghton, B., Kuno, N., Milani, S.,
  Mohanty, S., Liebana, D.~P., Salakhutdinov, R., Topin, N., et~al.
  \BBOP2019\BBCP.
\newblock \BBOQ The {M}ine{RL} competition on sample efficient reinforcement
  learning using human priors\BBCQ.

\bibitem[\protect\BCAY{Ha\ \BBA\ Schmidhuber}{Ha\ \BBA\
  Schmidhuber}{2018}]{worldmodels}
Ha, D.\BBACOMMA\  \BBA\ Schmidhuber, J. \BBOP2018\BBCP.
\newblock \BBOQ Recurrent world models facilitate policy evolution\BBCQ\
\newblock In {\Bem Proceedings of the 32Nd International Conference on Neural
  Information Processing Systems}, NeurIPS'18, \BPGS\ 2455--2467.

\bibitem[\protect\BCAY{Haarnoja, Zhou, Hartikainen, Tucker, Ha, Tan, Kumar,
  Zhu, Gupta, Abbeel,\ \BBA\ Levine}{Haarnoja et~al.}{2018}]{sac-v2}
Haarnoja, T., Zhou, A., Hartikainen, K., Tucker, G., Ha, S., Tan, J., Kumar,
  V., Zhu, H., Gupta, A., Abbeel, P., \BBA\ Levine, S. \BBOP2018\BBCP.
\newblock \BBOQ Soft actor-critic algorithms and applications\BBCQ\
\newblock {\Bem CoRR}, {\Bem abs/1812.05905}.

\bibitem[\protect\BCAY{Hafner, Lillicrap, Ba,\ \BBA\ Norouzi}{Hafner
  et~al.}{2020}]{dreamer}
Hafner, D., Lillicrap, T., Ba, J., \BBA\ Norouzi, M. \BBOP2020\BBCP.
\newblock \BBOQ Dream to control: Learning behaviors by latent
  imagination\BBCQ\
\newblock In {\Bem International Conference on Learning Representations}.

\bibitem[\protect\BCAY{Hafner, Lillicrap, Fischer, Villegas, Ha, Lee,\ \BBA\
  Davidson}{Hafner et~al.}{2019}]{planet}
Hafner, D., Lillicrap, T., Fischer, I., Villegas, R., Ha, D., Lee, H., \BBA\
  Davidson, J. \BBOP2019\BBCP.
\newblock \BBOQ Learning latent dynamics for planning from pixels\BBCQ\
\newblock In {\Bem Proceedings of the 36th International Conference on Machine
  Learning}, \BPGS\ 2555--2565.

\bibitem[\protect\BCAY{Hausknecht, Lehman, Miikkulainen,\ \BBA\
  Stone}{Hausknecht et~al.}{2014}]{hyperneat_atari}
Hausknecht, M., Lehman, J., Miikkulainen, R., \BBA\ Stone, P. \BBOP2014\BBCP.
\newblock \BBOQ A neuroevolution approach to general atari game playing\BBCQ\
\newblock {\Bem IEEE Transactions on Computational Intelligence and AI in
  Games}, {\Bem 6\/}(4), 355--366.

\bibitem[\protect\BCAY{He, Zhang, Ren,\ \BBA\ Sun}{He et~al.}{2016}]{resnet}
He, K., Zhang, X., Ren, S., \BBA\ Sun, J. \BBOP2016\BBCP.
\newblock \BBOQ Deep residual learning for image recognition\BBCQ\
\newblock In {\Bem 2016 {IEEE} Conference on Computer Vision and Pattern
  Recognition, {CVPR}, Las Vegas, NV, USA, June 27-30}, \BPGS\ 770--778. {IEEE}
  Computer Society.

\bibitem[\protect\BCAY{Henderson, Islam, Bachman, Pineau, Precup,\ \BBA\
  Meger}{Henderson et~al.}{2018}]{deeprlmatters}
Henderson, P., Islam, R., Bachman, P., Pineau, J., Precup, D., \BBA\ Meger, D.
  \BBOP2018\BBCP.
\newblock \BBOQ Deep reinforcement learning that matters\BBCQ\
\newblock In {\Bem Proceedings of the Thirty-Second {AAAI} Conference on
  Artificial Intelligence, (AAAI)}, \BPGS\ 3207--3214. {AAAI} Press.

\bibitem[\protect\BCAY{Hern{\'a}ndez-Lobato, Requeima, Pyzer-Knapp,\ \BBA\
  Aspuru-Guzik}{Hern{\'a}ndez-Lobato et~al.}{2017}]{hernandez2017parallel}
Hern{\'a}ndez-Lobato, J.~M., Requeima, J., Pyzer-Knapp, E.~O., \BBA\
  Aspuru-Guzik, A. \BBOP2017\BBCP.
\newblock \BBOQ Parallel and distributed thompson sampling for large-scale
  accelerated exploration of chemical space\BBCQ\
\newblock In {\Bem International conference on machine learning}, \BPGS\
  1470--1479. PMLR.

\bibitem[\protect\BCAY{Hertel, Baldi,\ \BBA\ Gillen}{Hertel
  et~al.}{2020}]{hpo_deep_rl_hertel_arxiv_20}
Hertel, L., Baldi, P., \BBA\ Gillen, D.~L. \BBOP2020\BBCP.
\newblock \BBOQ Quantity vs. quality: On hyperparameter optimization for deep
  reinforcement learning\BBCQ\
\newblock {\Bem CoRR}, {\Bem abs/2007.14604}.

\bibitem[\protect\BCAY{Hessel, Modayil, van Hasselt, Schaul, Ostrovski, Dabney,
  Horgan, Piot, Azar,\ \BBA\ Silver}{Hessel et~al.}{2018}]{rainbow}
Hessel, M., Modayil, J., van Hasselt, H., Schaul, T., Ostrovski, G., Dabney,
  W., Horgan, D., Piot, B., Azar, M.~G., \BBA\ Silver, D. \BBOP2018\BBCP.
\newblock \BBOQ Rainbow: Combining improvements in deep reinforcement
  learning\BBCQ.

\bibitem[\protect\BCAY{Hessel, van Hasselt, Modayil,\ \BBA\ Silver}{Hessel
  et~al.}{2019}]{inductivebiases}
Hessel, M., van Hasselt, H., Modayil, J., \BBA\ Silver, D. \BBOP2019\BBCP.
\newblock \BBOQ On inductive biases in deep reinforcement learning\BBCQ.

\bibitem[\protect\BCAY{Houthooft, Chen, Isola, Stadie, Wolski, Ho,\ \BBA\
  Abbeel}{Houthooft et~al.}{2018}]{evolvedpg}
Houthooft, R., Chen, Y., Isola, P., Stadie, B.~C., Wolski, F., Ho, J., \BBA\
  Abbeel, P. \BBOP2018\BBCP.
\newblock \BBOQ Evolved policy gradients\BBCQ\
\newblock In {\Bem Advances in Neural Information Processing Systems 31:
  NeurIPS, December 3-8, Montr{\'{e}}al, Canada}, \BPGS\ 5405--5414.

\bibitem[\protect\BCAY{Hu, Wang, Jia, Wang, Chen, Hao, Wu,\ \BBA\ Fan}{Hu
  et~al.}{2020}]{learn_reward_shaping_neurips_20}
Hu, Y., Wang, W., Jia, H., Wang, Y., Chen, Y., Hao, J., Wu, F., \BBA\ Fan, C.
  \BBOP2020\BBCP.
\newblock \BBOQ Learning to utilize shaping rewards: {A} new approach of reward
  shaping\BBCQ\
\newblock In {\Bem Advances in Neural Information Processing Systems 33:
  NeurIPS, December 6-12, virtual}.

\bibitem[\protect\BCAY{Hutter, Hoos,\ \BBA\ Leyton-Brown}{Hutter
  et~al.}{2014}]{hutter-icml14a}
Hutter, F., Hoos, H.~H., \BBA\ Leyton-Brown, K. \BBOP2014\BBCP.
\newblock \BBOQ An efficient approach for assessing hyperparameter
  importance\BBCQ\
\newblock In {\Bem Proceedings of the 31st International Conference on Machine
  Learning, ({ICML}'14)}, \BPGS\ 754--762. Omnipress.

\bibitem[\protect\BCAY{Hutter, Hoos, Leyton-Brown,\ \BBA\ Stützle}{Hutter
  et~al.}{2009}]{HutHooLeyStu09a}
Hutter, F., Hoos, H.~H., Leyton-Brown, K., \BBA\ Stützle, T. \BBOP2009\BBCP.
\newblock \BBOQ Paramils: An automatic algorithm configuration framework\BBCQ\
\newblock {\Bem Journal of Artificial Intelligence Research}, {\Bem 36},
  267--306.

\bibitem[\protect\BCAY{Hutter, Kotthoff,\ \BBA\ Vanschoren}{Hutter
  et~al.}{2019}]{automl_book}
Hutter, F., Kotthoff, L., \BBA\ Vanschoren, J.\BEDS. \BBOP2019\BBCP.
\newblock {\Bem Automated Machine Learning - Methods, Systems, Challenges}.
\newblock The Springer Series on Challenges in Machine Learning. Springer.

\bibitem[\protect\BCAY{Igl, Farquhar, Luketina, Boehmer,\ \BBA\ Whiteson}{Igl
  et~al.}{2021}]{igl2021transient}
Igl, M., Farquhar, G., Luketina, J., Boehmer, W., \BBA\ Whiteson, S.
  \BBOP2021\BBCP.
\newblock \BBOQ Transient non-stationarity and generalisation in deep
  reinforcement learning\BBCQ\
\newblock In {\Bem International Conference on Learning Representations}.

\bibitem[\protect\BCAY{Ioffe\ \BBA\ Szegedy}{Ioffe\ \BBA\
  Szegedy}{2015}]{batch_norm}
Ioffe, S.\BBACOMMA\  \BBA\ Szegedy, C. \BBOP2015\BBCP.
\newblock \BBOQ Batch normalization: Accelerating deep network training by
  reducing internal covariate shift\BBCQ\
\newblock In {\Bem Proceedings of the 32nd International Conference on Machine
  Learning, {ICML}, Lille, France, 6-11 July}, \lowercase{\BVOL}~37 of {\Bem
  {JMLR} Workshop and Conference Proceedings}, \BPGS\ 448--456. JMLR.org.

\bibitem[\protect\BCAY{Islam, Henderson, Gomrokchi,\ \BBA\ Precup}{Islam
  et~al.}{2017}]{reproducibility_cont_control_islam_arxiv17}
Islam, R., Henderson, P., Gomrokchi, M., \BBA\ Precup, D. \BBOP2017\BBCP.
\newblock \BBOQ Reproducibility of benchmarked deep reinforcement learning
  tasks for continuous control\BBCQ\
\newblock {\Bem CoRR}, {\Bem abs/1708.04133}.

\bibitem[\protect\BCAY{Jacot, Hongler,\ \BBA\ Gabriel}{Jacot
  et~al.}{2018}]{ntk}
Jacot, A., Hongler, C., \BBA\ Gabriel, F. \BBOP2018\BBCP.
\newblock \BBOQ Neural tangent kernel: Convergence and generalization in neural
  networks\BBCQ\
\newblock In {\Bem Advances in Neural Information Processing Systems 31: Annual
  Conference on Neural Information Processing Systems, NeurIPS, December 3-8,
  Montr{\'{e}}al, Canada}, \BPGS\ 8580--8589.

\bibitem[\protect\BCAY{Jaderberg, Czarnecki, Dunning, Marris, Lever,
  Casta{\~n}eda, Beattie, Rabinowitz, Morcos, Ruderman, Sonnerat, Green,
  Deason, Leibo, Silver, Hassabis, Kavukcuoglu,\ \BBA\ Graepel}{Jaderberg
  et~al.}{2019}]{Jaderberg859}
Jaderberg, M., Czarnecki, W.~M., Dunning, I., Marris, L., Lever, G.,
  Casta{\~n}eda, A.~G., Beattie, C., Rabinowitz, N.~C., Morcos, A.~S.,
  Ruderman, A., Sonnerat, N., Green, T., Deason, L., Leibo, J.~Z., Silver, D.,
  Hassabis, D., Kavukcuoglu, K., \BBA\ Graepel, T. \BBOP2019\BBCP.
\newblock \BBOQ Human-level performance in 3d multiplayer games with
  population-based reinforcement learning\BBCQ\
\newblock {\Bem Science}, {\Bem 364\/}(6443), 859--865.

\bibitem[\protect\BCAY{Jaderberg, Dalibard, Osindero, Czarnecki, Donahue,
  Razavi, Vinyals, Green, Dunning, Simonyan, Fernando,\ \BBA\
  Kavukcuoglu}{Jaderberg et~al.}{2017}]{pbt}
Jaderberg, M., Dalibard, V., Osindero, S., Czarnecki, W.~M., Donahue, J.,
  Razavi, A., Vinyals, O., Green, T., Dunning, I., Simonyan, K., Fernando, C.,
  \BBA\ Kavukcuoglu, K. \BBOP2017\BBCP.
\newblock \BBOQ Population based training of neural networks\BBCQ\
\newblock {\Bem CoRR}, {\Bem abs/1711.09846}.

\bibitem[\protect\BCAY{Jamieson\ \BBA\ Talwalkar}{Jamieson\ \BBA\
  Talwalkar}{2016}]{successive_halving_jamieson_aistats_16}
Jamieson, K.~G.\BBACOMMA\  \BBA\ Talwalkar, A. \BBOP2016\BBCP.
\newblock \BBOQ Non-stochastic best arm identification and hyperparameter
  optimization\BBCQ\
\newblock In {\Bem Proceedings of the 19th International Conference on
  Artificial Intelligence and Statistics, {AISTATS}, Cadiz, Spain, May 9-11},
  \lowercase{\BVOL}~51 of {\Bem {JMLR} Workshop and Conference Proceedings},
  \BPGS\ 240--248. JMLR.org.

\bibitem[\protect\BCAY{Janner, Fu, Zhang,\ \BBA\ Levine}{Janner
  et~al.}{2019}]{mbpo}
Janner, M., Fu, J., Zhang, M., \BBA\ Levine, S. \BBOP2019\BBCP.
\newblock \BBOQ When to trust your model: Model-based policy optimization\BBCQ\
\newblock In {\Bem Advances in Neural Information Processing Systems}.

\bibitem[\protect\BCAY{Jiang, Meng, Zhao, Shan,\ \BBA\ Hauptmann}{Jiang
  et~al.}{2015}]{jian-spl-aaai15}
Jiang, L., Meng, D., Zhao, Q., Shan, S., \BBA\ Hauptmann, A.~G. \BBOP2015\BBCP.
\newblock \BBOQ Self-paced curriculum learning\BBCQ\
\newblock In {\Bem Proceedings of the Twenty-Ninth {AAAI} Conference on
  Artificial Intelligence, January 25-30, Austin, Texas, {USA}}, \BPGS\
  2694--2700. {AAAI} Press.

\bibitem[\protect\BCAY{Jiang, Dennis, Parker{-}Holder, Foerster, Grefenstette,\
  \BBA\ Rockt{\"{a}}schel}{Jiang et~al.}{2021a}]{jiang2021robustplr}
Jiang, M., Dennis, M., Parker{-}Holder, J., Foerster, J., Grefenstette, E.,
  \BBA\ Rockt{\"{a}}schel, T. \BBOP2021a\BBCP.
\newblock \BBOQ Replay-guided adversarial environment design\BBCQ\
\newblock In {\Bem Advances in Neural Information Processing Systems}.

\bibitem[\protect\BCAY{Jiang, Grefenstette,\ \BBA\ Rockt{\"a}schel}{Jiang
  et~al.}{2021b}]{jiang2020prioritized}
Jiang, M., Grefenstette, E., \BBA\ Rockt{\"a}schel, T. \BBOP2021b\BBCP.
\newblock \BBOQ Prioritized {l}evel {r}eplay\BBCQ\
\newblock In {\Bem The International Conference on Machine Learning}.

\bibitem[\protect\BCAY{Jin}{Jin}{2006}]{moml_book_jin_2006}
Jin, Y.\BED. \BBOP2006\BBCP.
\newblock {\Bem Multi-Objective Machine Learning}, \lowercase{\BVOL}~16 of
  {\Bem Studies in Computational Intelligence}.
\newblock Springer.

\bibitem[\protect\BCAY{Jones, Schonlau,\ \BBA\ Welch}{Jones
  et~al.}{1998}]{jones1998efficient}
Jones, D.~R., Schonlau, M., \BBA\ Welch, W.~J. \BBOP1998\BBCP.
\newblock \BBOQ Efficient global optimization of expensive black-box
  functions\BBCQ\
\newblock {\Bem Journal of Global optimization}, {\Bem 13\/}(4), 455--492.

\bibitem[\protect\BCAY{Jordan, Chandak, Cohen, Zhang,\ \BBA\ Thomas}{Jordan
  et~al.}{2020}]{pmlr-v119-jordan20a}
Jordan, S., Chandak, Y., Cohen, D., Zhang, M., \BBA\ Thomas, P. \BBOP2020\BBCP.
\newblock \BBOQ Evaluating the performance of reinforcement learning
  algorithms\BBCQ\
\newblock In {\Bem Proceedings of the 37th International Conference on Machine
  Learning}, \lowercase{\BVOL}\ 119 of {\Bem Proceedings of Machine Learning
  Research}, \BPGS\ 4962--4973. PMLR.

\bibitem[\protect\BCAY{Kadioglu, Malitsky, Sellmann,\ \BBA\ Tierney}{Kadioglu
  et~al.}{2010}]{kadioglu-ecai10}
Kadioglu, S., Malitsky, Y., Sellmann, M., \BBA\ Tierney, K. \BBOP2010\BBCP.
\newblock \BBOQ {ISAC} - instance-specific algorithm configuration\BBCQ\
\newblock In {\Bem Proceedings of the Nineteenth European Conference on
  Artificial Intelligence ({ECAI}'10)}, \BPGS\ 751--756. {IOS} Press.

\bibitem[\protect\BCAY{Kaiser, Babaeizadeh, Milos, Osinski, Campbell,
  Czechowski, Erhan, Finn, Kozakowski, Levine, Mohiuddin, Sepassi, Tucker,\
  \BBA\ Michalewski}{Kaiser et~al.}{2020}]{simple}
Kaiser, L., Babaeizadeh, M., Milos, P., Osinski, B., Campbell, R.~H.,
  Czechowski, K., Erhan, D., Finn, C., Kozakowski, P., Levine, S., Mohiuddin,
  A., Sepassi, R., Tucker, G., \BBA\ Michalewski, H. \BBOP2020\BBCP.
\newblock \BBOQ Model based reinforcement learning for {A}tari\BBCQ\
\newblock In {\Bem International Conference on Learning Representations}.

\bibitem[\protect\BCAY{Kandasamy, Dasarathy, Oliva, Schneider,\ \BBA\
  Poczos}{Kandasamy et~al.}{2016}]{bayesopt_multifidelity}
Kandasamy, K., Dasarathy, G., Oliva, J.~B., Schneider, J., \BBA\ Poczos, B.
  \BBOP2016\BBCP.
\newblock \BBOQ Gaussian process bandit optimisation with multi-fidelity
  evaluations\BBCQ\
\newblock In Lee, D., Sugiyama, M., Luxburg, U., Guyon, I., \BBA\ Garnett,
  R.\BEDS, {\Bem Advances in Neural Information Processing Systems},
  \lowercase{\BVOL}~29. Curran Associates, Inc.

\bibitem[\protect\BCAY{Kearns\ \BBA\ Singh}{Kearns\ \BBA\
  Singh}{2000}]{kearnssingh}
Kearns, M.~J.\BBACOMMA\  \BBA\ Singh, S.~P. \BBOP2000\BBCP.
\newblock \BBOQ Bias-variance error bounds for temporal difference
  updates\BBCQ\
\newblock In {\Bem Proceedings of the Thirteenth Annual Conference on
  Computational Learning Theory}, COLT '00, \BPG\ 142–147, San Francisco, CA,
  USA. Morgan Kaufmann Publishers Inc.

\bibitem[\protect\BCAY{Kingma\ \BBA\ Welling}{Kingma\ \BBA\
  Welling}{2014}]{vae_kingma_iclr_14}
Kingma, D.~P.\BBACOMMA\  \BBA\ Welling, M. \BBOP2014\BBCP.
\newblock \BBOQ Auto-encoding variational bayes\BBCQ\
\newblock In Bengio, Y.\BBACOMMA\  \BBA\ LeCun, Y.\BEDS, {\Bem 2nd
  International Conference on Learning Representations, {ICLR} 2014, Banff, AB,
  Canada, April 14-16, 2014, Conference Track Proceedings}.

\bibitem[\protect\BCAY{Kirk, Zhang, Grefenstette,\ \BBA\
  Rockt{\"{a}}schel}{Kirk et~al.}{2021}]{rl_generalization_survey}
Kirk, R., Zhang, A., Grefenstette, E., \BBA\ Rockt{\"{a}}schel, T.
  \BBOP2021\BBCP.
\newblock \BBOQ A survey of generalisation in deep reinforcement learning\BBCQ\
\newblock {\Bem CoRR}, {\Bem abs/2111.09794}.

\bibitem[\protect\BCAY{Kirsch, Flennerhag, van Hasselt, Friesen, Oh,\ \BBA\
  Chen}{Kirsch et~al.}{2021}]{symmetries_bbml2021}
Kirsch, L., Flennerhag, S., van Hasselt, H., Friesen, A.~L., Oh, J., \BBA\
  Chen, Y. \BBOP2021\BBCP.
\newblock \BBOQ Introducing symmetries to black box meta reinforcement
  learning\BBCQ\
\newblock {\Bem CoRR}, {\Bem abs/2109.10781}.

\bibitem[\protect\BCAY{Kirsch, van Steenkiste,\ \BBA\ Schmidhuber}{Kirsch
  et~al.}{2020}]{learned_objectives_meta_rl_kirsch_iclr20}
Kirsch, L., van Steenkiste, S., \BBA\ Schmidhuber, J. \BBOP2020\BBCP.
\newblock \BBOQ Improving generalization in meta reinforcement learning using
  learned objectives\BBCQ\
\newblock In {\Bem 8th International Conference on Learning Representations,
  {ICLR}, Addis Ababa, Ethiopia, April 26-30}. OpenReview.net.

\bibitem[\protect\BCAY{Klein, Falkner, Bartels, Hennig,\ \BBA\ Hutter}{Klein
  et~al.}{2017}]{klein-aistats17}
Klein, A., Falkner, S., Bartels, S., Hennig, P., \BBA\ Hutter, F.
  \BBOP2017\BBCP.
\newblock \BBOQ Fast bayesian optimization of machine learning hyperparameters
  on large datasets\BBCQ\
\newblock In {\Bem Proceedings of the AISTATS conference}.

\bibitem[\protect\BCAY{Klink, D'Eramo, Peters,\ \BBA\ Pajarinen}{Klink
  et~al.}{2020}]{spdrl_klink_nips_20}
Klink, P., D'Eramo, C., Peters, J., \BBA\ Pajarinen, J. \BBOP2020\BBCP.
\newblock \BBOQ Self-paced deep reinforcement learning\BBCQ\
\newblock In {\Bem Advances in Neural Information Processing Systems 33:
  NeurIPS, December 6-12, virtual}.

\bibitem[\protect\BCAY{Konidaris\ \BBA\ Barto}{Konidaris\ \BBA\
  Barto}{2006}]{autonomous_shaping_konidaris_icml_06}
Konidaris, G.~D.\BBACOMMA\  \BBA\ Barto, A.~G. \BBOP2006\BBCP.
\newblock \BBOQ Autonomous shaping: knowledge transfer in reinforcement
  learning\BBCQ\
\newblock In {\Bem Machine Learning, Proceedings of the Twenty-Third
  International Conference {(ICML}), Pittsburgh, Pennsylvania, USA, June
  25-29}, \lowercase{\BVOL}\ 148 of {\Bem {ACM} International Conference
  Proceeding Series}, \BPGS\ 489--496. {ACM}.

\bibitem[\protect\BCAY{Krizhevsky, Sutskever,\ \BBA\ Hinton}{Krizhevsky
  et~al.}{2012}]{alexnet}
Krizhevsky, A., Sutskever, I., \BBA\ Hinton, G.~E. \BBOP2012\BBCP.
\newblock \BBOQ Imagenet classification with deep convolutional neural
  networks\BBCQ\
\newblock In {\Bem Advances in Neural Information Processing Systems 25:},
  \BPGS\ 1106--1114.

\bibitem[\protect\BCAY{Kumar, Agarwal, Ghosh,\ \BBA\ Levine}{Kumar
  et~al.}{2020a}]{kumar2020implicit}
Kumar, A., Agarwal, R., Ghosh, D., \BBA\ Levine, S. \BBOP2020a\BBCP.
\newblock \BBOQ Implicit under-parameterization inhibits data-efficient deep
  reinforcement learning\BBCQ.

\bibitem[\protect\BCAY{Kumar, Zhou, Tucker,\ \BBA\ Levine}{Kumar
  et~al.}{2020b}]{cql}
Kumar, A., Zhou, A., Tucker, G., \BBA\ Levine, S. \BBOP2020b\BBCP.
\newblock \BBOQ Conservative q-learning for offline reinforcement
  learning\BBCQ\
\newblock In {\Bem Advances in Neural Information Processing Systems 33:
  NeurIPS, December 6-12, virtual}.

\bibitem[\protect\BCAY{K{\"{u}}ttler, Nardelli, Miller, Raileanu, Selvatici,
  Grefenstette,\ \BBA\ Rockt{\"{a}}schel}{K{\"{u}}ttler et~al.}{2020}]{nle}
K{\"{u}}ttler, H., Nardelli, N., Miller, A.~H., Raileanu, R., Selvatici, M.,
  Grefenstette, E., \BBA\ Rockt{\"{a}}schel, T. \BBOP2020\BBCP.
\newblock \BBOQ {The NetHack Learning Environment}\BBCQ\
\newblock In {\Bem Proceedings of the Conference on Neural Information
  Processing Systems (NeurIPS)}.

\bibitem[\protect\BCAY{Laroche\ \BBA\ F{\'{e}}raud}{Laroche\ \BBA\
  F{\'{e}}raud}{2018a}]{rl_as_laroche_iclr18}
Laroche, R.\BBACOMMA\  \BBA\ F{\'{e}}raud, R. \BBOP2018a\BBCP.
\newblock \BBOQ Reinforcement learning algorithm selection\BBCQ\
\newblock In {\Bem 6th International Conference on Learning Representations,
  {ICLR}, Vancouver, BC, Canada, April 30 - May 3, Conference Track
  Proceedings}. OpenReview.net.

\bibitem[\protect\BCAY{Laroche\ \BBA\ F{\'{e}}raud}{Laroche\ \BBA\
  F{\'{e}}raud}{2018b}]{DBLP:conf/iclr/LarocheF18}
Laroche, R.\BBACOMMA\  \BBA\ F{\'{e}}raud, R. \BBOP2018b\BBCP.
\newblock \BBOQ Reinforcement learning algorithm selection\BBCQ\
\newblock In {\Bem 6th International Conference on Learning Representations,
  {ICLR}, Vancouver, BC, Canada, April 30 - May 3, Conference Track
  Proceedings}. OpenReview.net.

\bibitem[\protect\BCAY{Lee, Hwangbo, Wellhausen, Koltun,\ \BBA\ Hutter}{Lee
  et~al.}{2020}]{quadrup_poetlike}
Lee, J., Hwangbo, J., Wellhausen, L., Koltun, V., \BBA\ Hutter, M.
  \BBOP2020\BBCP.
\newblock \BBOQ Learning quadrupedal locomotion over challenging terrain\BBCQ\
\newblock {\Bem Sci. Robotics}, {\Bem 5\/}(47), 5986.

\bibitem[\protect\BCAY{Lee, Yosinski, Glette, Lipson,\ \BBA\ Clune}{Lee
  et~al.}{2013}]{hyperneatgaits2013}
Lee, S., Yosinski, J., Glette, K., Lipson, H., \BBA\ Clune, J. \BBOP2013\BBCP.
\newblock \BBOQ Evolving gaits for physical robots with the hyperneat
  generative encoding: The benefits of simulation\BBCQ.
\newblock \BPGS\ 540--549.

\bibitem[\protect\BCAY{Lehman, Chen, Clune,\ \BBA\ Stanley}{Lehman
  et~al.}{2018}]{lehman2018es}
Lehman, J., Chen, J., Clune, J., \BBA\ Stanley, K.~O. \BBOP2018\BBCP.
\newblock \BBOQ Es is more than just a traditional finite-difference
  approximator\BBCQ.

\bibitem[\protect\BCAY{Li, Santu, Gupta, Nguyen, Venkatesh, Sutti, Leal,
  Slezak, Height, Mohammed, et~al.}{Li et~al.}{2018}]{li2018accelerating}
Li, C., Santu, R., Gupta, S., Nguyen, V., Venkatesh, S., Sutti, A., Leal, D. R.
  D.~C., Slezak, T., Height, M., Mohammed, M., et~al. \BBOP2018\BBCP.
\newblock \BBOQ Accelerating experimental design by incorporating experimenter
  hunches\BBCQ\
\newblock In {\Bem 18th IEEE International Conference on Data Mining, ICDM},
  \BPGS\ 257--266. IEEE.

\bibitem[\protect\BCAY{Li, Jamieson, DeSalvo, Rostamizadeh,\ \BBA\
  Talwalkar}{Li et~al.}{2017}]{hyperband}
Li, L., Jamieson, K., DeSalvo, G., Rostamizadeh, A., \BBA\ Talwalkar, A.
  \BBOP2017\BBCP.
\newblock \BBOQ Hyperband: A novel bandit-based approach to hyperparameter
  optimization\BBCQ\
\newblock In {\Bem Journal Machine Learning Research}.

\bibitem[\protect\BCAY{Lillicrap, Hunt, Pritzel, Heess, Erez, Tassa, Silver,\
  \BBA\ Wierstra}{Lillicrap et~al.}{2016}]{ddpg}
Lillicrap, T.~P., Hunt, J.~J., Pritzel, A., Heess, N., Erez, T., Tassa, Y.,
  Silver, D., \BBA\ Wierstra, D. \BBOP2016\BBCP.
\newblock \BBOQ Continuous control with deep reinforcement learning\BBCQ\
\newblock In {\Bem 4th International Conference on Learning Representations,
  {ICLR}, San Juan, Puerto Rico, May 2-4, Conference Track Proceedings}.

\bibitem[\protect\BCAY{Liu, Simonyan,\ \BBA\ Yang}{Liu et~al.}{2019a}]{darts}
Liu, H., Simonyan, K., \BBA\ Yang, Y. \BBOP2019a\BBCP.
\newblock \BBOQ {DARTS:} differentiable architecture search\BBCQ\
\newblock In {\Bem 7th International Conference on Learning Representations,
  {ICLR}, New Orleans, LA, USA, May 6-9}. OpenReview.net.

\bibitem[\protect\BCAY{Liu, Lever, Merel, Tunyasuvunakool, Heess,\ \BBA\
  Graepel}{Liu et~al.}{2019b}]{liu2018emergent}
Liu, S., Lever, G., Merel, J., Tunyasuvunakool, S., Heess, N., \BBA\ Graepel,
  T. \BBOP2019b\BBCP.
\newblock \BBOQ Emergent coordination through competition\BBCQ\
\newblock In {\Bem 7th International Conference on Learning Representations,
  {ICLR}, New Orleans, LA, USA, May 6-9}. OpenReview.net.

\bibitem[\protect\BCAY{Liu, Lever, Wang, Merel, Eslami, Hennes, Czarnecki,
  Tassa, Omidshafiei, Abdolmaleki, Siegel, Hasenclever, Marris,
  Tunyasuvunakool, Song, Wulfmeier, Muller, Haarnoja, Tracey, Tuyls, Graepel,\
  \BBA\ Heess}{Liu et~al.}{2021a}]{pbtfootball}
Liu, S., Lever, G., Wang, Z., Merel, J., Eslami, S. M.~A., Hennes, D.,
  Czarnecki, W.~M., Tassa, Y., Omidshafiei, S., Abdolmaleki, A., Siegel, N.~Y.,
  Hasenclever, L., Marris, L., Tunyasuvunakool, S., Song, H.~F., Wulfmeier, M.,
  Muller, P., Haarnoja, T., Tracey, B.~D., Tuyls, K., Graepel, T., \BBA\ Heess,
  N. \BBOP2021a\BBCP.
\newblock \BBOQ From motor control to team play in simulated humanoid
  football\BBCQ\
\newblock {\Bem CoRR}, {\Bem abs/2105.12196}.

\bibitem[\protect\BCAY{Liu, Li, Kang,\ \BBA\ Darrell}{Liu
  et~al.}{2021b}]{liu2021regularization}
Liu, Z., Li, X., Kang, B., \BBA\ Darrell, T. \BBOP2021b\BBCP.
\newblock \BBOQ Regularization matters in policy optimization - an empirical
  study on continuous control\BBCQ\
\newblock In {\Bem International Conference on Learning Representations}.

\bibitem[\protect\BCAY{Lu, Ball, Parker{-}Holder, Osborne,\ \BBA\ Roberts}{Lu
  et~al.}{2021}]{lu2021revisiting}
Lu, C., Ball, P.~J., Parker{-}Holder, J., Osborne, M., \BBA\ Roberts, S.
  \BBOP2021\BBCP.
\newblock \BBOQ Revisiting design choices in offline model based reinforcement
  learning\BBCQ\
\newblock In {\Bem RL for Real Life Workshop at ICML}.

\bibitem[\protect\BCAY{Mankowitz, Mann, Bacon, Precup,\ \BBA\ Mannor}{Mankowitz
  et~al.}{2018}]{robust_options}
Mankowitz, D.~J., Mann, T.~A., Bacon, P., Precup, D., \BBA\ Mannor, S.
  \BBOP2018\BBCP.
\newblock \BBOQ Learning robust options\BBCQ\
\newblock In {\Bem Proceedings of the Thirty-Second {AAAI} Conference on
  Artificial Intelligence, (AAAI)}, \BPGS\ 6409--6416. {AAAI} Press.

\bibitem[\protect\BCAY{Marthi}{Marthi}{2007}]{automatic_shaping_and_decomposition_marthi_icml_2007}
Marthi, B. \BBOP2007\BBCP.
\newblock \BBOQ Automatic shaping and decomposition of reward functions\BBCQ.
\newblock ICML '07, \BPG\ 601–608, New York, NY, USA. Association for
  Computing Machinery.

\bibitem[\protect\BCAY{Matiisen, Oliver, Cohen,\ \BBA\ Schulman}{Matiisen
  et~al.}{2020}]{teacher_student_curri_matiisen_ieee_20}
Matiisen, T., Oliver, A., Cohen, T., \BBA\ Schulman, J. \BBOP2020\BBCP.
\newblock \BBOQ Teacher-student curriculum learning\BBCQ\
\newblock {\Bem {IEEE} Trans. Neural Networks Learn. Syst.}, {\Bem 31\/}(9),
  3732--3740.

\bibitem[\protect\BCAY{Miao, Song, Peng, Yue, Brevdo,\ \BBA\ Faust}{Miao
  et~al.}{2021}]{rl_darts}
Miao, Y., Song, X., Peng, D., Yue, S., Brevdo, E., \BBA\ Faust, A.
  \BBOP2021\BBCP.
\newblock \BBOQ {RL-DARTS:} differentiable architecture search for
  reinforcement learning\BBCQ\
\newblock {\Bem CoRR}, {\Bem abs/2106.02229}.

\bibitem[\protect\BCAY{Michalewicz}{Michalewicz}{2013}]{michalewicz2013genetic}
Michalewicz, Z. \BBOP2013\BBCP.
\newblock {\Bem Genetic algorithms+ data structures= evolution programs}.
\newblock Springer Science \& Business Media.

\bibitem[\protect\BCAY{Mirjalili\ \BBA\ Lewis}{Mirjalili\ \BBA\
  Lewis}{2016}]{WhaleOptAlgo}
Mirjalili, S.\BBACOMMA\  \BBA\ Lewis, A. \BBOP2016\BBCP.
\newblock \BBOQ The whale optimization algorithm\BBCQ\
\newblock {\Bem Advances in Engineering Software}, {\Bem 95}, 51--67.

\bibitem[\protect\BCAY{Mnih, Badia, Mirza, Graves, Lillicrap, Harley, Silver,\
  \BBA\ Kavukcuoglu}{Mnih et~al.}{2016}]{a3c_mnih_icml_16}
Mnih, V., Badia, A.~P., Mirza, M., Graves, A., Lillicrap, T.~P., Harley, T.,
  Silver, D., \BBA\ Kavukcuoglu, K. \BBOP2016\BBCP.
\newblock \BBOQ Asynchronous methods for deep reinforcement learning\BBCQ\
\newblock In Balcan, M.\BBACOMMA\  \BBA\ Weinberger, K.~Q.\BEDS, {\Bem
  Proceedings of the 33nd International Conference on Machine Learning, {ICML}
  2016, New York City, NY, USA, June 19-24, 2016}, \lowercase{\BVOL}~48 of
  {\Bem {JMLR} Workshop and Conference Proceedings}, \BPGS\ 1928--1937.
  JMLR.org.

\bibitem[\protect\BCAY{Mnih, Kavukcuoglu, Silver, Rusu, Veness, Bellemare,
  Graves, Riedmiller, Fidjeland, Ostrovski, Petersen, Beattie, Sadik,
  Antonoglou, King, Kumaran, Wierstra, Legg,\ \BBA\ Hassabis}{Mnih
  et~al.}{2015}]{dqn}
Mnih, V., Kavukcuoglu, K., Silver, D., Rusu, A., Veness, J., Bellemare, M.,
  Graves, A., Riedmiller, M., Fidjeland, A., Ostrovski, G., Petersen, S.,
  Beattie, C., Sadik, A., Antonoglou, I., King, H., Kumaran, D., Wierstra, D.,
  Legg, S., \BBA\ Hassabis, D. \BBOP2015\BBCP.
\newblock \BBOQ Human-level control through deep reinforcement learning\BBCQ\
\newblock {\Bem Nature}, {\Bem 518}, 529--33.

\bibitem[\protect\BCAY{Mockus}{Mockus}{1974}]{original_bayesopt}
Mockus, J. \BBOP1974\BBCP.
\newblock \BBOQ On bayesian methods for seeking the extremum\BBCQ\
\newblock In {\Bem Optimization Techniques, {IFIP} Technical Conference,
  Novosibirsk, USSR, July 1-7}, \lowercase{\BVOL}~27 of {\Bem Lecture Notes in
  Computer Science}, \BPGS\ 400--404. Springer.

\bibitem[\protect\BCAY{Moffaert\ \BBA\ Now{\'{e}}}{Moffaert\ \BBA\
  Now{\'{e}}}{2014}]{morl_moffaert_jmlr_15}
Moffaert, K.~V.\BBACOMMA\  \BBA\ Now{\'{e}}, A. \BBOP2014\BBCP.
\newblock \BBOQ Multi-objective reinforcement learning using sets of pareto
  dominating policies\BBCQ\
\newblock {\Bem J. Mach. Learn. Res.}, {\Bem 15\/}(1), 3483--3512.

\bibitem[\protect\BCAY{Moskovitz, Parker{-}Holder, Pacchiano,\ \BBA\
  Arbel}{Moskovitz et~al.}{2021}]{top}
Moskovitz, T., Parker{-}Holder, J., Pacchiano, A., \BBA\ Arbel, M.
  \BBOP2021\BBCP.
\newblock \BBOQ Deep reinforcement learning with dynamic optimism\BBCQ\
\newblock In {\Bem Advances in Neural Information Processing Systems}.

\bibitem[\protect\BCAY{Moulines\ \BBA\ Bach}{Moulines\ \BBA\
  Bach}{2011}]{moulines-neurips11}
Moulines, E.\BBACOMMA\  \BBA\ Bach, F. \BBOP2011\BBCP.
\newblock \BBOQ Non-asymptotic analysis of stochastic approximation algorithms
  for machine learning\BBCQ\
\newblock In {\Bem Advances in Neural Information Processing Systems},
  \lowercase{\BVOL}~24. Curran Associates, Inc.

\bibitem[\protect\BCAY{Narvekar, Peng, Leonetti, Sinapov, Taylor,\ \BBA\
  Stone}{Narvekar et~al.}{2020}]{narvekar-cl-survey-jmlr20}
Narvekar, S., Peng, B., Leonetti, M., Sinapov, J., Taylor, M.~E., \BBA\ Stone,
  P. \BBOP2020\BBCP.
\newblock \BBOQ Curriculum learning for reinforcement learning domains: {A}
  framework and survey\BBCQ\
\newblock {\Bem J. Mach. Learn. Res.}, {\Bem 21}, 181:1--181:50.

\bibitem[\protect\BCAY{Neyshabur}{Neyshabur}{2017}]{implicit_regularization}
Neyshabur, B. \BBOP2017\BBCP.
\newblock \BBOQ Implicit regularization in deep learning\BBCQ\
\newblock {\Bem CoRR}, {\Bem abs/1709.01953}.

\bibitem[\protect\BCAY{Neyshabur, Li, Bhojanapalli, LeCun,\ \BBA\
  Srebro}{Neyshabur et~al.}{2019}]{width_complexity_measure_neyshabur19}
Neyshabur, B., Li, Z., Bhojanapalli, S., LeCun, Y., \BBA\ Srebro, N.
  \BBOP2019\BBCP.
\newblock \BBOQ The role of over-parametrization in generalization of neural
  networks\BBCQ\
\newblock In {\Bem 7th International Conference on Learning Representations,
  {ICLR}, New Orleans, LA, USA, May 6-9}. OpenReview.net.

\bibitem[\protect\BCAY{Neyshabur, Tomioka,\ \BBA\ Srebro}{Neyshabur
  et~al.}{2015}]{practical_overparameterization_neyshabur15}
Neyshabur, B., Tomioka, R., \BBA\ Srebro, N. \BBOP2015\BBCP.
\newblock \BBOQ In search of the real inductive bias: On the role of implicit
  regularization in deep learning\BBCQ\
\newblock In {\Bem 3rd International Conference on Learning Representations,
  {ICLR}, San Diego, CA, USA, May 7-9, Workshop Track Proceedings}.

\bibitem[\protect\BCAY{Ng, Harada,\ \BBA\ Russell}{Ng
  et~al.}{1999}]{reward_shaping_ng_icml_99}
Ng, A.~Y., Harada, D., \BBA\ Russell, S.~J. \BBOP1999\BBCP.
\newblock \BBOQ Policy invariance under reward transformations: Theory and
  application to reward shaping\BBCQ\
\newblock In {\Bem Proceedings of the Sixteenth International Conference on
  Machine Learning {(ICML}), Bled, Slovenia, June 27 - 30}, \BPGS\ 278--287.
  Morgan Kaufmann.

\bibitem[\protect\BCAY{Nguyen, Orbell, Lennon, Moon, Vigneau, Camenzind, Yu,
  Zumb{\"u}hl, Briggs, Osborne, et~al.}{Nguyen et~al.}{2021}]{nguyen2021deep}
Nguyen, V., Orbell, S., Lennon, D.~T., Moon, H., Vigneau, F., Camenzind, L.~C.,
  Yu, L., Zumb{\"u}hl, D.~M., Briggs, G. A.~D., Osborne, M.~A., et~al.
  \BBOP2021\BBCP.
\newblock \BBOQ Deep reinforcement learning for efficient measurement of
  quantum devices\BBCQ\
\newblock {\Bem npj Quantum Information}, {\Bem 7\/}(1), 1--9.

\bibitem[\protect\BCAY{Nguyen\ \BBA\ Osborne}{Nguyen\ \BBA\
  Osborne}{2020}]{nguyen2019knowing}
Nguyen, V.\BBACOMMA\  \BBA\ Osborne, M.~A. \BBOP2020\BBCP.
\newblock \BBOQ Knowing the what but not the where in bayesian
  optimization\BBCQ\
\newblock In {\Bem Proceedings of the 37th International Conference on Machine
  Learning, {ICML}, 13-18 July, Virtual Event}, \lowercase{\BVOL}\ 119 of {\Bem
  Proceedings of Machine Learning Research}, \BPGS\ 7317--7326. {PMLR}.

\bibitem[\protect\BCAY{Nguyen, Schulze,\ \BBA\ Osborne}{Nguyen
  et~al.}{2020}]{boil}
Nguyen, V., Schulze, S., \BBA\ Osborne, M.~A. \BBOP2020\BBCP.
\newblock \BBOQ Bayesian optimization for iterative learning\BBCQ\
\newblock In {\Bem Advances in Neural Information Processing Systems
  (NeurIPS)}.

\bibitem[\protect\BCAY{Obando-Ceron\ \BBA\ Castro}{Obando-Ceron\ \BBA\
  Castro}{2020}]{obando20revisiting}
Obando-Ceron, J.~S.\BBACOMMA\  \BBA\ Castro, P.~S. \BBOP2020\BBCP.
\newblock \BBOQ Revisiting rainbow: Promoting more insightful and inclusive
  deep reinforcement learning research\BBCQ\
\newblock In {\Bem Deep Reinforcement Learning Workshop, NeurIPS}.

\bibitem[\protect\BCAY{Oh, Hessel, Czarnecki, Xu, van Hasselt, Singh,\ \BBA\
  Silver}{Oh et~al.}{2020}]{learnedPG}
Oh, J., Hessel, M., Czarnecki, W.~M., Xu, Z., van Hasselt, H., Singh, S., \BBA\
  Silver, D. \BBOP2020\BBCP.
\newblock \BBOQ Discovering reinforcement learning algorithms\BBCQ\
\newblock In {\Bem Advances in Neural Information Processing Systems 33}.

\bibitem[\protect\BCAY{OpenAI, Akkaya, Andrychowicz, Chociej, Litwin, McGrew,
  Petron, Paino, Plappert, Powell, Ribas, Schneider, Tezak, Tworek, Welinder,
  Weng, Yuan, Zaremba,\ \BBA\ Zhang}{OpenAI et~al.}{2019}]{openai-rubikscube}
OpenAI, Akkaya, I., Andrychowicz, M., Chociej, M., Litwin, M., McGrew, B.,
  Petron, A., Paino, A., Plappert, M., Powell, G., Ribas, R., Schneider, J.,
  Tezak, N., Tworek, J., Welinder, P., Weng, L., Yuan, Q., Zaremba, W., \BBA\
  Zhang, L. \BBOP2019\BBCP.
\newblock \BBOQ Solving rubik's cube with a robot hand\BBCQ\
\newblock {\Bem CoRR}, {\Bem abs/1910.07113}.

\bibitem[\protect\BCAY{OpenAI, Andrychowicz, Baker, Chociej, J{\'{o}}zefowicz,
  McGrew, Pachocki, Pachocki, Petron, Plappert, Powell, Ray, Schneider, Sidor,
  Tobin, Welinder, Weng,\ \BBA\ Zaremba}{OpenAI
  et~al.}{2018}]{dexterous_openai}
OpenAI, Andrychowicz, M., Baker, B., Chociej, M., J{\'{o}}zefowicz, R., McGrew,
  B., Pachocki, J.~W., Pachocki, J., Petron, A., Plappert, M., Powell, G., Ray,
  A., Schneider, J., Sidor, S., Tobin, J., Welinder, P., Weng, L., \BBA\
  Zaremba, W. \BBOP2018\BBCP.
\newblock \BBOQ Learning dexterous in-hand manipulation\BBCQ\
\newblock {\Bem CoRR}, {\Bem abs/1808.00177}.

\bibitem[\protect\BCAY{OpenAI, Plappert, Sampedro, Xu, Akkaya, Kosaraju,
  Welinder, D'Sa, Petron, de~Oliveira~Pinto, Paino, Noh, Weng, Yuan, Chu,\
  \BBA\ Zaremba}{OpenAI et~al.}{2021}]{openai2021asymmetric}
OpenAI, O., Plappert, M., Sampedro, R., Xu, T., Akkaya, I., Kosaraju, V.,
  Welinder, P., D'Sa, R., Petron, A., de~Oliveira~Pinto, H.~P., Paino, A., Noh,
  H., Weng, L., Yuan, Q., Chu, C., \BBA\ Zaremba, W. \BBOP2021\BBCP.
\newblock \BBOQ Asymmetric self-play for automatic goal discovery in robotic
  manipulation\BBCQ.

\bibitem[\protect\BCAY{Osband, Doron, Hessel, Aslanides, Sezener, Saraiva,
  McKinney, Lattimore, {Sz}epesv{\'a}ri, Singh, Van~Roy, Sutton, Silver,\ \BBA\
  van Hasselt}{Osband et~al.}{2020}]{osband2020bsuite}
Osband, I., Doron, Y., Hessel, M., Aslanides, J., Sezener, E., Saraiva, A.,
  McKinney, K., Lattimore, T., {Sz}epesv{\'a}ri, C., Singh, S., Van~Roy, B.,
  Sutton, R., Silver, D., \BBA\ van Hasselt, H. \BBOP2020\BBCP.
\newblock \BBOQ Behaviour suite for reinforcement learning\BBCQ\
\newblock In {\Bem International Conference on Learning Representations}.

\bibitem[\protect\BCAY{Parisotto, Song, Rae, Pascanu, G{\"{u}}l{\c{c}}ehre,
  Jayakumar, Jaderberg, Kaufman, Clark, Noury, Botvinick, Heess,\ \BBA\
  Hadsell}{Parisotto et~al.}{2020}]{transformersforrl}
Parisotto, E., Song, H.~F., Rae, J.~W., Pascanu, R., G{\"{u}}l{\c{c}}ehre,
  {\c{C}}., Jayakumar, S.~M., Jaderberg, M., Kaufman, R.~L., Clark, A., Noury,
  S., Botvinick, M., Heess, N., \BBA\ Hadsell, R. \BBOP2020\BBCP.
\newblock \BBOQ Stabilizing transformers for reinforcement learning\BBCQ\
\newblock In {\Bem Proceedings of the 37th International Conference on Machine
  Learning, {ICML}, 13-18 July, Virtual Event}, \lowercase{\BVOL}\ 119 of {\Bem
  Proceedings of Machine Learning Research}, \BPGS\ 7487--7498. {PMLR}.

\bibitem[\protect\BCAY{Parker-Holder, Nguyen, Desai,\ \BBA\
  Roberts}{Parker-Holder et~al.}{2021}]{parkerholder2021tuning}
Parker-Holder, J., Nguyen, V., Desai, S., \BBA\ Roberts, S. \BBOP2021\BBCP.
\newblock \BBOQ Tuning {M}ixed {I}nput {H}yperparameters on the {F}ly for
  {E}fficient {P}opulation {B}ased {A}uto{RL}\BBCQ\
\newblock In {\Bem Advances in Neural Information Processing Systems},
  \lowercase{\BVOL}~34.

\bibitem[\protect\BCAY{Parker-Holder, Nguyen,\ \BBA\ Roberts}{Parker-Holder
  et~al.}{2020a}]{pb2}
Parker-Holder, J., Nguyen, V., \BBA\ Roberts, S.~J. \BBOP2020a\BBCP.
\newblock \BBOQ Provably efficient online hyperparameter optimization with
  population-based bandits\BBCQ\
\newblock {\Bem Advances in Neural Information Processing Systems}, {\Bem 33}.

\bibitem[\protect\BCAY{Parker{-}Holder, Pacchiano, Choromanski,\ \BBA\
  Roberts}{Parker{-}Holder et~al.}{2020b}]{dvd}
Parker{-}Holder, J., Pacchiano, A., Choromanski, K., \BBA\ Roberts, S.
  \BBOP2020b\BBCP.
\newblock \BBOQ Effective diversity in population-based reinforcement
  learning\BBCQ\
\newblock In {\Bem Advances in Neural Information Processing Systems 33}.

\bibitem[\protect\BCAY{Paszke, Gross, Massa, Lerer, Bradbury, Chanan, Killeen,
  Lin, Gimelshein, Antiga, Desmaison, Kopf, Yang, DeVito, Raison, Tejani,
  Chilamkurthy, Steiner, Fang, Bai,\ \BBA\ Chintala}{Paszke
  et~al.}{2019}]{pytorch_neurips_2019}
Paszke, A., Gross, S., Massa, F., Lerer, A., Bradbury, J., Chanan, G., Killeen,
  T., Lin, Z., Gimelshein, N., Antiga, L., Desmaison, A., Kopf, A., Yang, E.,
  DeVito, Z., Raison, M., Tejani, A., Chilamkurthy, S., Steiner, B., Fang, L.,
  Bai, J., \BBA\ Chintala, S. \BBOP2019\BBCP.
\newblock \BBOQ Pytorch: An imperative style, high-performance deep learning
  library\BBCQ\
\newblock In Wallach, H., Larochelle, H., Beygelzimer, A., d\textquotesingle
  Alch\'{e}-Buc, F., Fox, E., \BBA\ Garnett, R.\BEDS, {\Bem Advances in Neural
  Information Processing Systems 32}, \BPGS\ 8024--8035. Curran Associates,
  Inc.

\bibitem[\protect\BCAY{Patterson, Neumann, White, Kumaraswamy,\ \BBA\
  White}{Patterson et~al.}{2021}]{crossenvhyperopt-2021}
Patterson, A., Neumann, S., White, A.~M., Kumaraswamy, R., \BBA\ White, M.
  \BBOP2021\BBCP.
\newblock \BBOQ The cross-environment hyperparameter setting benchmark for
  reinforcement learning\BBCQ.

\bibitem[\protect\BCAY{Paul, Kurin,\ \BBA\ Whiteson}{Paul et~al.}{2019}]{hoof}
Paul, S., Kurin, V., \BBA\ Whiteson, S. \BBOP2019\BBCP.
\newblock \BBOQ Fast efficient hyperparameter tuning for policy gradients\BBCQ.

\bibitem[\protect\BCAY{P{\'{e}}rez{-}Cruz, Vaerenbergh, Murillo{-}Fuentes,
  L{\'{a}}zaro{-}Gredilla,\ \BBA\ Santamar{\'{\i}}a}{P{\'{e}}rez{-}Cruz
  et~al.}{2013}]{gp_picture}
P{\'{e}}rez{-}Cruz, F., Vaerenbergh, S.~V., Murillo{-}Fuentes, J.~J.,
  L{\'{a}}zaro{-}Gredilla, M., \BBA\ Santamar{\'{\i}}a, I. \BBOP2013\BBCP.
\newblock \BBOQ Gaussian processes for nonlinear signal processing: An overview
  of recent advances\BBCQ\
\newblock {\Bem {IEEE} Signal Process. Mag.}, {\Bem 30\/}(4), 40--50.

\bibitem[\protect\BCAY{Perrone, Shen, Zolic, Shcherbatyi, Ahmed, Bansal,
  Donini, Winkelmolen, Jenatton, Faddoul, Pogorzelska, Miladinovic, Kenthapadi,
  Seeger,\ \BBA\ Archambeau}{Perrone et~al.}{2021}]{perrone2021amazon}
Perrone, V., Shen, H., Zolic, A., Shcherbatyi, I., Ahmed, A., Bansal, T.,
  Donini, M., Winkelmolen, F., Jenatton, R., Faddoul, J.~B., Pogorzelska, B.,
  Miladinovic, M., Kenthapadi, K., Seeger, M.~W., \BBA\ Archambeau, C.
  \BBOP2021\BBCP.
\newblock \BBOQ Amazon sagemaker automatic model tuning: Scalable gradient-free
  optimization\BBCQ\
\newblock In Zhu, F., Ooi, B.~C., \BBA\ Miao, C.\BEDS, {\Bem {KDD} '21: The
  27th {ACM} {SIGKDD} Conference on Knowledge Discovery and Data Mining,
  Virtual Event, Singapore, August 14-18, 2021}, \BPGS\ 3463--3471. {ACM}.

\bibitem[\protect\BCAY{Precup, Sutton,\ \BBA\ Singh}{Precup
  et~al.}{2000}]{precup2000ope}
Precup, D., Sutton, R.~S., \BBA\ Singh, S.~P. \BBOP2000\BBCP.
\newblock \BBOQ Eligibility traces for off-policy policy evaluation\BBCQ\
\newblock In Langley, P.\BED, {\Bem Proceedings of the Seventeenth
  International Conference on Machine Learning {(ICML} 2000), Stanford
  University, Stanford, CA, USA, June 29 - July 2, 2000}, \BPGS\ 759--766.
  Morgan Kaufmann.

\bibitem[\protect\BCAY{Prokhorov\ \BBA\ Wunsch}{Prokhorov\ \BBA\
  Wunsch}{1997}]{adaptivecriticdesigns}
Prokhorov, D.\BBACOMMA\  \BBA\ Wunsch, D. \BBOP1997\BBCP.
\newblock \BBOQ Adaptive critic designs\BBCQ\
\newblock {\Bem IEEE Transactions on Neural Networks}, {\Bem 8\/}(5),
  997--1007.

\bibitem[\protect\BCAY{Raileanu, Goldstein, Yarats, Kostrikov,\ \BBA\
  Fergus}{Raileanu et~al.}{2020}]{ucb_drac}
Raileanu, R., Goldstein, M., Yarats, D., Kostrikov, I., \BBA\ Fergus, R.
  \BBOP2020\BBCP.
\newblock \BBOQ Automatic data augmentation for generalization in deep
  reinforcement learning\BBCQ\
\newblock {\Bem CoRR}, {\Bem abs/2006.12862}.

\bibitem[\protect\BCAY{Rajan\ \BBA\ Hutter}{Rajan\ \BBA\
  Hutter}{2019}]{mdp_playground_rajan_neurips_workshop_19}
Rajan, R.\BBACOMMA\  \BBA\ Hutter, F. \BBOP2019\BBCP.
\newblock \BBOQ Mdp playground: Meta-features in reinforcement learning\BBCQ\
\newblock In {\Bem {NeurIPS} {Deep RL} Workshop}.

\bibitem[\protect\BCAY{Rakelly, Zhou, Finn, Levine,\ \BBA\ Quillen}{Rakelly
  et~al.}{2019}]{pearl_rakelly_icml19}
Rakelly, K., Zhou, A., Finn, C., Levine, S., \BBA\ Quillen, D. \BBOP2019\BBCP.
\newblock \BBOQ Efficient off-policy meta-reinforcement learning via
  probabilistic context variables\BBCQ\
\newblock In {\Bem Proceedings of the 36th International Conference on Machine
  Learning, {ICML}, 9-15 June, Long Beach, California, {USA}},
  \lowercase{\BVOL}~97 of {\Bem Proceedings of Machine Learning Research},
  \BPGS\ 5331--5340. {PMLR}.

\bibitem[\protect\BCAY{Real, Aggarwal, Huang,\ \BBA\ Le}{Real
  et~al.}{2019}]{regevo}
Real, E., Aggarwal, A., Huang, Y., \BBA\ Le, Q.~V. \BBOP2019\BBCP.
\newblock \BBOQ Regularized evolution for image classifier architecture
  search\BBCQ\
\newblock In {\Bem The Thirty-Third {AAAI} Conference on Artificial
  Intelligence, {AAAI}}, \BPGS\ 4780--4789. {AAAI} Press.

\bibitem[\protect\BCAY{Rechenberg}{Rechenberg}{1973}]{es1973}
Rechenberg, I. \BBOP1973\BBCP.
\newblock \BBOQ Evolutionsstrategie : Optimierung technischer systeme nach
  prinzipien der biologischen evolution\BBCQ.

\bibitem[\protect\BCAY{Rice}{Rice}{1976}]{rice1976algorithm}
Rice, J.~R. \BBOP1976\BBCP.
\newblock \BBOQ The algorithm selection problem\BBCQ\
\newblock In {\Bem Advances in computers}, \lowercase{\BVOL}~15, \BPGS\
  65--118. Elsevier.

\bibitem[\protect\BCAY{Riquelme, Penedones, Vincent, Maennel, Gelly, Mann,
  Barreto,\ \BBA\ Neu}{Riquelme et~al.}{2019}]{adaptive_td_carlos_neurips_2019}
Riquelme, C., Penedones, H., Vincent, D., Maennel, H., Gelly, S., Mann, T.~A.,
  Barreto, A., \BBA\ Neu, G. \BBOP2019\BBCP.
\newblock \BBOQ Adaptive temporal-difference learning for policy evaluation
  with per-state uncertainty estimates\BBCQ\
\newblock In {\Bem Advances in Neural Information Processing Systems},
  \lowercase{\BVOL}~32.

\bibitem[\protect\BCAY{Risi\ \BBA\ Stanley}{Risi\ \BBA\
  Stanley}{2013}]{morphologycontrol2013}
Risi, S.\BBACOMMA\  \BBA\ Stanley, K.~O. \BBOP2013\BBCP.
\newblock \BBOQ Confronting the challenge of learning a flexible neural
  controller for a diversity of morphologies\BBCQ\
\newblock In {\Bem Proceedings of the 15th Annual Conference on Genetic and
  Evolutionary Computation}, GECCO '13, \BPG\ 255–262, New York, NY, USA.
  Association for Computing Machinery.

\bibitem[\protect\BCAY{Romac, Portelas, Hofmann,\ \BBA\ Oudeyer}{Romac
  et~al.}{2021}]{romac-tma-icml21}
Romac, C., Portelas, R., Hofmann, K., \BBA\ Oudeyer, P. \BBOP2021\BBCP.
\newblock \BBOQ Teachmyagent: a benchmark for automatic curriculum learning in
  deep {RL}\BBCQ\
\newblock In {\Bem Proceedings of the 38th International Conference on Machine
  Learning, {ICML}, 18-24 July, Virtual Event}, \lowercase{\BVOL}\ 139 of {\Bem
  Proceedings of Machine Learning Research}, \BPGS\ 9052--9063. {PMLR}.

\bibitem[\protect\BCAY{Rowland, Dabney,\ \BBA\ Munos}{Rowland
  et~al.}{2020}]{rowland2020adaptive}
Rowland, M., Dabney, W., \BBA\ Munos, R. \BBOP2020\BBCP.
\newblock \BBOQ Adaptive trade-offs in off-policy learning\BBCQ\
\newblock In {\Bem Proceedings of the Twenty Third International Conference on
  Artificial Intelligence and Statistics}. PMLR.

\bibitem[\protect\BCAY{Runge, Stoll, Falkner,\ \BBA\ Hutter}{Runge
  et~al.}{2019}]{runge2018learning}
Runge, F., Stoll, D., Falkner, S., \BBA\ Hutter, F. \BBOP2019\BBCP.
\newblock \BBOQ Learning to design {RNA}\BBCQ\
\newblock In {\Bem 7th International Conference on Learning Representations,
  {ICLR}, New Orleans, LA, USA, May 6-9}. OpenReview.net.

\bibitem[\protect\BCAY{Samvelyan, Kirk, Kurin, Parker-Holder, Jiang, Hambro,
  Petroni, Kuttler, Grefenstette,\ \BBA\ Rockt{\"a}schel}{Samvelyan
  et~al.}{2021}]{samvelyan2021minihack}
Samvelyan, M., Kirk, R., Kurin, V., Parker-Holder, J., Jiang, M., Hambro, E.,
  Petroni, F., Kuttler, H., Grefenstette, E., \BBA\ Rockt{\"a}schel, T.
  \BBOP2021\BBCP.
\newblock \BBOQ Minihack the planet: A sandbox for open-ended reinforcement
  learning research\BBCQ\
\newblock In {\Bem Thirty-fifth Conference on Neural Information Processing
  Systems Datasets and Benchmarks Track}.

\bibitem[\protect\BCAY{Santurkar, Tsipras, Ilyas,\ \BBA\ Madry}{Santurkar
  et~al.}{2018}]{batch_norm_smoothness}
Santurkar, S., Tsipras, D., Ilyas, A., \BBA\ Madry, A. \BBOP2018\BBCP.
\newblock \BBOQ How does batch normalization help optimization?\BBCQ\
\newblock In {\Bem Advances in Neural Information Processing Systems 31:
  NeurIPS, December 3-8, Montr{\'{e}}al, Canada}, \BPGS\ 2488--2498.

\bibitem[\protect\BCAY{Schaul, Borsa, Ding, Szepesvari, Ostrovski, Dabney,\
  \BBA\ Osindero}{Schaul et~al.}{2019}]{schaul2020adapting}
Schaul, T., Borsa, D., Ding, D., Szepesvari, D., Ostrovski, G., Dabney, W.,
  \BBA\ Osindero, S. \BBOP2019\BBCP.
\newblock \BBOQ Adapting behaviour for learning progress\BBCQ\
\newblock {\Bem CoRR}, {\Bem abs/1912.06910}.

\bibitem[\protect\BCAY{Schaul, Quan, Antonoglou,\ \BBA\ Silver}{Schaul
  et~al.}{2016}]{prioritized_exprep}
Schaul, T., Quan, J., Antonoglou, I., \BBA\ Silver, D. \BBOP2016\BBCP.
\newblock \BBOQ Prioritized experience replay\BBCQ\
\newblock In {\Bem 4th International Conference on Learning Representations,
  {ICLR}, San Juan, Puerto Rico, May 2-4, Conference Track Proceedings}.

\bibitem[\protect\BCAY{Schmitt, Hudson, Z{\'{\i}}dek, Osindero, Doersch,
  Czarnecki, Leibo, K{\"{u}}ttler, Zisserman, Simonyan,\ \BBA\ Eslami}{Schmitt
  et~al.}{2018}]{kickstarting}
Schmitt, S., Hudson, J.~J., Z{\'{\i}}dek, A., Osindero, S., Doersch, C.,
  Czarnecki, W.~M., Leibo, J.~Z., K{\"{u}}ttler, H., Zisserman, A., Simonyan,
  K., \BBA\ Eslami, S. M.~A. \BBOP2018\BBCP.
\newblock \BBOQ Kickstarting deep reinforcement learning\BBCQ\
\newblock {\Bem CoRR}, {\Bem abs/1803.03835}.

\bibitem[\protect\BCAY{Schrittwieser, Antonoglou, Hubert, Simonyan, Sifre,
  Schmitt, Guez, Lockhart, Hassabis, Graepel, et~al.}{Schrittwieser
  et~al.}{2020}]{schrittwieser2020mastering}
Schrittwieser, J., Antonoglou, I., Hubert, T., Simonyan, K., Sifre, L.,
  Schmitt, S., Guez, A., Lockhart, E., Hassabis, D., Graepel, T., et~al.
  \BBOP2020\BBCP.
\newblock \BBOQ Mastering atari, go, chess and shogi by planning with a learned
  model\BBCQ\
\newblock {\Bem Nature}, {\Bem 588\/}(7839), 604--609.

\bibitem[\protect\BCAY{Schulman, Levine, Abbeel, Jordan,\ \BBA\
  Moritz}{Schulman et~al.}{2015}]{schulman2015trust}
Schulman, J., Levine, S., Abbeel, P., Jordan, M.~I., \BBA\ Moritz, P.
  \BBOP2015\BBCP.
\newblock \BBOQ Trust region policy optimization\BBCQ\
\newblock In {\Bem Proceedings of the 32nd International Conference on Machine
  Learning, {ICML}, Lille, France, 6-11 July}, \lowercase{\BVOL}~37 of {\Bem
  {JMLR} Workshop and Conference Proceedings}, \BPGS\ 1889--1897. JMLR.org.

\bibitem[\protect\BCAY{Schulman, Wolski, Dhariwal, Radford,\ \BBA\
  Klimov}{Schulman et~al.}{2017a}]{schulman2017proximal}
Schulman, J., Wolski, F., Dhariwal, P., Radford, A., \BBA\ Klimov, O.
  \BBOP2017a\BBCP.
\newblock \BBOQ Proximal policy optimization algorithms\BBCQ.

\bibitem[\protect\BCAY{Schulman, Wolski, Dhariwal, Radford,\ \BBA\
  Klimov}{Schulman et~al.}{2017b}]{ppo_schulman_arxiv_17}
Schulman, J., Wolski, F., Dhariwal, P., Radford, A., \BBA\ Klimov, O.
  \BBOP2017b\BBCP.
\newblock \BBOQ Proximal policy optimization algorithms\BBCQ\
\newblock {\Bem CoRR}, {\Bem abs/1707.06347}.

\bibitem[\protect\BCAY{Sehgal, La, Louis,\ \BBA\ Nguyen}{Sehgal
  et~al.}{2019}]{ddpg_ga}
Sehgal, A., La, H.~M., Louis, S.~J., \BBA\ Nguyen, H. \BBOP2019\BBCP.
\newblock \BBOQ Deep reinforcement learning using genetic algorithm for
  parameter optimization\BBCQ\
\newblock {\Bem CoRR}, {\Bem abs/1905.04100}.

\bibitem[\protect\BCAY{Sharma, Lakshminarayanan,\ \BBA\ Ravindran}{Sharma
  et~al.}{2017}]{learning_to_repeat}
Sharma, S., Lakshminarayanan, A.~S., \BBA\ Ravindran, B. \BBOP2017\BBCP.
\newblock \BBOQ Learning to repeat: Fine grained action repetition for deep
  reinforcement learning\BBCQ\
\newblock In {\Bem 5th International Conference on Learning Representations,
  {ICLR}, Toulon, France, April 24-26, Conference Track Proceedings}.
  OpenReview.net.

\bibitem[\protect\BCAY{Silver, Huang, Maddison, Guez, Sifre, van~den Driessche,
  Schrittwieser, Antonoglou, Panneershelvam, Lanctot, Dieleman, Grewe, Nham,
  Kalchbrenner, Sutskever, Lillicrap, Leach, Kavukcuoglu, Graepel,\ \BBA\
  Hassabis}{Silver et~al.}{2016}]{alphago}
Silver, D., Huang, A., Maddison, C.~J., Guez, A., Sifre, L., van~den Driessche,
  G., Schrittwieser, J., Antonoglou, I., Panneershelvam, V., Lanctot, M.,
  Dieleman, S., Grewe, D., Nham, J., Kalchbrenner, N., Sutskever, I.,
  Lillicrap, T.~P., Leach, M., Kavukcuoglu, K., Graepel, T., \BBA\ Hassabis, D.
  \BBOP2016\BBCP.
\newblock \BBOQ Mastering the game of {G}o with deep neural networks and tree
  search\BBCQ\
\newblock {\Bem Nature}, {\Bem 529}, 484--489.

\bibitem[\protect\BCAY{Singh\ \BBA\ Dayan}{Singh\ \BBA\
  Dayan}{1996}]{singhdayan}
Singh, S.\BBACOMMA\  \BBA\ Dayan, P. \BBOP1996\BBCP.
\newblock \BBOQ Analytical mean squared error curves in temporal difference
  learning\BBCQ\
\newblock In {\Bem Proceedings of the 9th International Conference on Neural
  Information Processing Systems}, NIPS'96, \BPG\ 1054–1060, Cambridge, MA,
  USA. MIT Press.

\bibitem[\protect\BCAY{Sinha, Bharadhwaj, Srinivas,\ \BBA\ Garg}{Sinha
  et~al.}{2020}]{impact_nn_arch_rl_sinha_neurips20_deep_rl_workshop}
Sinha, S., Bharadhwaj, H., Srinivas, A., \BBA\ Garg, A. \BBOP2020\BBCP.
\newblock \BBOQ {D2RL:} deep dense architectures in reinforcement
  learning\BBCQ\
\newblock In {\Bem Deep Reinforcement Learning Workshop, NeurIPS},
  \lowercase{\BVOL}\ abs/2010.09163.

\bibitem[\protect\BCAY{Snel\ \BBA\ Whiteson}{Snel\ \BBA\
  Whiteson}{2010}]{multi_task_evolutionary_shaping_snel_gecco_10}
Snel, M.\BBACOMMA\  \BBA\ Whiteson, S. \BBOP2010\BBCP.
\newblock \BBOQ Multi-task evolutionary shaping without pre-specified
  representations\BBCQ\
\newblock In {\Bem Genetic and Evolutionary Computation Conference, {GECCO},
  Proceedings, Portland, Oregon, USA, July 7-11}, \BPGS\ 1031--1038. {ACM}.

\bibitem[\protect\BCAY{Song, Chen,\ \BBA\ Yue}{Song
  et~al.}{2019}]{multifidelity_bo}
Song, J., Chen, Y., \BBA\ Yue, Y. \BBOP2019\BBCP.
\newblock \BBOQ A general framework for multi-fidelity bayesian optimization
  with gaussian processes\BBCQ\
\newblock In Chaudhuri, K.\BBACOMMA\  \BBA\ Sugiyama, M.\BEDS, {\Bem
  Proceedings of the Twenty-Second International Conference on Artificial
  Intelligence and Statistics}, \lowercase{\BVOL}~89 of {\Bem Proceedings of
  Machine Learning Research}, \BPGS\ 3158--3167. PMLR.

\bibitem[\protect\BCAY{Song, Choromanski, Parker{-}Holder, Tang, Peng, Jain,
  Gao, Pacchiano, Sarl{\'{o}}s,\ \BBA\ Yang}{Song
  et~al.}{2021}]{es_enas_song21}
Song, X., Choromanski, K., Parker{-}Holder, J., Tang, Y., Peng, D., Jain, D.,
  Gao, W., Pacchiano, A., Sarl{\'{o}}s, T., \BBA\ Yang, Y. \BBOP2021\BBCP.
\newblock \BBOQ {ES-ENAS:} combining evolution strategies with neural
  architecture search at no extra cost for reinforcement learning\BBCQ\
\newblock {\Bem CoRR}, {\Bem abs/2101.07415}.

\bibitem[\protect\BCAY{Song, Du,\ \BBA\ Jackson}{Song
  et~al.}{2019}]{hyp_dep_gen}
Song, X., Du, Y., \BBA\ Jackson, J. \BBOP2019\BBCP.
\newblock \BBOQ An empirical study on hyperparameters and their interdependence
  for {RL} generalization\BBCQ.

\bibitem[\protect\BCAY{Song, Jiang, Tu, Du,\ \BBA\ Neyshabur}{Song
  et~al.}{2020}]{observational_overfitting}
Song, X., Jiang, Y., Tu, S., Du, Y., \BBA\ Neyshabur, B. \BBOP2020\BBCP.
\newblock \BBOQ Observational overfitting in reinforcement learning\BBCQ\
\newblock In {\Bem 8th International Conference on Learning Representations,
  {ICLR}, Addis Ababa, Ethiopia, April 26-30}. OpenReview.net.

\bibitem[\protect\BCAY{Spears}{Spears}{1995}]{AdaptingCrossOverinEA}
Spears, W.~M. \BBOP1995\BBCP.
\newblock \BBOQ Adapting crossover in evolutionary algorithms\BBCQ\
\newblock In McDonnell, J.~R., Reynolds, R.~G., \BBA\ Fogel, D.~B.\BEDS, {\Bem
  Proceedings of the Fourth Annual Conference on Evolutionary Programming, {EP}
  1995, San Diego, CA, USA, March 1-3, 1995}, \BPGS\ 367--384. A Bradford Book,
  {MIT} Press. Cambridge, Massachusetts.

\bibitem[\protect\BCAY{Srinivas, Krause, Kakade,\ \BBA\ Seeger}{Srinivas
  et~al.}{2010}]{gp_ucb}
Srinivas, N., Krause, A., Kakade, S.~M., \BBA\ Seeger, M.~W. \BBOP2010\BBCP.
\newblock \BBOQ Gaussian process optimization in the bandit setting: No regret
  and experimental design\BBCQ\
\newblock In {\Bem Proceedings of the 27th International Conference on Machine
  Learning (ICML-10), June 21-24, Haifa, Israel}, \BPGS\ 1015--1022. Omnipress.

\bibitem[\protect\BCAY{Stanley, Clune, Lehman,\ \BBA\ Miikkulainen}{Stanley
  et~al.}{2019}]{neuroevolution_nature}
Stanley, K., Clune, J., Lehman, J., \BBA\ Miikkulainen, R. \BBOP2019\BBCP.
\newblock \BBOQ Designing neural networks through neuroevolution\BBCQ\
\newblock {\Bem Nature Machine Intelligence}, {\Bem 1}.

\bibitem[\protect\BCAY{Stanley, D'Ambrosio,\ \BBA\ Gauci}{Stanley
  et~al.}{2009}]{hyperneat}
Stanley, K.~O., D'Ambrosio, D.~B., \BBA\ Gauci, J. \BBOP2009\BBCP.
\newblock \BBOQ A hypercube-based encoding for evolving large-scale neural
  networks\BBCQ\
\newblock {\Bem Artificial Life}, {\Bem 15\/}(2), 185--212.

\bibitem[\protect\BCAY{Stanley\ \BBA\ Miikkulainen}{Stanley\ \BBA\
  Miikkulainen}{2002}]{neat}
Stanley, K.~O.\BBACOMMA\  \BBA\ Miikkulainen, R. \BBOP2002\BBCP.
\newblock \BBOQ Evolving neural networks through augmenting topologies\BBCQ.
\newblock {\Bem 10\/}(2).

\bibitem[\protect\BCAY{Such, Madhavan, Conti, Lehman, Stanley,\ \BBA\
  Clune}{Such et~al.}{2018}]{such2018deep}
Such, F.~P., Madhavan, V., Conti, E., Lehman, J., Stanley, K.~O., \BBA\ Clune,
  J. \BBOP2018\BBCP.
\newblock \BBOQ Deep neuroevolution: Genetic algorithms are a competitive
  alternative for training deep neural networks for reinforcement
  learning\BBCQ.

\bibitem[\protect\BCAY{Sukhbaatar, Lin, Kostrikov, Synnaeve, Szlam,\ \BBA\
  Fergus}{Sukhbaatar et~al.}{2018}]{auto_curricula_sukhbaatar_iclr_18}
Sukhbaatar, S., Lin, Z., Kostrikov, I., Synnaeve, G., Szlam, A., \BBA\ Fergus,
  R. \BBOP2018\BBCP.
\newblock \BBOQ Intrinsic motivation and automatic curricula via asymmetric
  self-play\BBCQ\
\newblock In {\Bem 6th International Conference on Learning Representations,
  {ICLR}, Vancouver, BC, Canada, April 30 - May 3, Conference Track
  Proceedings}. OpenReview.net.

\bibitem[\protect\BCAY{Sutton\ \BBA\ Singh}{Sutton\ \BBA\
  Singh}{1994}]{Sutton94onstep-size}
Sutton, R.\BBACOMMA\  \BBA\ Singh, S.~P. \BBOP1994\BBCP.
\newblock \BBOQ On step-size and bias in temporal-difference learning\BBCQ\
\newblock In {\Bem Center for Systems Science, Yale University}, \BPGS\ 91--96.

\bibitem[\protect\BCAY{Sutton}{Sutton}{1991}]{dyna}
Sutton, R.~S. \BBOP1991\BBCP.
\newblock \BBOQ Dyna, an integrated architecture for learning, planning, and
  reacting\BBCQ\
\newblock {\Bem SIGART Bull.}, {\Bem 2\/}(4), 160–163.

\bibitem[\protect\BCAY{Sutton\ \BBA\ Barto}{Sutton\ \BBA\
  Barto}{2018}]{sutton_book_2018}
Sutton, R.~S.\BBACOMMA\  \BBA\ Barto, A.~G. \BBOP2018\BBCP.
\newblock {\Bem Reinforcement Learning: An Introduction\/} (Second \BEd).
\newblock The MIT Press.

\bibitem[\protect\BCAY{Sutton, McAllester, Singh,\ \BBA\ Mansour}{Sutton
  et~al.}{1999a}]{policy_gradients}
Sutton, R.~S., McAllester, D., Singh, S., \BBA\ Mansour, Y. \BBOP1999a\BBCP.
\newblock \BBOQ Policy gradient methods for reinforcement learning with
  function approximation\BBCQ\
\newblock In {\Bem Proceedings of the 12th International Conference on Neural
  Information Processing Systems}, NIPS'99, \BPG\ 1057–1063, Cambridge, MA,
  USA. MIT Press.

\bibitem[\protect\BCAY{Sutton, Precup,\ \BBA\ Singh}{Sutton
  et~al.}{1999b}]{options}
Sutton, R.~S., Precup, D., \BBA\ Singh, S.~P. \BBOP1999b\BBCP.
\newblock \BBOQ Between mdps and semi-mdps: {A} framework for temporal
  abstraction in reinforcement learning\BBCQ\
\newblock {\Bem Artif. Intell.}, {\Bem 112\/}(1-2), 181--211.

\bibitem[\protect\BCAY{Tan, Zhang, Coumans, Iscen, Bai, Hafner, Bohez,\ \BBA\
  Vanhoucke}{Tan et~al.}{2018}]{tan2018sim}
Tan, J., Zhang, T., Coumans, E., Iscen, A., Bai, Y., Hafner, D., Bohez, S.,
  \BBA\ Vanhoucke, V. \BBOP2018\BBCP.
\newblock \BBOQ Sim-to-real: Learning agile locomotion for quadruped
  robots\BBCQ\
\newblock In Kress{-}Gazit, H., Srinivasa, S.~S., Howard, T., \BBA\ Atanasov,
  N.\BEDS, {\Bem Robotics: Science and Systems XIV, Carnegie Mellon University,
  Pittsburgh, Pennsylvania, USA, June 26-30, 2018}.

\bibitem[\protect\BCAY{Tang, Nguyen,\ \BBA\ Ha}{Tang
  et~al.}{2020}]{attentionagent2020}
Tang, Y., Nguyen, D., \BBA\ Ha, D. \BBOP2020\BBCP.
\newblock \BBOQ Neuroevolution of self-interpretable agents\BBCQ\
\newblock In {\Bem Proceedings of the Genetic and Evolutionary Computation
  Conference}.

\bibitem[\protect\BCAY{Tassa, Erez,\ \BBA\ Todorov}{Tassa
  et~al.}{2012}]{tassa_ilqg_iros_12}
Tassa, Y., Erez, T., \BBA\ Todorov, E. \BBOP2012\BBCP.
\newblock \BBOQ Synthesis and stabilization of complex behaviors through online
  trajectory optimization\BBCQ\
\newblock In {\Bem 2012 {IEEE/RSJ} International Conference on Intelligent
  Robots and Systems, {IROS}, Vilamoura, Algarve, Portugal, October 7-12,
  2012}, \BPGS\ 4906--4913. {IEEE}.

\bibitem[\protect\BCAY{Team, Stooke, Mahajan, Barros, Deck, Bauer, Sygnowski,
  Trebacz, Jaderberg, Mathieu, McAleese, Bradley{-}Schmieg, Wong, Porcel,
  Raileanu, Hughes{-}Fitt, Dalibard,\ \BBA\ Czarnecki}{Team
  et~al.}{2021}]{xland2021}
Team, O. E.~L., Stooke, A., Mahajan, A., Barros, C., Deck, C., Bauer, J.,
  Sygnowski, J., Trebacz, M., Jaderberg, M., Mathieu, M., McAleese, N.,
  Bradley{-}Schmieg, N., Wong, N., Porcel, N., Raileanu, R., Hughes{-}Fitt, S.,
  Dalibard, V., \BBA\ Czarnecki, W.~M. \BBOP2021\BBCP.
\newblock \BBOQ Open-ended learning leads to generally capable agents\BBCQ\
\newblock {\Bem CoRR}, {\Bem abs/2107.12808}.

\bibitem[\protect\BCAY{Todorov, Erez,\ \BBA\ Tassa}{Todorov
  et~al.}{2012}]{todorov2012mujoco}
Todorov, E., Erez, T., \BBA\ Tassa, Y. \BBOP2012\BBCP.
\newblock \BBOQ {MuJoCo}: {A} physics engine for model-based control\BBCQ\
\newblock In {\Bem International Conference on Intelligent Robots and Systems
  ({IROS}'12)}, \BPGS\ 5026--5033. {IEEE}.

\bibitem[\protect\BCAY{Tran, Nguyen, Bruce, Crockett, Formenti, Phan, Payne,\
  \BBA\ Farmery}{Tran et~al.}{2021}]{tran2021simulation}
Tran, M., Nguyen, V., Bruce, R., Crockett, D., Formenti, F., Phan, P., Payne,
  S., \BBA\ Farmery, A. \BBOP2021\BBCP.
\newblock \BBOQ Simulation-based optimisation to quantify heterogeneity of
  specific ventilation and perfusion in the lung by the inspired sinewave
  test\BBCQ\
\newblock {\Bem Scientific reports}, {\Bem 11\/}(1), 1--10.

\bibitem[\protect\BCAY{Turchetta, Kolobov, Shah, Krause,\ \BBA\
  Agarwal}{Turchetta et~al.}{2020}]{turchetta-neurips20}
Turchetta, M., Kolobov, A., Shah, S., Krause, A., \BBA\ Agarwal, A.
  \BBOP2020\BBCP.
\newblock \BBOQ Safe reinforcement learning via curriculum induction\BBCQ\
\newblock In {\Bem Advances in Neural Information Processing Systems 33:
  NeurIPS, December 6-12, virtual}.

\bibitem[\protect\BCAY{van Bueren, Reed, Nguyen, Sheffield, van~der Ven,
  Osborne, Kroesbergen,\ \BBA\ Kadosh}{van Bueren
  et~al.}{2021}]{van2021personalized}
van Bueren, N., Reed, T., Nguyen, V., Sheffield, J., van~der Ven, S., Osborne,
  M., Kroesbergen, E., \BBA\ Kadosh, R.~C. \BBOP2021\BBCP.
\newblock \BBOQ Personalized closed-loop brain stimulation for effective
  neurointervention across participants\BBCQ\
\newblock {\Bem PLoS Computational Biology}, {\Bem 17}.

\bibitem[\protect\BCAY{van Hasselt, Guez,\ \BBA\ Silver}{van Hasselt
  et~al.}{2016}]{doubleQ}
van Hasselt, H., Guez, A., \BBA\ Silver, D. \BBOP2016\BBCP.
\newblock \BBOQ Deep reinforcement learning with double q-learning\BBCQ\
\newblock In {\Bem Proceedings of the Thirtieth {AAAI} Conference on Artificial
  Intelligence, February 12-17, Phoenix, Arizona, {USA}}, \BPGS\ 2094--2100.
  {AAAI} Press.

\bibitem[\protect\BCAY{van Rijn\ \BBA\ Hutter}{van Rijn\ \BBA\
  Hutter}{2018}]{Rijn-KDD18}
van Rijn, J.~N.\BBACOMMA\  \BBA\ Hutter, F. \BBOP2018\BBCP.
\newblock \BBOQ Hyperparameter importance across datasets\BBCQ\
\newblock In Guo, Y.\BBACOMMA\  \BBA\ Farooq, F.\BEDS, {\Bem Proceedings of the
  24th {ACM} {SIGKDD} International Conference on Knowledge Discovery {\&} Data
  Mining, {KDD} 2018, London, UK, August 19-23, 2018}, \BPGS\ 2367--2376.
  {ACM}.

\bibitem[\protect\BCAY{Vaswani, Shazeer, Parmar, Uszkoreit, Jones, Gomez,
  Kaiser,\ \BBA\ Polosukhin}{Vaswani et~al.}{2017}]{vaswanitransformers}
Vaswani, A., Shazeer, N., Parmar, N., Uszkoreit, J., Jones, L., Gomez, A.~N.,
  Kaiser, L.~u., \BBA\ Polosukhin, I. \BBOP2017\BBCP.
\newblock \BBOQ Attention is all you need\BBCQ\
\newblock In {\Bem Advances in Neural Information Processing Systems}.

\bibitem[\protect\BCAY{Veeriah, Hessel, Xu, Rajendran, Lewis, Oh, van Hasselt,
  Silver,\ \BBA\ Singh}{Veeriah et~al.}{2019}]{veeriah2019discovery}
Veeriah, V., Hessel, M., Xu, Z., Rajendran, J., Lewis, R.~L., Oh, J., van
  Hasselt, H.~P., Silver, D., \BBA\ Singh, S. \BBOP2019\BBCP.
\newblock \BBOQ Discovery of useful questions as auxiliary tasks\BBCQ\
\newblock In {\Bem Advances in Neural Information Processing Systems},
  \lowercase{\BVOL}~32.

\bibitem[\protect\BCAY{Veeriah, Zahavy, Hessel, Xu, Oh, Kemaev, van Hasselt,
  Silver,\ \BBA\ Singh}{Veeriah et~al.}{2021}]{veeriah2021options}
Veeriah, V., Zahavy, T., Hessel, M., Xu, Z., Oh, J., Kemaev, I., van Hasselt,
  H., Silver, D., \BBA\ Singh, S. \BBOP2021\BBCP.
\newblock \BBOQ Discovery of options via meta-learned subgoals\BBCQ\
\newblock {\Bem CoRR}, {\Bem abs/2102.06741}.

\bibitem[\protect\BCAY{Vieillard, Pietquin,\ \BBA\ Geist}{Vieillard
  et~al.}{2020}]{m-dqn}
Vieillard, N., Pietquin, O., \BBA\ Geist, M. \BBOP2020\BBCP.
\newblock \BBOQ Munchausen reinforcement learning\BBCQ\
\newblock In {\Bem Advances in Neural Information Processing Systems 33:
  NeurIPS, December 6-12, virtual}.

\bibitem[\protect\BCAY{Vinyals, Babuschkin, Czarnecki, Mathieu, Dudzik, Chung,
  Choi, Powell, Ewalds, Georgiev, Oh, Horgan, Kroiss, Danihelka, Huang, Sifre,
  Cai, Agapiou, Jaderberg,\ \BBA\ Silver}{Vinyals et~al.}{2019}]{alphastar}
Vinyals, O., Babuschkin, I., Czarnecki, W., Mathieu, M., Dudzik, A., Chung, J.,
  Choi, D., Powell, R., Ewalds, T., Georgiev, P., Oh, J., Horgan, D., Kroiss,
  M., Danihelka, I., Huang, A., Sifre, L., Cai, T., Agapiou, J., Jaderberg, M.,
  \BBA\ Silver, D. \BBOP2019\BBCP.
\newblock \BBOQ Grandmaster level in starcraft ii using multi-agent
  reinforcement learning\BBCQ\
\newblock {\Bem Nature}, {\Bem 575}.

\bibitem[\protect\BCAY{Volz, Schrum, Liu, Lucas, Smith,\ \BBA\ Risi}{Volz
  et~al.}{2018}]{evolving_mario_gan_gecco_18}
Volz, V., Schrum, J., Liu, J., Lucas, S.~M., Smith, A.~M., \BBA\ Risi, S.
  \BBOP2018\BBCP.
\newblock \BBOQ Evolving mario levels in the latent space of a deep
  convolutional generative adversarial network\BBCQ\
\newblock In {\Bem Proceedings of the Genetic and Evolutionary Computation
  Conference, {GECCO}, Kyoto, Japan, July 15-19}, \BPGS\ 221--228. {ACM}.

\bibitem[\protect\BCAY{Wan, Nguyen, Ha, Ru, Lu,\ \BBA\ Osborne}{Wan
  et~al.}{2021}]{wan2021casmopolitan}
Wan, X., Nguyen, V., Ha, H., Ru, B., Lu, C., \BBA\ Osborne, M.~A.
  \BBOP2021\BBCP.
\newblock \BBOQ Think global and act local: Bayesian optimisation over
  high-dimensional categorical and mixed search spaces\BBCQ\
\newblock In Meila, M.\BBACOMMA\  \BBA\ Zhang, T.\BEDS, {\Bem Proceedings of
  the 38th International Conference on Machine Learning}, \lowercase{\BVOL}\
  139 of {\Bem Proceedings of Machine Learning Research}, \BPGS\ 10663--10674.
  PMLR.

\bibitem[\protect\BCAY{Wang, King, Porcel, Kurth{-}Nelson, Zhu, Deck, Choy,
  Cassin, Reynolds, Song, Buttimore, Reichert, Rabinowitz, Matthey, Hassabis,
  Lerchner,\ \BBA\ Botvinick}{Wang et~al.}{2021}]{alchemy_wang_arxiv_21}
Wang, J.~X., King, M., Porcel, N., Kurth{-}Nelson, Z., Zhu, T., Deck, C., Choy,
  P., Cassin, M., Reynolds, M., Song, H.~F., Buttimore, G., Reichert, D.~P.,
  Rabinowitz, N.~C., Matthey, L., Hassabis, D., Lerchner, A., \BBA\ Botvinick,
  M. \BBOP2021\BBCP.
\newblock \BBOQ Alchemy: {A} structured task distribution for
  meta-reinforcement learning\BBCQ\
\newblock {\Bem CoRR}, {\Bem abs/2102.02926}.

\bibitem[\protect\BCAY{Wang, Kurth{-}Nelson, Tirumala, Soyer, Leibo, Munos,
  Blundell, Kumaran,\ \BBA\ Botvinick}{Wang et~al.}{2016}]{wang-l2rl-arxiv16}
Wang, J.~X., Kurth{-}Nelson, Z., Tirumala, D., Soyer, H., Leibo, J.~Z., Munos,
  R., Blundell, C., Kumaran, D., \BBA\ Botvinick, M. \BBOP2016\BBCP.
\newblock \BBOQ Learning to reinforcement learn\BBCQ\
\newblock {\Bem CoRR}, {\Bem abs/1611.05763}.

\bibitem[\protect\BCAY{Wang, Lehman, Clune,\ \BBA\ Stanley}{Wang
  et~al.}{2019}]{poet_wang_gecco_19}
Wang, R., Lehman, J., Clune, J., \BBA\ Stanley, K.~O. \BBOP2019\BBCP.
\newblock \BBOQ {POET:} open-ended coevolution of environments and their
  optimized solutions\BBCQ\
\newblock In {\Bem Proceedings of the Genetic and Evolutionary Computation
  Conference, {GECCO}, Prague, Czech Republic, July 13-17}, \BPGS\ 142--151.
  {ACM}.

\bibitem[\protect\BCAY{Wang, Lehman, Rawal, Zhi, Li, Clune,\ \BBA\
  Stanley}{Wang et~al.}{2020}]{enhanced_poet_wang_icml_20}
Wang, R., Lehman, J., Rawal, A., Zhi, J., Li, Y., Clune, J., \BBA\ Stanley,
  K.~O. \BBOP2020\BBCP.
\newblock \BBOQ Enhanced {POET:} open-ended reinforcement learning through
  unbounded invention of learning challenges and their solutions\BBCQ\
\newblock In {\Bem Proceedings of the 37th International Conference on Machine
  Learning, {ICML}, 13-18 July, Virtual Event}, \lowercase{\BVOL}\ 119 of {\Bem
  Proceedings of Machine Learning Research}, \BPGS\ 9940--9951. {PMLR}.

\bibitem[\protect\BCAY{Wang, Schaul, Hessel, van Hasselt, Lanctot,\ \BBA\
  de~Freitas}{Wang et~al.}{2016}]{dueling}
Wang, Z., Schaul, T., Hessel, M., van Hasselt, H., Lanctot, M., \BBA\
  de~Freitas, N. \BBOP2016\BBCP.
\newblock \BBOQ Dueling network architectures for deep reinforcement
  learning\BBCQ\
\newblock In {\Bem Proceedings of the 33nd International Conference on Machine
  Learning, {ICML}, New York City, NY, USA, June 19-24}, \lowercase{\BVOL}~48
  of {\Bem {JMLR} Workshop and Conference Proceedings}, \BPGS\ 1995--2003.
  JMLR.org.

\bibitem[\protect\BCAY{Watkins\ \BBA\ Dayan}{Watkins\ \BBA\
  Dayan}{1992}]{watkins1992q}
Watkins, C.~J.\BBACOMMA\  \BBA\ Dayan, P. \BBOP1992\BBCP.
\newblock \BBOQ Q-learning\BBCQ\
\newblock {\Bem Machine learning}, {\Bem 8\/}(3-4), 279--292.

\bibitem[\protect\BCAY{White\ \BBA\ White}{White\ \BBA\
  White}{2016}]{white2016greedy}
White, M.\BBACOMMA\  \BBA\ White, A. \BBOP2016\BBCP.
\newblock \BBOQ A greedy approach to adapting the trace parameter for temporal
  difference learning\BBCQ\
\newblock In {\Bem AAMAS}.

\bibitem[\protect\BCAY{Whiteson, Tanner, Taylor,\ \BBA\ Stone}{Whiteson
  et~al.}{2009}]{whiteson2009generalized}
Whiteson, S., Tanner, B., Taylor, M.~E., \BBA\ Stone, P. \BBOP2009\BBCP.
\newblock \BBOQ Generalized domains for empirical evaluations in reinforcement
  learning\BBCQ.

\bibitem[\protect\BCAY{Whitley, Gordon,\ \BBA\ Mathias}{Whitley
  et~al.}{1994}]{lamarckian}
Whitley, D., Gordon, V.~S., \BBA\ Mathias, K. \BBOP1994\BBCP.
\newblock \BBOQ Lamarckian evolution, the baldwin effect and function
  optimization\BBCQ\
\newblock In {\Bem Parallel Problem Solving from Nature --- PPSN III}.

\bibitem[\protect\BCAY{W{\"{o}}hlke, Schmitt,\ \BBA\ van Hoof}{W{\"{o}}hlke
  et~al.}{2020}]{woehlke-spl-aamas20}
W{\"{o}}hlke, J., Schmitt, F., \BBA\ van Hoof, H. \BBOP2020\BBCP.
\newblock \BBOQ A performance-based start state curriculum framework for
  reinforcement learning\BBCQ\
\newblock In {\Bem Proceedings of the 19th International Conference on
  Autonomous Agents and Multiagent Systems, {AAMAS} '20, Auckland, New Zealand,
  May 9-13, 2020}, \BPGS\ 1503--1511. International Foundation for Autonomous
  Agents and Multiagent Systems.

\bibitem[\protect\BCAY{Xu, Hoos,\ \BBA\ Leyton-Brown}{Xu
  et~al.}{2010}]{xu-aaai10a}
Xu, L., Hoos, H., \BBA\ Leyton-Brown, K. \BBOP2010\BBCP.
\newblock \BBOQ Hydra: Automatically configuring algorithms for portfolio-based
  selection\BBCQ\
\newblock In {\Bem Proceedings of the Twenty-fourth National Conference on
  Artificial Intelligence ({AAAI}'10)}, \BPGS\ 210--216. {AAAI} Press.

\bibitem[\protect\BCAY{Xu, van Hasselt,\ \BBA\ Silver}{Xu
  et~al.}{2018}]{metagradients}
Xu, Z., van Hasselt, H., \BBA\ Silver, D. \BBOP2018\BBCP.
\newblock \BBOQ Meta-gradient reinforcement learning\BBCQ\
\newblock In {\Bem Advances in Neural Information Processing Systems 31:
  NeurIPS, December 3-8, Montr{\'{e}}al, Canada}, \BPGS\ 2402--2413.

\bibitem[\protect\BCAY{Xu, van Hasselt, Hessel, Oh, Singh,\ \BBA\ Silver}{Xu
  et~al.}{2020}]{frodo}
Xu, Z., van Hasselt, H.~P., Hessel, M., Oh, J., Singh, S., \BBA\ Silver, D.
  \BBOP2020\BBCP.
\newblock \BBOQ Meta-gradient reinforcement learning with an objective
  discovered online\BBCQ\
\newblock In {\Bem Advances in Neural Information Processing Systems},
  \lowercase{\BVOL}~33, \BPGS\ 15254--15264.

\bibitem[\protect\BCAY{Yang, Sun,\ \BBA\ Narasimhan}{Yang
  et~al.}{2019}]{morl_yang_neurips_19}
Yang, R., Sun, X., \BBA\ Narasimhan, K. \BBOP2019\BBCP.
\newblock \BBOQ A generalized algorithm for multi-objective reinforcement
  learning and policy adaptation\BBCQ\
\newblock In Wallach, H.~M., Larochelle, H., Beygelzimer, A.,
  d'Alch{\'{e}}{-}Buc, F., Fox, E.~B., \BBA\ Garnett, R.\BEDS, {\Bem Advances
  in Neural Information Processing Systems 32: Annual Conference on Neural
  Information Processing Systems 2019, NeurIPS 2019, December 8-14, 2019,
  Vancouver, BC, Canada}, \BPGS\ 14610--14621.

\bibitem[\protect\BCAY{Yu, Quillen, He, Julian, Hausman, Finn,\ \BBA\
  Levine}{Yu et~al.}{2019}]{meta_world_yu_corl_2019}
Yu, T., Quillen, D., He, Z., Julian, R., Hausman, K., Finn, C., \BBA\ Levine,
  S. \BBOP2019\BBCP.
\newblock \BBOQ Meta-world: A benchmark and evaluation for multi-task and meta
  reinforcement learning\BBCQ\
\newblock In {\Bem Conference on Robot Learning (CoRL)}.

\bibitem[\protect\BCAY{Zahavy, Xu, Veeriah, Hessel, Oh, van Hasselt, Silver,\
  \BBA\ Singh}{Zahavy et~al.}{2020}]{stac}
Zahavy, T., Xu, Z., Veeriah, V., Hessel, M., Oh, J., van Hasselt, H., Silver,
  D., \BBA\ Singh, S. \BBOP2020\BBCP.
\newblock \BBOQ A self-tuning actor-critic algorithm\BBCQ\
\newblock In {\Bem Advances in Neural Information Processing Systems}.

\bibitem[\protect\BCAY{Zambaldi, Raposo, Santoro, Bapst, Li, Babuschkin, Tuyls,
  Reichert, Lillicrap, Lockhart, Shanahan, Langston, Pascanu, Botvinick,
  Vinyals,\ \BBA\ Battaglia}{Zambaldi et~al.}{2019}]{relational_rl}
Zambaldi, V.~F., Raposo, D., Santoro, A., Bapst, V., Li, Y., Babuschkin, I.,
  Tuyls, K., Reichert, D.~P., Lillicrap, T.~P., Lockhart, E., Shanahan, M.,
  Langston, V., Pascanu, R., Botvinick, M., Vinyals, O., \BBA\ Battaglia, P.~W.
  \BBOP2019\BBCP.
\newblock \BBOQ Deep reinforcement learning with relational inductive
  biases\BBCQ\
\newblock In {\Bem 7th International Conference on Learning Representations,
  {ICLR}, New Orleans, LA, USA, May 6-9}. OpenReview.net.

\bibitem[\protect\BCAY{Zhang, Rajan, Pineda, Lambert, Biedenkapp, Chua,
  Hutter,\ \BBA\ Calandra}{Zhang et~al.}{2021}]{pbtbt}
Zhang, B., Rajan, R., Pineda, L., Lambert, N.~O., Biedenkapp, A., Chua, K.,
  Hutter, F., \BBA\ Calandra, R. \BBOP2021\BBCP.
\newblock \BBOQ On the importance of hyperparameter optimization for
  model-based reinforcement learning\BBCQ\
\newblock In {\Bem The 24th International Conference on Artificial Intelligence
  and Statistics, {AISTATS}, April 13-15, Virtual Event}, \lowercase{\BVOL}\
  130 of {\Bem Proceedings of Machine Learning Research}, \BPGS\ 4015--4023.
  {PMLR}.

\bibitem[\protect\BCAY{Zhang, Vinyals, Munos,\ \BBA\ Bengio}{Zhang
  et~al.}{2018}]{overfitting_rl}
Zhang, C., Vinyals, O., Munos, R., \BBA\ Bengio, S. \BBOP2018\BBCP.
\newblock \BBOQ A study on overfitting in deep reinforcement learning\BBCQ\
\newblock {\Bem CoRR}, {\Bem abs/1804.06893}.

\bibitem[\protect\BCAY{Zhang, Abbeel,\ \BBA\ Pinto}{Zhang
  et~al.}{2020}]{zhang-vds-neurips20}
Zhang, Y., Abbeel, P., \BBA\ Pinto, L. \BBOP2020\BBCP.
\newblock \BBOQ Automatic curriculum learning through value disagreement\BBCQ\
\newblock In {\Bem Advances in Neural Information Processing Systems 33:
  NeurIPS, December 6-12, virtual}.

\bibitem[\protect\BCAY{Zheng, Oh,\ \BBA\ Singh}{Zheng
  et~al.}{2018}]{learned_intrinsic_rewards_zheng_neurips_18}
Zheng, Z., Oh, J., \BBA\ Singh, S. \BBOP2018\BBCP.
\newblock \BBOQ On learning intrinsic rewards for policy gradient methods\BBCQ\
\newblock In {\Bem Advances in Neural Information Processing Systems 31:
  NeurIPS, December 3-8, Montr{\'{e}}al, Canada}, \BPGS\ 4649--4659.

\bibitem[\protect\BCAY{Zintgraf, Shiarlis, Igl, Schulze, Gal, Hofmann,\ \BBA\
  Whiteson}{Zintgraf et~al.}{2020}]{varibad_zintgraf_iclr20}
Zintgraf, L.~M., Shiarlis, K., Igl, M., Schulze, S., Gal, Y., Hofmann, K.,
  \BBA\ Whiteson, S. \BBOP2020\BBCP.
\newblock \BBOQ Varibad: {A} very good method for bayes-adaptive deep {RL} via
  meta-learning\BBCQ\
\newblock In {\Bem 8th International Conference on Learning Representations,
  {ICLR}, Addis Ababa, Ethiopia, April 26-30}. OpenReview.net.

\bibitem[\protect\BCAY{Zou, Ren, Yan, Su,\ \BBA\ Zhu}{Zou
  et~al.}{2019}]{reward_shaping_via_meta_learning_arxiv_19}
Zou, H., Ren, T., Yan, D., Su, H., \BBA\ Zhu, J. \BBOP2019\BBCP.
\newblock \BBOQ Reward shaping via meta-learning\BBCQ\
\newblock {\Bem CoRR}, {\Bem abs/1901.09330}.

\end{thebibliography}
\bibliographystyle{theapa}

\newpage

\end{document}